\documentclass[letterpaper]{article} %

\newif\ifarxiv
\arxivtrue  %

\ifarxiv
\usepackage[draft]{aaai25-custom}
\usepackage[hyphens]{url}
\usepackage{hyperref}
\else
\usepackage{aaai25}  %
\usepackage[hyphens]{url}  %
\fi

\usepackage{graphicx} %
\usepackage{times}  %
\usepackage{helvet}  %
\usepackage{courier}  %
\usepackage{natbib}  %
\usepackage{caption} %
\usepackage{algorithm}
\usepackage{algorithmic}

\usepackage{newfloat}
\usepackage{listings}
\DeclareCaptionStyle{ruled}{labelfont=normalfont,labelsep=colon,strut=off} %
\floatstyle{ruled}
\newfloat{listing}{tb}{lst}{}
\floatname{listing}{Listing}
\usepackage{amsmath} %
\usepackage{amssymb} %
\usepackage{booktabs} %
\usepackage{adjustbox} %
\usepackage{etoolbox} %
\usepackage{import} %
\usepackage{enumitem} %
\usepackage{subcaption} %
\usepackage{tikz} %
\usepackage{svg} %

\newif\ifgetaaaiapprefs
\ifarxiv
  \getaaaiapprefsfalse
\else
  \getaaaiapprefsfalse

  \ifgetaaaiapprefs
    \let\oldlabel\label
    \renewcommand{\label}[1]{}

    \renewcommand{\cite}[1]{}
    \renewcommand{\Citeauthor}[1]{}
    \renewcommand{\citeauthor}[1]{}
    \renewcommand{\citep}[1]{}
    \renewcommand{\Citet}[1]{}
    \renewcommand{\citet}[1]{}
  \else
    \usepackage{xr}
  \fi
\fi

\newif\ifshowappendix
\showappendixfalse
\ifarxiv\showappendixtrue\fi
\ifgetaaaiapprefs\showappendixtrue\fi

\newcommand{\cpfivezerofive}{\texttt{base-victim}}
\newcommand{\cpfivezerofiveshort}{\texttt{base-v}}

\newcommand{\bsixty}{\texttt{may23-victim}}
\newcommand{\origcyclic}{\texttt{base-adversary}}  %
\newcommand{\origcyclicshort}{\texttt{base-a}}
\newcommand{\origcyclicmedium}{\texttt{base-adv}}
\newcommand{\attackbsixty}{\texttt{may23-adversary}} %
\newcommand{\attackbsixtymedium}{\texttt{may23-adv}}
\newcommand{\origtwotwoseven}{\texttt{base-adv-early}} %

\newcommand{\dectwentythree}{\texttt{dec23-victim}}
\newcommand{\dectwentythreeshort}{\texttt{dec23-v}}

\newcommand{\maytwentyfour}{\texttt{may24-victim}}
\newcommand{\maytwentyfourshort}{\texttt{may24-v}}

\newcommand{\vitvictim}{\texttt{ViT-victim}}

\newcommand{\vitvictimshort}{\texttt{ViT-v}}
\newcommand{\vitadversary}{\texttt{ViT-adversary}}
\newcommand{\vitadversaryshort}{\texttt{ViT-a}}
\newcommand{\vitadversarymedium}{\texttt{ViT-adv}}
\newcommand{\controlbten}{\texttt{control-victim}}

\newcommand{\attackiter}[1]{%
    \ifnum0<0#1\relax
        \texttt{a\textsubscript{#1}}%
    \else%
        $\texttt{a}_{#1}$%
    \fi
}
\newcommand{\defenseiter}[1]{%
    \ifnum0<0#1\relax
        \texttt{v\textsubscript{#1}}%
    \else%
        $\texttt{v}_{#1}$%
    \fi
}

\newcommand{\attackhnine}{\texttt{atari-adversary}} %
\newcommand{\attackhnineshort}{\texttt{atari-a}}
\newcommand{\stalladv}{\texttt{stall-adversary}}
\newcommand{\stalladvshort}{\texttt{stall-a}}
\newcommand{\contadv}{\texttt{continuous-adversary}} %
\newcommand{\contadvshort}{\texttt{cont-a}} %
\newcommand{\contadvmedium}{\texttt{cont-adv}} %
\newcommand{\largeadv}{\texttt{big-adversary}}
\newcommand{\largeadvmedium}{\texttt{big-adv}}
\newcommand{\largeadvshort}{\texttt{big-a}}
\newcommand{\koadv}{\texttt{gift-adversary}} %
\newcommand{\koadvshort}{\texttt{gift-a}} %
\newcommand{\koadvmedium}{\texttt{gift-adv}} %

\newcommand{\attacker}{adversary}
\newcommand{\Attacker}{Adversary}
\newcommand{\attackers}{adversaries}
\newcommand{\Attackers}{Adversaries}
\newcommand{\defender}{victim}
\newcommand{\Defender}{Victim}
\newcommand{\defenders}{victims}
\newcommand{\Defenders}{Victims}

\newcommand{\plotdiamond}{$\blacklozenge$}

\newif\ifsubmission
\makeatletter
\ifx\showauthors@on\@empty
  \submissiontrue
\else
  \submissionfalse
\fi
\makeatother

\ifsubmission
\newcommand{\demosite}{https://go-defense.netlify.app}
\newcommand{\codebase}{https://anonymized-see-supplementary-material}
\newcommand{\vitimplementation}{https://anonymized-see-supplementary-material/engines/KataGo-custom/blob/stable/python/model\_pytorch.py}

\newcommand{\anonkellinpelrine}{Anonymized}
\newcommand{\anonpelrine}{Anonymized}
\else
\newcommand{\demosite}{https://goattack.far.ai}
\newcommand{\codebase}{https://github.com/AlignmentResearch/go\_attack}
\newcommand{\vitimplementation}{https://github.com/AlignmentResearch/KataGo-custom/blob/stable/python/model\_pytorch.py}

\newcommand{\anonkellinpelrine}{Kellin Pelrine}
\newcommand{\anonpelrine}{Pelrine}
\fi

\newcommand{\maybehideurl}[2]{%
  \ifarxiv%
    \href{#1}{#2}%
  \else%
    #2%
  \fi%
}

\title{Can Go AIs Be Adversarially Robust?\ifarxiv\else\footnote{See the full
      version~\cite{tseng2024canarxiv} of this paper for appendices, which also
      contain some referenced figures and tables.}\fi}
\author{
  Tom Tseng\textsuperscript{\rm 1},
  Euan McLean\textsuperscript{\rm 1},
  Kellin Pelrine\footnote{Equal advising contribution.}\textsuperscript{\rm 1,\rm 2},
  Tony T.\ Wang\footnotemark[\ifarxiv1\else2\fi]\textsuperscript{\rm 3},
  Adam Gleave\footnotemark[\ifarxiv1\else2\fi]\textsuperscript{\rm 1}}
\affiliations{
    \textsuperscript{\rm 1}FAR.AI,
    \textsuperscript{\rm 2}Mila,
    \textsuperscript{\rm 3}MIT,\\
    \{tom, kellin, adam\}@far.ai,
    emc031@gmail.com,
    twang6@mit.edu
}

\begin{document}

\maketitle

\begin{abstract}
Prior work found that superhuman Go AIs can be defeated by simple adversarial strategies, especially ``cyclic'' attacks. In this paper, we study whether adding natural countermeasures can achieve robustness in Go, a favorable domain for robustness since it benefits from incredible average-case capability and a narrow, innately adversarial setting. We test three defenses: adversarial training on hand-constructed positions, iterated adversarial training, and changing the network architecture. We find that though some of these defenses protect against previously discovered attacks, none withstand freshly trained adversaries. Furthermore, most of the reliably effective attacks these adversaries discover are different realizations of the same overall class of cyclic attacks. Our results suggest that building robust AI systems is challenging even with extremely superhuman systems in some of the most tractable settings, and highlight two key gaps: efficient generalization of defenses, and diversity in training.
\ifarxiv For interactive examples of attacks and a link to our codebase, see \href{\demosite/}{\demosite}.\fi
\end{abstract}

\section{Introduction}
\label{sec:intro}

It is essential that AI systems work robustly, especially when they are deployed at a societal scale or are used in safety-critical systems. Unfortunately, although the \emph{average-case} performance of AI systems is rapidly improving, building AI systems with good \emph{worst-case} performance remains an unsolved problem. Indeed, Go AIs~\citep{wang2023adversarial}, image classifiers~\citep{liu2023comprehensive, croce2020robustbench}, and large language models~\citep{mazeika2024harmbench} all fail catastrophically when presented with adversarial inputs.

However, Go seems like it ought to be especially favorable for robustness. First, unlike computer vision models and LLMs where average-case performance struggles to surpass humans, Go AIs' average-case play massively outstrips human ability~\citep{silver2016,wang2023adversarial}. Second, the attack surface is narrow, with only a limited set of valid moves available to the adversary each turn. Third, unlike many non-game domains, Go is inherently adversarial and zero-sum, and there is no fundamental trade-off between clean and robust accuracy~\citep{tsipras2018robustness}. Furthermore, there is even a clear initial target: robustness to the ``cyclic attack'' found by \citet{wang2023adversarial}. This class of attacks is well-defined and consistent enough for humans to understand it, replicate it, and easily tell novice players how to defend against it---suggesting it should be feasible for a superhuman AI to defend against it.
The Go models \citet{wang2023adversarial} attacked were not designed with robustness in mind, giving
us hope that with more effort they could be made robust.

So, in this paper, we study if it is possible to make superhuman agents robust---that is, to make them have good worst-case performance---by focusing on the domain of Go.
Go has a proven track record of driving progress in AI, motivating the development of algorithms like AlphaZero~\citep{silver2016} and MuZero~\citep{schrittwieser2020}. If we could solve robustness here, it would provide a natural starting point for designing robust AI systems more broadly. In particular, we aim for a Go AI that a) does not make mistakes that humans can easily correct and b) cannot be reliably defeated by adversaries trained with a small amount of compute. These criteria, elaborated in Section~\ref{sec:background}, reflect two critical types of robustness: assurance that a system will not open up new vulnerabilities that humans could avoid, and that it will not be cheaply exploitable.

\begin{figure*}[t]
\centering
\includegraphics[width=0.75\textwidth]{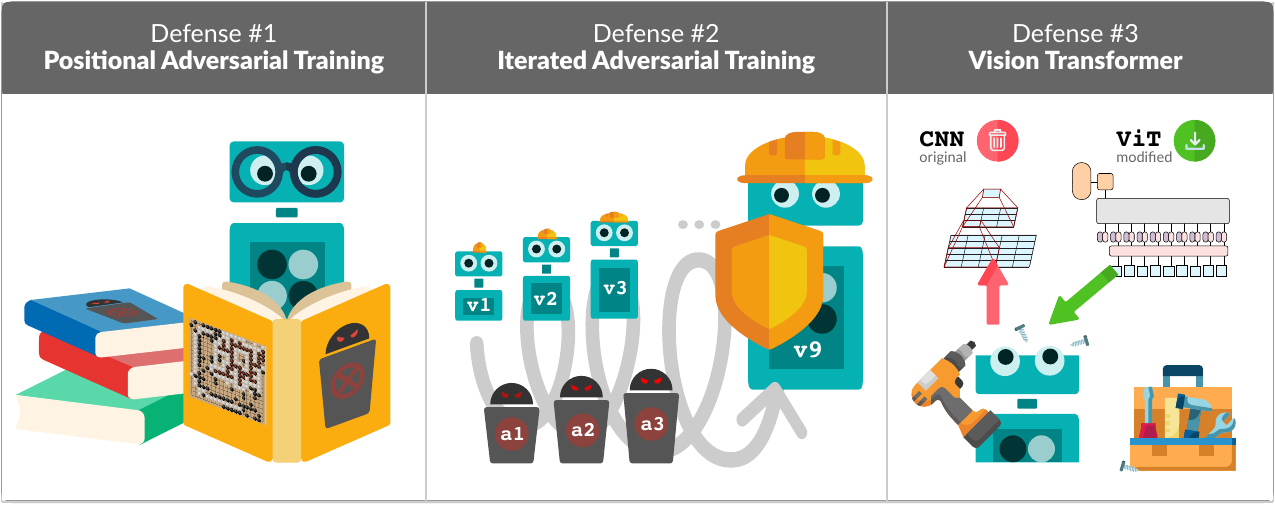}
\caption{Three strategies for defending Go AIs against adversarial attack. \textbf{Left}: Positional adversarial training has an agent ``study'' adversarial positions by performing self-play starting from those positions. \textbf{Middle}: Iterated adversarial training consists of multiple rounds of an adversary finding attacks and a victim learning to defend. \textbf{Right}: We replace KataGo's convolutional neural network (CNN) backbone with a vision transformer (ViT) backbone to see which vulnerabilities of Go AIs are caused by the inductive biases of CNNs.
}
\label{fig:summary}
\end{figure*}

Leveraging the state-of-the-art open-source Go AI KataGo~\citep{wu2021katago}, we investigate three natural defenses (Figure~\ref{fig:summary}). But we find that all three are ineffective---
given query access to the defended systems,
it is relatively cheap to train new adversaries that reliably defeat them and cause them to blunder in ways that humans would not. In fact, never mind all attacks---each defended system is still vulnerable to cyclic attacks,
demonstrating the defenses' poor generalization even in the favorable domain of Go.

More specifically, the first defense we study is \textbf{positional adversarial training}, which augments KataGo's training data with examples of \citet{wang2023adversarial}'s cyclic attack (Section~\ref{sec:internaladvtraining}). This intervention produces a defended agent that never loses against \citeauthor{wang2023adversarial}'s adversary.
Unfortunately, we find that adaptively fine-tuning \citeauthor{wang2023adversarial}'s adversary against the defended agent brings its win rate from 0\% back up to 92\%.
Moreover, this fine-tuned adversary wins using only a slight variant of its original strategy, and can be fine-tuned using just 8\% of the compute used for the defense.
(Appendix~\ref{app:networks} lists the compute used for each attack and defense.)
Furthermore, the defended agent is also vulnerable to a qualitatively new ``gift attack''
(Figure~\ref{fig:giftexample}).

Though ultimately unsuccessful, this first defense shows that defending against a \emph{fixed} attack is possible. This motivates our second defense, \textbf{iterated adversarial training}, which simulates an ``arms race'' between an \attacker{} continuously searching for new attacks and a \defender{} continuously building defenses against those attacks (Section~\ref{sec:advtraining}). Regrettably, we find that this scheme has the same weakness as positional adversarial training.
The defended agent is robust to \citet{wang2023adversarial}'s original cyclic attack, but
through fine-tuning we find a variant of the cyclic attack that defeats the
defended agent 90\% of the time using 26\% of the total compute used to train
the defended agent (Figure~\ref{fig:a9example}). Another variant wins 81\% of
the time using 5\% of the compute, though it loses efficacy when the defended agent uses more search (Figure~\ref{fig:validateex}).

The final defense we test is replacing the convolutional neural network (CNN) backbone used by KataGo with a \textbf{vision transformer} (ViT) backbone (Section~\ref{sec:vit}). Is the cyclic vulnerability caused by poor inductive biases of CNNs? We disprove that hypothesis by training the world’s first professional-level ViT-based Go AI and show that it too is beaten by cyclic attacks. Given \citet{wang2023adversarial} found the cyclic vulnerability was widespread across different Go AIs, with CNNs ruled out as the cause, a likely remaining candidate is the self-play training process. Playing only one opponent identical in strength and reasoning process may discourage exploration and get the model stuck in local equilibria, learning a way to assess vulnerability of groups that fails catastrophically on cyclic ones.

Our negative results in Go are evidence that achieving neural net robustness in
other domains will require concerted effort---we cannot assume that merely
boosting average-case performance to superhuman levels and adding simple
defenses will be enough to solve robustness.
State-of-the-art Go agents fail to represent a feature (cyclic shapes) that
amateur humans easily learn, even when using methods such as adversarial
training that provide numerous training examples where this feature is critical
to solving the task correctly. This failure makes them easily exploitable,
even by humans (Appendix~\ref{app:humanatk}). Building robust Go AIs may be
possible but will likely require significant design changes
(Section~\ref{sec:discussion}).

\section{Threat Model, Robustness, Attack Method}
\label{sec:background}

\paragraph{Threat Model}
We follow the threat model of \citet{wang2023adversarial} set in a two-player zero-sum Markov game~\citep{shapley1953}.
A threat actor trains an \emph{\attacker{}} agent to win against a \emph{\defender{}} agent.
The threat actor has \emph{gray-box access} to the \defender{}: they can run inference on the \defender{}'s policy network on arbitrary inputs.
However, the \attacker{} does not have direct access to the \defender{}'s weights and cannot take gradients through the \defender{}.

\paragraph{Defining Robustness}
We aim to make agents that are robust. Unlike settings like $\epsilon$-ball-robust image classification~\citep{croce2020robustbench}, it is not immediately obvious what it means for a Go agent to be robust. In this work, we introduce and use three complementary definitions of robustness centered around the notion of being minimally exploitable by adversaries.

Firstly, we aim to make our agents \textit{human robust}, meaning that they cannot be made to commit game-losing blunders that a human would not commit (Appendix~\ref{app:humanrobustness}). Secondly, we aim to
make our agents have high \emph{training-compute robustness}, meaning that it should take a large amount of compute to train an adversary that can defeat a \defender{} (Appendix~\ref{app:train-compute-robustness}). Finally, and more speculatively, we aim to make our agents have high \emph{inference-compute robustness}, meaning that our agents should be able to efficiently overcome vulnerabilities by using additional compute at inference-time (Appendix~\ref{app:test-compute-robustness}). We pick these definitions of robustness because they are applicable to both Go policies and more general AI agents.

\paragraph{Attack Method}
The \defenders{} are based on KataGo~\citep{wu2020accelerating}, the strongest open-source Go AI system. Like AlphaZero~\citep{silver2018}, it trains with self-play and selects moves using Monte Carlo tree search (MCTS; the amount of search is quantified by \emph{visits}) combined with a neural net.

We use \citet{wang2023adversarial}'s state-of-the-art attack method to produce \attackers{} for adversarial training and testing defenses. \citeauthor{wang2023adversarial} train an \attacker{} with \emph{victim-play} where the \attacker{} plays games against a frozen copy of the \defender{}, and training data is saved only from the \attacker{}'s moves.
The \attacker{} selects moves using Adversarial MCTS (A-MCTS), a modification of MCTS that queries the \defender{}'s network when traversing MCTS nodes corresponding to the opponent's move.
We follow \citeauthor{wang2023adversarial} by evaluating adversaries with 600 A-MCTS visits per move.
We typically train our adversaries by initializing them to \origcyclic{} (\citeauthor{wang2023adversarial}'s original cyclic adversary), which achieved a 97\% win rate against a 2022 KataGo network \cpfivezerofive{} (which \citeauthor{wang2023adversarial} calls \texttt{Latest}) at 4096 victim visits.
To find diverse attacks, we initialize some experiments to the non-cyclic \origtwotwoseven{}, which is the first \origcyclic{} checkpoint that beats \cpfivezerofive{} at 1 victim visit.
See Appendices~\ref{app:networks} and \ref{app:training-background} for details on these networks and training parameters.

\section{Positional Adversarial Training}
\label{sec:internaladvtraining}

KataGo's official training run performed adversarial training on board positions exhibiting the cyclic attack.
We show that KataGo's adversarially trained networks remain exploitable despite this by training new adversaries that
beat the strongest KataGo network from the end of 2023, which we call \dectwentythree{}, and a stronger model \maytwentyfour{} from May 2024.
One \attacker{},
\koadv{}, defeats \dectwentythree{} in 91\% of games (at 512 victim visits) using a new non-cyclic exploit where the \defender{} repeatedly gifts the \attacker{} two stones, though it does not scale to high victim visits as well as cyclic attacks.
We then find that \dectwentythree{} and \maytwentyfour{} are both vulnerable to cyclic attacks, with
another \attacker{} \largeadv{} winning 56\% of games against \maytwentyfour{} (65,536 victim visits).
The attacks can be replicated by a human expert (Appendix~\ref{app:humanatk}).

\subsection{Defense Methodology}\label{sec:katagoadvtrain-description}

We target models from KataGo's main training run, which began to include adversarial training against cyclic positions after the discovery of \origcyclic{}.
Since December 2022,
0.08\% of KataGo's self-play games have been initialized from a set of hand-written positions based on \origcyclic{}'s strategy~\citep{wu2022advtraining,wu2023advtrainingannouncement}.
Other positions were added as online Go players found different configurations of cyclic positions, growing the fraction of seeded self-play games to a few tenths of a percent~\citep{wu2023advtrainingdata}.
The resulting models defended well against \origcyclic{}.

Despite this positive result, \citet{wang2023adversarial} were able to fine-tune \origcyclic{} to produce \attackbsixty{} achieving a 47\% win rate against an adversarially trained KataGo checkpoint \bsixty{} at 4096 visits.
Building on \citeauthor{wang2023adversarial}'s evaluation, we test \dectwentythree{} and \maytwentyfour{} which have had over twice as much adversarial training as \bsixty{}.

\begin{figure*}[tb]
\centering
\input{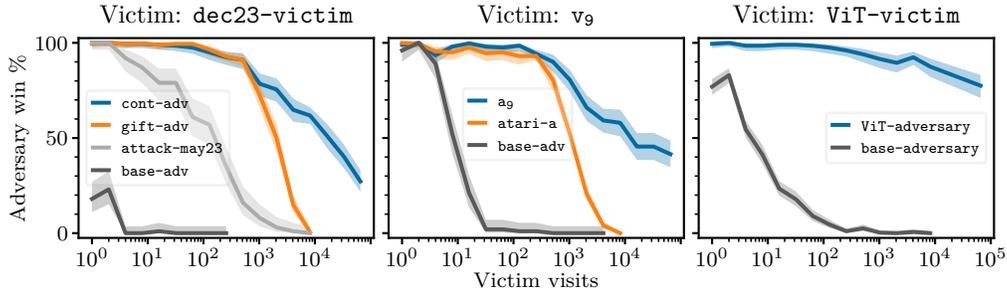}
\caption{Win rate (y-axis) of adversaries (legend) for varying amounts of search visits (x-axis) given to victims (plot title). The adversary win rate declines with victim search budget; however, some adversaries generalize better to higher victim visit counts than others. Shaded regions are 95\% Clopper-Pearson confidence intervals in this and following figures.}
\label{fig:vs-visits-combined}
\end{figure*}

\begin{figure*}[tb]
\centering
\begin{subfigure}[t]{0.24\textwidth}
    \includesvg[width=0.98\textwidth]{figs/boards/gift_example-2col.svg}
    \caption{
    \maybehideurl{\demosite/positional-adversarial-training?row=0\#dec23-vs-gift-board}{\dectwentythree{} (W) \\ vs. \koadv{} (B)}}
    \label{fig:giftexample}
\end{subfigure}
\begin{subfigure}[t]{0.24\textwidth}
    \includesvg[width=0.98\textwidth]{figs/boards/large_example-2col.svg}
    \caption{
    \maybehideurl{\demosite/positional-adversarial-training?row=0\#may24-vs-big-board}{\maytwentyfour{} (W) \\ vs. \largeadv{} (B)}}
    \label{fig:largeadv_boardstate}
\end{subfigure}
\begin{subfigure}[t]{0.24\textwidth}
    \includesvg[width=0.98\textwidth]{figs/boards/h9_example-2col.svg}
    \caption{
    \maybehideurl{\demosite/iterated-adversarial-training?row=0\#v9-vs-a9-board}{\defenseiter{9} (W) vs.\ \attackiter{9} (B)}}
    \label{fig:a9example}
\end{subfigure}
\begin{subfigure}[t]{0.24\textwidth}
    \includesvg[width=0.98\textwidth]{figs/boards/vit_example-2col.svg}
    \caption{
    \maybehideurl{\demosite/vit?row=0\#vit-vs-vit-adversary-board}{\vitvictim{} (W) \\ vs. \vitadversary{} (B)}}
    \label{fig:vitexample}
\end{subfigure}
\caption{Our learned adversarial strategies are qualitatively distinct.
\subref{fig:largeadv_boardstate}, \subref{fig:a9example}, \subref{fig:vitexample} show cyclic attacks with the \textcolor{red}{$\boldsymbol{\times}$} groups soon to be captured; these attacks use different styles of inside shapes, though these shapes have little impact on optimal play and are all easy for a human to navigate correctly. The \koadv{} in \subref{fig:giftexample} follows a different strategy, inducing the victim (white) to play the stone marked \textcolor{red}{$\boldsymbol{\times}$} ``gifting'' the adversary two stones it can capture by playing at $\triangle$. %
\ifarxiv Each subcaption links to a complete game history on our \href{\demosite}{website}.
\fi
}
\label{fig:boardstates}
\end{figure*}

\subsection{The Gift \Attacker{}}
\label{sec:attack-b18}

The \koadv{} was initialized from the early \origtwotwoseven{} checkpoint, encouraging exploration, and fine-tuned using victim-play against \dectwentythree{}.
Appendix~\ref{app:giftattack} gives more training details.
The attack wins 91\% of games against \dectwentythree{} (at 512 victim visits---above the superhuman threshold of 64 visits, see Appendix~\ref{app:elo}) after training with just 6\% as much compute as the victim.
The \koadv{} does not scale to high victim visits as well as the cyclic adversaries from Section~\ref{sec:large-adv} (Figure~\ref{fig:vs-visits-combined}).
However, the attack reveals a non-cyclic, qualitatively new exploit against KataGo (Figure~\ref{fig:giftexample}).

In particular, the \attacker{} sets up a ``sending-two-returning-one'' situation where the \defender{} unnecessarily gifts the \attacker{} two stones and needs to capture one back.
However, the \defender{}'s recapture is blocked by positional superko rules.\footnote{To prevent infinite loops, most rule sets include a \emph{superko rule} forbidding repetition of a previous board state (``positional superko'', or ``situational superko'' if the repeated state is expanded to include not only the board position but also whose turn it is to move).}
The \attacker{} sets up the position such that it resurrects of one of its dead groups, leading to a disaster for the \defender{}. %
The \defender{}'s gift is strange given that the \defender{} was trained with superko rules and has a neural net input feature marking moves that are illegal due to superko, and moreover
the scenario would not benefit the \defender{} even if superko rules were not in play.

\subsection{Cyclic Attacks}
\label{sec:large-adv}

We also find that \dectwentythree{} is still vulnerable to cyclic attacks by fine-tuning \attackbsixty{} to produce a cyclic adversary \contadv{} (Appendix~\ref{app:continuous}) that wins 65\% of games against \dectwentythree{} at 4096 victim visits.
In February 2024, KataGo's developer added positions from \koadv{} and \contadv{} into KataGo' adversarial training data set, but still
we were able to fine-tune \attackbsixty{} to produce yet another cyclic adversary, \largeadv{}, that beats \maytwentyfour{} (see Appendix~\ref{app:largeadv} for training details).

Cyclic attacks still work even at high victim visits, as \largeadv{} achieves a 56\% win rate against \maytwentyfour{} at 65536 victim visits.
The shape of the cyclic group is oblong and bigger than \citeauthor{wang2023adversarial}'s original cyclic attack, hence the name \largeadv{} (Figure~\ref{fig:largeadv_boardstate}).
The win rate is not as high as
\citeauthor{wang2023adversarial} achieved against the non-adversarially trained \cpfivezerofive{},
suggesting that adversarial training complicates and may narrow the range of attacks.
But at its current scale in KataGo, adversarial training has failed to comprehensively eliminate the cyclic vulnerability.

\section{Iterated Adversarial Training}
\label{sec:advtraining}

The previous section shows that an adversarially trained agent can still be vulnerable to new attacks.
Can we create a robust agent by repeatedly defending against new attacks until the space of possible attacks is exhausted?
In this section, we design an iterated adversarial training procedure that alternately trains a \defender{} and an \attacker{}.
Our procedure produced a \defender{} that was largely robust to the attacks it observed, losing only a low single-digit percentage of games.
However, the \defender{} still was not robust to new attacks, as we were able to train a new \attacker{} to exploit the final \defender{}.

\subsection{Methodology}
\label{sec:advtraining_method}

Our approach differs from KataGo's adversarial training (Section~\ref{sec:katagoadvtrain-description}) in three key ways.
First, we perform \emph{iterated} adversarial training, with multiple rounds of attack and defense to train against a broader range of attacks.
Second, we include a higher proportion of adversarial games in the training data: since our priority is robustness, we are more willing to take a hit in average-case capabilities than KataGo's developer.
Third, we play games directly against the adversary: this method is less sample-efficient than starting from hand-curated positions, but is more scalable and does not require domain-specific knowledge.

We label the \attacker{} and \defender{} at iteration $n$ of adversarial training ``\attackiter{n}'' and ``\defenseiter{n}''.
We initialize the \defender{} to \defenseiter{0} = \cpfivezerofive{} and the \attacker{} to \attackiter{0} = \origcyclic{} which \Citeauthor{wang2023adversarial} trained to defeat \cpfivezerofive{}.
Each subsequent iteration involves training the victim to be robust against the latest adversary, then training an adversary to attack this hardened victim.
We repeat this process for 9 iterations.

\defenseiter{n} is fine-tuned from \defenseiter{n-1}. 18\% of games were played against a frozen copy of \attackiter{n-1} and all other games were self-play.
This mixture teaches the \defender{} to be robust to the attack while preserving its Go capabilities.
We stop the training when the \defender{}'s win rate plateaus.
\attackiter{n} is fine-tuned from \attackiter{n-1} using victim-play.
We stop its training after either the \attacker{} wins most games at a threshold victim visit count, or a set maximum compute budget is reached.
See Appendix~\ref{app:adversarialtraining} for details.

\begin{figure*}[btp]
\centering
\import{figs/plots}{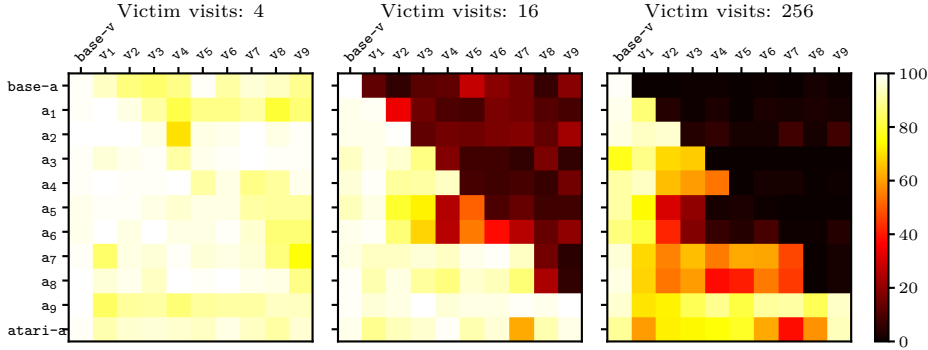}
\caption{Win rate of all \attackers{} ($x$-axis) against all \defenders{} ($y$-axis) throughout iterated adversarial training for varying victim visits (plot title). The \attacker{} \attackiter{n} is typically able to beat the \defender{} \defenseiter{n} it is trained to exploit (top-left-to-bottom-right diagonal), especially at 16 visits or less (middle and left plots). However, given at least 16 visits (middle and right) the \defender{} \defenseiter{n} is typically able to beat the \attacker{} \attackiter{n-1} it trained against (elements immediately below main diagonal) along with all previous iterations \attackiter{n-2}, \attackiter{n-3}, $\cdots$. See Figure~\ref{fig:win-rate-heatmap-full} for an extended version including other \attackers{}, \defenders{} and visit counts.}
\label{fig:win-rate-heatmap-iterated-adv}
\end{figure*}

\subsection{Results}
\label{sec:advtraining_results}

Each defender \defenseiter{n} learned an effective defense against the \attacker{} \attackiter{n-1} but rarely reached a 100\% win rate despite \attackiter{n-1} following a weak, degenerate strategy.
We tested the final \defender{} \defenseiter{9} by pitting it against the final attacker \attackiter{9} as well as validation \attackers{} trained separately from the iterated \attackers{} \attackiter{1}, $\ldots$, \attackiter{9}.
We found the \defender{} was still vulnerable to both attacks but is somewhat less so at high visit counts, indicating that iterated adversarial training offers partial protection against attacks.

\subsubsection{Robustness Against the Iterated \Attackers{}}
\label{sec:robustness_against_training_\attacker{}}

Each \defender{} \defenseiter{n} achieves a high win rate against the \attacker{} \attackiter{n-1} it was trained against when \defenseiter{n} uses at least 16 visits of search (Figure~\ref{fig:win-rate-heatmap-iterated-adv}, middle).
\defenseiter{n} quickly learns to beat \attackiter{n-1} $>95\%$ of the time (Figure~\ref{fig:vs-gpu-days-h-each}; 256 visits).
However, our defense runs rarely reached a 100\% win rate, and the \defenders{} were persistently vulnerable at extremely low visits (Figure~\ref{fig:win-rate-heatmap-iterated-adv}, left).

Both \defenders{} and \attackers{} are able to beat opponents from all previous iterations.
This is clearly shown by the adversary win rate in Figure~\ref{fig:win-rate-heatmap-iterated-adv} being much higher \emph{above} the diagonal (\attacker{} playing against older \defender{}) than \emph{below} the diagonal (\defender{} playing against older \attacker{}).
In the \defender{} case, this can be explained by the training window being large enough to contain data from all previous iterations. By contrast, since \attacker{} training takes many more time steps, all data from one iteration exits the training window during the next iteration. This suggests that the \attacker{} strategy transfers well to previous \defenders{}.

The final \defender{} \defenseiter{9} is still vulnerable even at high visits.
Our final \attacker{} \attackiter{9} wins 42\% of the time against \defenseiter{9} even at 65,536 visits.
We trained \attackiter{9} for longer than preceding \attackers{}, but its total training compute was still only 26\% of \defenseiter{9}'s.
(Though \attackiter{5}, \attackiter{6}, and \attackiter{8} have relatively poor
win rates against their corresponding \defenders{}, \attackiter{9} is able to
beat every \defenseiter{n} even at high \defender{} visits
(Figure~\ref{fig:win-rate-heatmap-full}). Therefore we think that
\defenseiter{5}, \defenseiter{6}, and \defenseiter{8} are no more robust than
\defenseiter{9} but instead benefited from their \attackers{}
getting stuck in local minima during training.)

All \attackers{} \attackiter{n} exploit a cyclic group, but there are qualitative variations in the size and location of that group, other stones, and especially the \attacker{}'s group inside the \defender{}'s cyclic group (elaborated in Appendix~\ref{app:iterated-qualitative}).
For example, \attackiter{9} favors a small, minimal interior group (Figure~\ref{fig:a9example}). To humans, the differences are subtle, and the difficulty of defending against them does not vary significantly. But to the \defenders{} \defenseiter{n}, the representations learned do not appear to generalize smoothly between these variations.

\subsubsection{Robustness Against New \Attackers{}}\label{sec:attack-h9}
\label{sec:robustness_against_validation_\attacker{}}

\begin{figure*}[tb]
    \centering
    \input{figs/plots/compare-gpu-days-cyclic-vs-attack-h9.pgf}
    \caption{Win rate ($y$-axis) of \origcyclic{} vs \cpfivezerofive{} (---) and \attackhnine{} vs \defenseiter{9} ($\cdots$) by training compute ($x$-axis), including the 164 GPU days training \attackhnine{}'s initialization checkpoint \origtwotwoseven{}. The checkpoint marked \plotdiamond{} is used for evaluation.}
    \label{fig:compare-gpu-days-cyclic-vs-attack-h9}
\end{figure*}

Though the final iterated \defender{} \defenseiter{9}
bests all previous \attackers{} \attackiter{1} to \attackiter{8},
the ultimate judge of a defense is whether it works against real attacks.
To evaluate this, we train a new \attacker{} \attackhnine{} (initialized to \origtwotwoseven{}) against \defenseiter{9} (Figure~\ref{fig:compare-gpu-days-cyclic-vs-attack-h9}).
This is analogous to a randomly initialized \attacker{} trained to first beat the publicly available KataGo checkpoint \cpfivezerofive{} at 1 visit and then---without access to any intermediate adversarial training checkpoints \defenseiter{1},\ldots,\defenseiter{8}---trained to attack \defenseiter{9}.
\attackhnine{} still learns a cyclic attack, and we name it after its tendency to leave many stones in ``atari'' (capturable next move) while forming an elongated cyclic shape (Fig.~\ref{fig:validateex}).

\attackhnine{} wins 81\% of the time against \defenseiter{9} playing with 512 visits despite being trained with less than 5\% of \defenseiter{9}'s compute,
and even without any A-MCTS search, \attackhnine{} wins 13\% of games (Appendix~\ref{app:adv-visits}).
The attack quickly learns to exploit \defenseiter{9} at low visits, winning over $60\%$ of the time against \defenseiter{9} at 256 visits after just 500 V100 GPU days (Figure~\ref{fig:compare-gpu-days-cyclic-vs-attack-h9}), sooner than our original \origcyclic{} \attacker{} learned to exploit \cpfivezerofive{}.

However, \defenseiter{9} proves harder to attack at high visits than \cpfivezerofive{}.
Quadrupling to 1024 visits takes slightly more than $4\times$ the compute, largely due to the increased cost of playing training games at higher visits.
\attackhnine{} plateaus after 1401 GPU days (\plotdiamond) with a meager 4\% win rate at 4096 visits.
We also trained another adversary \stalladv{} (Appendix~\ref{app:stall}), initialized from the later checkpoint \origcyclic{}, to attack \defenseiter{9}, but despite learning a distinct cyclic attack, its attack fails at high victim visits and its training plateaus like \attackhnine{}.
In contrast, \origcyclic{} generalized rapidly to beat \cpfivezerofive{} at higher visits.

\section{Vision Transformers}
\label{sec:vit}

\citet{wang2023adversarial}'s attack works not only against KataGo but also against a range of other superhuman Go AIs such as
\maybehideurl{https://online-go.com/game/51321265}{ELF OpenGo}~\citep{tian2019elf},
\maybehideurl{https://online-go.com/game/51356405}{Leela Zero}~\citep{pascutto2019leela},
\maybehideurl{https://online-go.com/game/51375020}{Sai}~\citep{Morandin2019},
\maybehideurl{https://www.bilibili.com/video/BV1Ls4y147Es/?share\_source=copy\_web\&t=97}{Golaxy}~\citep{golaxy},
and
\maybehideurl{https://h5.foxwq.com/txwqshare/index.html?chessid=1676910620010001365&boardsize=19}{FineArt}~\citep{fineart}.
Though it is possible that each of these systems is vulnerable to the cyclic attack for a different reason, it is more likely that shared properties such as their convolutional neural network (CNN) backbone cause their shared vulnerability.%
\footnote{Golaxy and FineArt are closed-source but likely use the same design principles as other Go AIs.}
Indeed, KataGo's developer proposed that vulnerability to cyclic attacks may be a result of the CNN backbone learning a local algorithm for classifying if a group is alive that fails to generalize to larger groups~\citep{polytope2023}.
However, we demonstrate that superhuman Go AIs with vision transformer (ViT) backbones are also susceptible to cyclic attacks.
This suggests the shared weakness is either AlphaZero-style training or deep learning more generally.

Since no prior work has trained strong Go AIs with a ViT architecture,%
\footnote{\citet{sagri2024vision} and \citet{wu2024restnet} train ViT-based and hybrid CNN-transformer Go AIs but did not validate the strength of their systems or release weights. Moreover, they used supervised learning on games generated by CNN agents and humans, whereas we trained our ViT agents only on ViT-generated self-play data.}
we first trained the world's first professional-level ViT-based Go AI.
We follow a training recipe similar to the one used by KataGo~\citep{katagotraining:2022}, except we replace the CNN backbone with a ViT (Appendix~\ref{app:vit}). Our strongest ViT network, a 16-layer model \vitvictim{}, was trained for 563 V100 GPU-days. It is slower to train than a CNN agent and weaker at the same inference budget, but by our estimation it still reaches top-human levels when playing with 32,768 visits. This estimate is derived from benchmarking against KataGo, pitting our agent against players on the KGS Go Server, and winning two out of three games against Go professionals (Appendix~\ref{app:elo}).

Despite the new architecture, Figure~\ref{fig:vs-visits-combined} shows that our \vitvictim{} (at 65536 visits) remains vulnerable to the cyclic attack, losing 78\% of games to a fine-tuned version of \origcyclic{}, which we call \vitadversary{} (Appendix~\ref{app:vit:adversary}).
\vitadversary{}'s strategy resembles other cyclic attacks but is qualitatively distinct in its tendency to produce small groups inside the cyclic one, and dense board states with limited open space (Figure~\ref{fig:vitexample}).
This attack can be replicated by a human expert (Appendix~\ref{app:humanatk}).

Remarkably, \vitvictim{} (at 512 visits) also loses 2.5\% of games to the original \origcyclic{}---similar to the zero-shot transfer to CNN Go AIs such as ELF OpenGo reported by \citet{wang2023adversarial}.
\origcyclic{} certainly does not win through strong Go ability: it is a very weak strategy that loses to amateur human players~\citep{wang2023adversarial}.
\vitvictim{} even misevaluates cyclic positions on boards as small as
11\texttimes11 (Appendix~\ref{app:board-size}).
We conclude that CNN architectures are not the cause of the cyclic
vulnerability, though we do not rule out the possibility that a much
deeper ViT or some other architecture would solve the cyclic vulnerability.

\section{Related Work}

We focus on robustness against \emph{adversarial policies}: strategies designed to make an opponent perform poorly.
Adversarial policies give an empirical lower bound for an agent's \emph{exploitability}: its worst-case loss relative to
Nash equilibria~\citep{timbers2022}.
\Citet{gleave2020} previously explored such policies in a zero-sum game between simulated humanoids trained with self-play.
The policies~\citep{bansal2018emergent} attacked by \citeauthor{gleave2020} were below human performance, raising the question: were the agents vulnerable because of their limited capability?
To investigate this, \Citet{wang2023adversarial} searched for adversarial policies against the superhuman Go AI KataGo~\cite{wu2020accelerating}, finding a strategy that beats KataGo in $97\%$ of games.

We focus on KataGo~\citep{wu2020accelerating} as it is the most capable open-source Go AI.
Moreover, other superhuman open-source Go AIs such as ELF OpenGo~\citep{tian2019elf} and Leela OpenZero~\citep{pascutto2019} all follow the same basic AlphaZero-style training architecture.
However, alternative multi-agent reinforcement learning methods may be more robust.
Approaches that maintain a population of strategies are promising~\citep{vinyals2019grandmaster,czempin2022,lanctot2017}.
Another alternative, counterfactual regret minimization~\cite{zinkevich2007}, has been used to beat professional human poker players~\cite{brown2017}.
Furthermore, \Citet{perolat2022mastering} found a way to approximate Nash equilibria~\cite{perolat2021} that scaled to the board game Stratego, whose game tree is $10^{175}$ times larger than Go's.

We replace the CNN backbone of KataGo with a vision transformer (ViT) and train the ViT agent to a superhuman level, finding it to be slower to train than a CNN agent and weaker at the same inference budget.
By contrast, \citet{sagri2024vision} found the transformer-based EfficientFormer architecture~\citep{li2022efficientformer} performed similarly to CNNs for Go---however, their models were trained only with supervised learning, not self-play.
Transformers have been investigated more thoroughly in chess.
Our results are consistent with \citet{czech2023representation} who found that CNNs are stronger at chess than both ViTs and a ViT-CNN hybrid at a given inference budget.
Yet transformers have shown strong performance, with the transformer-based Leela Chess Zero~\citep{pascutto2018lc0,lc02023smolgen} winning the Top Chess Engine Championship Cup 11~\citep{tcec2023cup11}.

Although we find ViTs weaker than CNNs in average-case capabilities, our primary metric is \emph{robustness}.
Past research in image classification indicates ViTs are modestly more robust than CNNs against adversarial perturbations and other out-of-distribution inputs~\citep{benz2021adversarial,shao2022on,bhojanapalli2021understanding,zhang2022delving,paul2022vision}, although some research contests this~\citep{bai2021transformers,mahmood2021on,tang2022robustart,pinto2022impartial,wang2023can}.
Even if ViTs are not more robust, their differing inductive biases might cause them to fail in \emph{different} ways to CNNs, with prior work finding that ViTs are more vulnerable to patch perturbations~\citep{fu2022patchfool}.
Surprisingly, we find that  \citet{wang2023adversarial}'s attack transfers zero-shot to our ViT agent.
In concurrent work, \citet{wu2024restnet} find that a hybrid CNN-transformer architecture beats a similarly sized CNN in the average case and is more robust
yet still vulnerable against a set of cyclic positions from
\citet{wang2023adversarial}.

\section{Discussion and Future Work}
\label{sec:discussion}

We explore three natural approaches for defending against adversarial attacks in Go: adversarial training with hand-constructed positions, iterated adversarial training, and using a ViT instead of a CNN. The defenses make attacks harder, increasing the amount of compute needed to successfully beat the defended systems. However, none of the defenses make attacks impossible---the attack algorithm from~\citet{wang2023adversarial} is always able to find a successful attack, often with a small fraction of the compute used to train the victims.
Moreover, none of the defenses achieve the robustness of a human (Appendix~\ref{app:humanrobustness}), and humans are even able to execute several attacks against the defenses (Appendix~\ref{app:humanatk}).
Our results highlight the challenge of defending against all possible attacks, or even all possible cyclic attacks, suggesting an offense-defense balance \citep{Jervis1978CooperationUT} favoring attackers.

We do, however, find it cheap to defend (relative to the cost to attack) against any \emph{fixed} (i.e. non-adaptive) attack.
This suggests that it may be possible to fully defend a Go AI by training against a large enough corpus of attacks.
We propose two complementary routes to fully robust Go AIs.

The first approach is to increase the size of the attack corpus by developing new attack algorithms that require less compute to train a variety of adversaries. Our version of iterated adversarial training was bottlenecked by the attack component, which took 18x more compute than the defense component (see Table~\ref{tab:iterated_training_cost}). Reducing the time taken to find new attacks, e.g., with relaxed~\citep{christiano2019worstcase,hubinger2019relaxed} or latent adversarial training~\citep{sankaranarayanan2018regularizing,casper2024defending}, would allow for training against many more attacks.

The second approach is to increase the sample efficiency of adversarial training by making the \defender{} generalize from a limited number of adversarial strategies. Existing algorithms for training Go AIs do not generalize in this way: \maytwentyfour{} and \defenseiter{9} remain vulnerable to cyclic attacks even after training against many variants of cyclic attacks.

There are also routes to robustness besides adversarial training. For example, multi-agent reinforcement learning schemes like PSRO~\citep{lanctot2017} or DeepNash~\citep{perolat2022mastering} may be able to automatically discover strategies like the cyclic attack and train them away.
Another possibility is to change the threat model and use online or \emph{stateful} defenses~\citep{chen2020stateful} which can dynamically update the \defender{} in tandem with \attackers{} who are trying to learn an exploit.

Our results highlight that existing difficulties in building robust AI systems may not be resolved by scaling up average-case capabilities and applying straightforward defenses. We recommend further systematic evaluations of robustness and how it behaves under increasing capabilities in other domains, especially ones where frontier models are beginning to attain new superhuman levels. But if we are unable to achieve robustness in strongly superhuman agents in the well-defined and self-contained domain of Go, achieving robustness in more open-ended real-world applications will likely be even more challenging.

\ifsubmission\else
\section*{Acknowledgments}
\addcontentsline{toc}{section}{Acknowledgments}

We thank David Wu for discussing KataGo and its adversarial training with us and for helping us qualitatively describe the two attacks we found in Section~\ref{sec:internaladvtraining},
Adrià Garriga-Alonso for infrastructure support when running experiments, and ChengCheng Tan for helping create Figure~\ref{fig:summary}. We also thank ChengCheng Tan, Derik Kauffmann, Siao Si Looi, David Wu, Micah Carroll, and Daniel Filan for feedback on early drafts. Finally, we thank Yilun Yang (7 dan professional) and Ryan Li (4 dan professional) for playing \vitvictim{} to evaluate its strength, and Matthew Harwit for helping connect the authors with professional Go players.

\section*{Disclosure of funding}
\addcontentsline{toc}{section}{Disclosure of funding}
Tom Tseng, Euan McLean, and Adam Gleave were employed by FAR.AI, a non-profit research institute, and supported by FAR.AI's unrestricted funds. Kellin Pelrine was supported by funding from IVADO and by the Fonds de recherche du Queb\'ec. Tony Wang was supported by a Lightspeed grant.

\section*{Author contributions}
\addcontentsline{toc}{section}{Author contributions}
Tom Tseng was the primary technical individual contributor, implementing the majority of the code and running the majority of the experiments.
Euan McLean prepared an initial draft of the paper from high-level comments provided by technical contributors, edited the resulting paper, and coordinated the write-up.
Kellin Pelrine, Tony Wang, and Adam Gleave were joint co-advisors throughout the project, and helped with paper writing and editing.
In addition, Kellin Pelrine analysed the Go games and replicated attacks by hand, Tony Wang set up KGS bots for human evaluation, and Adam Gleave managed the project.
\fi

{\small\bibliography{refs}}

\appendix

\ifgetaaaiapprefs
  \let\label\oldlabel
\fi

\ifarxiv\else
\fi
\ifshowappendix
\section{KataGo Networks Reference}\label{app:networks}

We built on top of KataGo~\cite{wu2020accelerating}, which was the strongest open-source Go AI system at the time of conducting our research.
KataGo learns via self-play using an AlphaZero-style training procedure~\citep{silver2018}.
The agent selects moves with Monte-Carlo Tree Search (MCTS), using a neural network to propose and evaluate moves.
The neural network contains a policy head that outputs a probability distribution over the next move and a value head that estimates the win rate from the current state.
KataGo trains its policy head to mimic the outcome of tree search and its value head to predict whether the agent wins the self-play game.

We evaluate and fine-tune a variety of KataGo models.
We refer to each model's architecture as \texttt{b$B$c$C$} where $B$ is the number of \emph{blocks} in the convolutional residual network and $C$ is the number of channels.
We refer to each model by \texttt{b$B$c$C$-s$S$m} where $S$ is the number of million time steps for which the model has been trained.
We may omit the channel term \texttt{c$C$} when there is no ambiguity. All of our \attackers{} have 6 blocks and 96 channels, abbreviated to \texttt{b6c96} or just \texttt{b6}.

The \defender{}s we attack are either \texttt{b40c256} or \texttt{b18c384} networks.
The \texttt{b18c384} networks use a new convolution-based architecture with modified bottleneck blocks~\cite{he2016deep,katago2024methods}.
They were introduced into KataGo's official training run in 2023, becoming the strongest networks by the end of the year.
The inference cost of these \texttt{b18} networks is similar to that of standard b40 networks.
Given the same inference compute budget per move to perform search, they outperform the best standard \texttt{b40} and \texttt{b60c320} networks.

\begin{figure*}[htbp]
\centering
\includegraphics[width=0.95\textwidth]{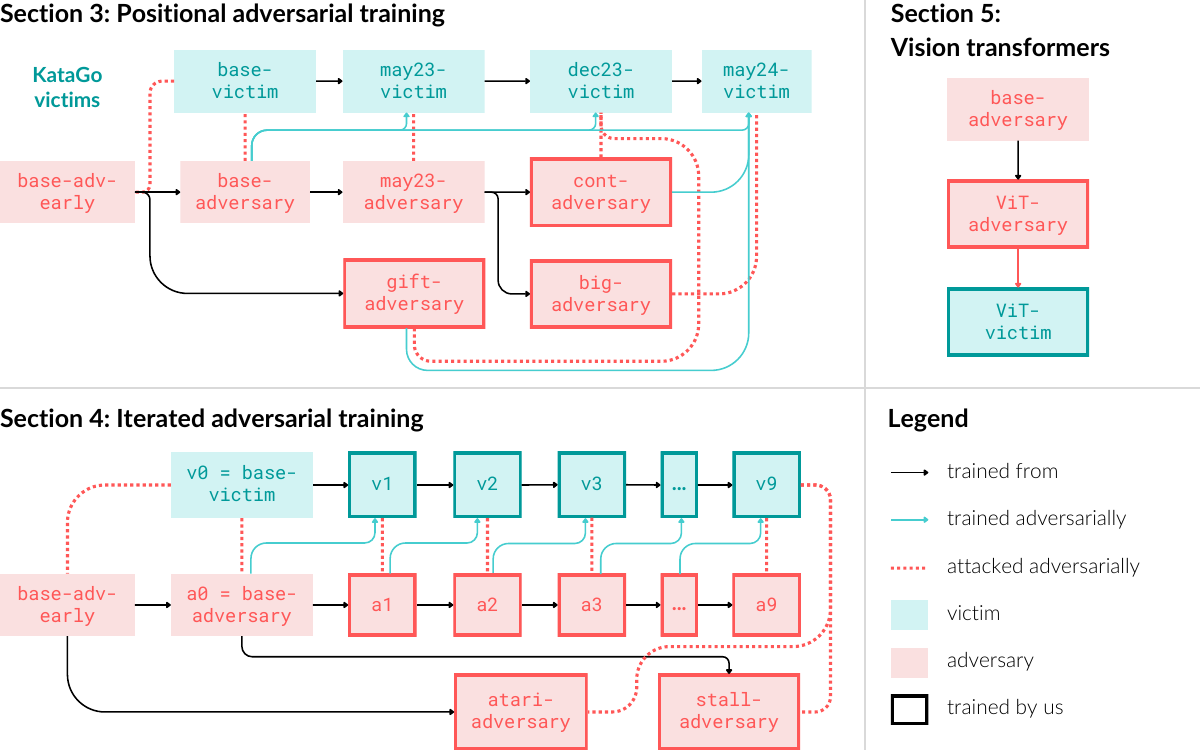}
\caption{A summary of the relationships among the Go-playing agents created by KataGo (the victims not trained by us), \Citet{wang2023adversarial} (the adversaries not trained by us), and our work (every model marked with a thick border). For instance, KataGo model \dectwentythree{} is trained from \bsixty{}, adversarially trains against board positions based on \origcyclic{}, and is adversarially attacked by \contadv{} and \koadv{}.
}
\label{fig:model-relationships}
\end{figure*}

In Table~\ref{tab:allvictims} we enumerate all \defenders{} used in this paper, comprising official KataGo networks, those developed by \citet{wang2023adversarial} and those developed in this work. In Table~\ref{tab:allattackers} we enumerate all \attackers{} used in this work, including our own and those developed by \citeauthor{wang2023adversarial}. Figure~\ref{fig:model-relationships} illustrates the relationships among these models.

\begin{table*}[hbtp]
\begin{adjustbox}{center} %
\begin{tabular}{llllllp{5cm}} %
\toprule %
\textbf{Name} & \multicolumn{2}{c}{\textbf{Params}} & \multicolumn{2}{c}{\textbf{Training}} & \textbf{Date} & \textbf{Description} \\ %
& \textbf{B} & \textbf{C} & \textbf{Steps (M)} & \textbf{GPU-days} & & \\
\midrule %

\cpfivezerofive{} & 40 & 256 & 11841 & 21681 & 2022-06 &

Original target KataGo network for \citet{wang2023adversarial}'s adversarial attack. 
Full name: kata1-b40c256-s11840935168-d2898845681.
\\

\texttt{b60-s6729m} & 60 & 320 & 6729 & 23099 & 2022-12 &
The last KataGo network before adversarial training began.
Not used in our paper but is a
reference point for how long KataGo's adversarial training has been running.
Full name: kata1-b60c320-s6729327872-d3057177418.
\\

\bsixty{} & 60 & 320 & 7702 & 25888 & 2023-05 &
KataGo network that had received 5 months' worth of adversarial training against cyclic positions. 
Full name: kata1-b60c320-s7701878528-d3323518127.
\\ %

\dectwentythree{} & 18 & 384 & 8527 & 33482 & 2023-12 &

KataGo network that had received 1 year's worth of adversarial training against cyclic positions. 
Full name: kata1-b18c384nbt-s8526915840-d3929217702.
\\ %

\maytwentyfour{} & 18 & 384 & 9997 & 41511 & 2024-05 &

KataGo network that had received 1.4 years' worth of adversarial training against cyclic positions. 
Full name: kata1-b18c384nbt-s9996604416-d4316597426.
\\ %

\hline
\\[-0.75em]
\defenseiter{n} & 40 & 256 & -- & -- & -- &

The victim at iteration $n$ of our iterated adversarial training (see Section~\ref{sec:advtraining}). \defenseiter{0} is warm-started from \cpfivezerofive{}. See Appendix~\ref{app:adversarialtraining} for breakdown. \\ %

\defenseiter{9} & 40 & 256 & 12097 & 28296 & 2024-01 &

Our final iterated adversarially trained victim (see Section~\ref{sec:advtraining}), warm-started from \cpfivezerofive{}. \\ %

\vitvictim{} & 16 & 384 & 650 & 563 & 2024-01 &

A network we trained from scratch using the same approach as KataGo but with the CNN backbone replaced with a vision transformer. \\ %

\bottomrule %
\end{tabular}
\end{adjustbox}
\vspace{1em}
\caption{All \defender{} networks used in this work with \textbf{B}locks, \textbf{C}hannels, training steps (in millions), and estimated compute cost in V100 GPU-days. Appendix~\ref{app:compute-katago} explains the compute estimates for the KataGo networks. The training cost for \defenseiter{9} includes both the training cost of all iterated \defenders{} and all iterated \attackers{} up to and including \attackiter{8}. The KataGo networks are available for download via their full name on https://katagotraining.org/networks/.} %
\label{tab:allvictims} 
\end{table*}

\begin{table*}[hbtp]
\begin{adjustbox}{center}
\begin{tabular}{lllp{1.8cm}p{6.2cm}} %
\toprule %
\textbf{Name} & \multicolumn{2}{c}{\textbf{Training}} & \textbf{Attack} & \textbf{Description} \\ %
& \textbf{Steps (M)} & \textbf{GPU-days} & \textbf{Style} & \\
\midrule %

\origcyclic{} & 545 & 2223 & cyclic & 

Original attack trained by \Citet{wang2023adversarial} from scratch to defeat the KataGo network \cpfivezerofive{}. \\

\origtwotwoseven{} & 227 & 164 & non-cyclic & 

The first checkpoint able to defeat \cpfivezerofive{} at one victim visit from the \origcyclic{} training run. \\

\attackbsixty{} & 713 & 3378 & cyclic & \origcyclic{} fine-tuned by \Citet{wang2023adversarial} to defeat KataGo's adversarially trained network \bsixty{}.\\
\hline
\\[-0.75em]

\koadv  & 878 & 1865 & gift & 

A network we trained using victim-play to defeat \dectwentythree{}, using a coarse-grained curriculum and fine-tuned from \origtwotwoseven{}. \\

\contadv{} & 1343 & 4476 & cyclic & 

A network we trained using victim-play to defeat \dectwentythree{}, using a fine-grained curriculum and fine-tuned from \attackbsixty{}. \\

\largeadv{} & 929 & 3671 & cyclic & 

A network we trained using victim-play to defeat \maytwentyfour{}, using a coarse-grained curriculum and fine-tuned from \attackbsixty{}. \\

\hline
\\[-0.75em]
\attackiter{n}  & -- & -- & cyclic & 

The adversary at iteration $n$ of our iterated adversarial training (see Section~\ref{sec:advtraining}). \attackiter{0} is fine-tuned from \origcyclic{}. See Appendix~\ref{app:adversarialtraining} for breakdown.
\\

\attackiter{9}  & 4132 & 7337 & cyclic & 

The final adversary resulting from our iterated adversarial training (see Section~\ref{sec:advtraining}).
\\

\attackhnine{} & 791 & 1401 & complex cyclic & 

A network we trained using victim-play to defeat \defenseiter{9} fine-tuned from \origtwotwoseven{}, to test the general robustness of iterated adversarial training.
\\

\stalladv{} & 642 & 2434 & cyclic & 

A network we trained using victim-play to defeat \defenseiter{9} fine-tuned from \origcyclic{}, to test the general robustness of iterated adversarial training.
\\

\hline
\\[-0.75em]
\vitadversary{}  & 871 & 2632 & cyclic & 

A network we trained using victim-play to defeat \vitvictim{}, fine-tuned from \origcyclic{}.
\\

\bottomrule %
\end{tabular}
\end{adjustbox}
\vspace{1em}
\caption{All \attacker{} networks used in this work with training steps (in millions) and estimated compute cost (in V100 GPU days). The \attackers{} use a 6 block, 96 channel KataGo CNN architecture \texttt{b6c96}.
The training cost includes the cost of the model the \attacker{} was initialized from. For example,
\largeadv{}'s cost includes the cost of \attackbsixty{} (not just the cost of fine-tuning), 
and the training cost for \attackiter{9} consists of the cost of all iterated \attackers{} as well as \origcyclic{}.} %
\label{tab:allattackers} %
\end{table*}

\begin{table*}[hbtp]
\begin{adjustbox}{center}
\begin{tabular}{ll|ll|ll} %
\toprule %
\multicolumn{2}{c|}{\textbf{\Defender{}}} & \multicolumn{2}{c|}{\textbf{Opponent}} & \multicolumn{2}{c}{\textbf{Opponent vs \Defender{}}} \\
\textbf{Name} & \textbf{Visits} & \textbf{Name} & \textbf{Visits} & \textbf{Compute (\%)} & \textbf{Win rate (\%)} \\[0.25em]

\hline & & & & & \\[-0.75em]

\cpfivezerofive{} & 4096 & \origcyclic{} & 600 & 10 & 97 \\

\cpfivezerofive{} & $10^7$ & \origcyclic{} & 600 & 10 & 72 \\

\bsixty{} & 4096 & \attackbsixty{} & 600 & 13 & 47 \\[0.25em] %

\hline & & & & & \\[-0.75em]

\dectwentythree{} & 512 & \koadv{} & 600 & 6 & 91 \\ %

\dectwentythree{} & 4096 & \contadv{} & 600 & 13 & 65 \\ %

\dectwentythree{} & 65536 & \contadv{} & 600 & 13 & 27 \\ %

\maytwentyfour{} & 4096 & \largeadv{} & 600 & 9 & 75 \\ %

\maytwentyfour{} & 65536 & \largeadv{} & 600 & 9 & 56 \\ %

\defenseiter{9} & 4096 & \cpfivezerofive{} & 4096 & 77 & 66 \\ %

\defenseiter{9} & 4096 & \attackiter{9} & 600 & 26 & 59 \\ %

\defenseiter{9} & 65536 & \attackiter{9} & 600 & 26 & 42 \\ %

\defenseiter{9} & 512 & \attackhnine{} & 600 & 5 & 81 \\ %

\defenseiter{9} & 512 & \stalladv{} & 600 & 9 & 91 \\ %

\vitvictim{} & 512 & \origcyclic{} & 600 & 393 & 2.5 \\ %

\vitvictim{} & 65536 & \vitadversary{} & 600 & 467 & 78 \\ %

\bottomrule %
\end{tabular}
\end{adjustbox}
\vspace{1em}
\caption{The \attacker{} win rate and fraction of opponent's compute used to train the opponent (right) for various \defenders{} (left) and opponents (middle). In most cases the \defender{} was trained with much more compute than the opponent; the exception is \vitvictim{} which was trained for a relatively brief period, $4\times$ less than \origcyclic{}, although the additional fine-tuning compute used to train \vitadversary{} was still less than that of \vitvictim{}. We standardize on an \attacker{} search budget of 600 visits for all evaluations. The non-adversarial opponent, \cpfivezerofive{}, is evaluated at the same number of visits as the \defender{}. The first three rows show evaluations performed by \Citet{wang2023adversarial}.} %
\label{tab:allgames} %
\end{table*}

\section{Definitions of Robustness}
\label{app:robustness-defs}

In this section, we propose three complementary definitions for a Go policy being ``robust''. Although our definitions are targeted at Go policies, we believe the core ideas behind them are also applicable to more general AI agents. Each of the following three subsections introduces a definition of robustness, states the motivations behind it, and discusses how the definition could be extended to more general agents.

\subsection{Human Robustness}
\label{app:humanrobustness}

Our first definition of robustness targets the concern that AI systems may fail in situations where humans would succeed. We call a system that does not have this failure mode \emph{human-robust}. Intuitively, a system is human-robust if its worst-case performance is better than human average-case performance -- that is, a human-robust system is not just sometimes but consistently superhuman.

We formalize this as follows: a system $S$ is \textbf{human-robust} in an environment $E$ if there are no points at which an omniscient observer could ask a human $H$ having ordinary skill in the art to make some decisions in place of the system $S$, such that $H$ -- without the benefit of hindsight --would consistently and intentionally produce a substantially better outcome compared to $S$ acting on its own.

Before elaborating on this definition, let's consider some examples of how this definition applies to different domains:
\begin{itemize}[leftmargin=1cm]
    \item[a.] In the context of Go, a human-robust policy must not lose against opponents where a human could take over temporarily and consistently produce a win. The set of opponents under consideration is $E$.

    None of the defense strategies studied in this work meet this standard when $E$ is the set of opponents that can be trained with a reasonable amount of compute and that have gray-box access to $S$ during both training and inference.

    \item[b.] Image classifiers which are not robust to $\epsilon$-ball perturbations on natural images  are probably not human-robust, since it is suspected that humans are highly invariant to small norm perturbations of natural images.\footnote{Though this is still an claim that is being actively researched~\citep{veerabadran2023subtle}.}
    $E$ here is the set of natural images.
\end{itemize}

Now that we have a couple examples in mind, let us unpack the different pieces of our definition in detail:

\begin{itemize}[leftmargin=1cm]
    \item
    The concept of a ``human having ordinary skill in the art'' is derived from patent law,\footnote{\url{https://en.wikipedia.org/wiki/Person_having_ordinary_skill_in_the_art}} referring to a person with ``normal skills and knowledge'' of a particular field ``without being a genius''. We can define related concepts by specifying a different group of humans, giving rise to notions such as lay- or expert-human-robustness, or amateur- or professional- or world-champion-human-robustness.

    \item
    The decisions for a human to control are chosen by an omniscient observer so that the human does not need hindsight. If the human themselves were choosing when to intervene without hindsight, they would have to anticipate the robustness failures of the system or perform better than the system in all cases. The former would be unrealistic in many cases, while the latter would be defining a standard that cannot apply to systems that are sometimes superhuman. On the other hand, if the human had hindsight, they would be able to choose where to intervene themselves, but they would also have more knowledge of how to intervene (gained e.g. by observing the consequence of a bad action) than a human with ordinary skill in the art would. Thus our omniscient observer criterion sits somewhere in the middle between a human having hindsight and a human not having hindsight.

    \item
    We can also define a stronger form of robustness by removing the ``intentionally'' condition. This would allow for humans to make correct decisions without legitimate reasons for them. For instance, some Go beginners will rush to capture stones even when the opponent already can't save them. This wastes many moves, but potentially blocks an adversary from using those stones later. In situations like these, humans might be more robust, but in an unstable way: when those beginners learn more and understand that their opponent couldn't save those stones, they will no longer rush to capture them, destroying that robustness.

    \item
    We say that a human must ``consistently ... produce [a] better outcome'' because there are plenty of situations where a human could randomly make the right decision. But in these situations, a flip of a coin could make the right decision too. Thus the consistency criterion makes our standard for human-robustness less stringent. The exact strength of our definition can be adjusted by specifying a particular probability that a human should make the right decision.

    \item
    The meaning of ``substantially better outcome'' should depend on the context and level of robustness needed. In the context of Go, the clearest standard is winning or losing the game. This could be made more strict by requiring that human interventions cannot increase the the final score differential by any significant amount (e.g. over 5 points) in addition to being unable to change the final win/loss outcome. We find this stricter standard is currently unnecessary since we find none of the Go AIs we test meet even the weaker standard.

    \item
    Finally, our definition applies to a specific environment $E$ (instead of quantifying over all possible environments) because of the existence of pathological environments that make it so that no non-human system can be human-robust in all possible environments.
    
    For example, imagine a deployment environment where the system $S$ fails due to some near-omnipotent entity $O$ stacking the cards against $S$ (for example in the Go domain this would correspond to $S$ facing an all-around much stronger Go program $O$). However, this near-omnipotent entity $O$ also has a backdoor -- if they detect any sign of human intervention, they will ``un-stack'' the odds in the opposite direction (for example instantly resigning the game in the case of Go). The existence of such a pathological environment implies that no computationally feasible non-human system can be human-robust in all possible environments.
\end{itemize}

We think our definition of human-robustness is a useful lens through which to think about robustness. In particular, if a system is human-robust in all non-pathological environments, that implies it will only fail in ways humans would too, and will not create new vulnerabilities. This is a practically useful desiderata. Moreover, our definition is sufficiently concrete to be falsifiable in real-world scenarios.

In the case of Go, we note that the Go AIs in our experiments do not even achieve the weakest version of our human-robustness criterion, since an amateur human can make correct decisions to defend against our attacks virtually 100\% of the time. Specifically, for all the cyclic attacks in Figure~\ref{fig:boardstates}, the human could simply capture the adversary's group inside the cyclic group, before the cyclic group itself is captured. This is trivial since the inside groups have few liberties and no options to defend against that. Meanwhile, to defend against \koadv{}, a human would simply not offer the gift, for example, connecting at the location marked with $\triangle$ in Figure~\ref{fig:giftexample}. Similarly, to defend against \attackhnine{}, a human just needs to avoid filling in their own last liberty, i.e., playing anywhere else such as one of the numerous captures available, instead of the location marked with $\triangle$ in Figure~\ref{fig:validateex:m2}.

\subsection{Training-Compute Robustness}
\label{app:train-compute-robustness}
Our second definition of robustness captures the property that a robust Go AI should only be beatable by using a large amount of computational resources. This definition naturally extends to more general agents -- \emph{training-compute-robust} agents should only be exploitable (i.e. induced to fail in dramatic ways) by adversaries that have large computational resources.

There are different types of training-compute-robustness for different threat models, e.g. black-box, gray-box, and white-box. We focus on gray-box training-compute-robustness in this work, but also comment briefly on the other types below.

We formalize this for Go as follows. Let $\pi$ be a policy for playing Go, which uses a fixed $I$ amount of inference compute per move. The \textbf{$p$-level training-compute-robustness} of $\pi$ is the minimum amount of compute needed to train an adversarial policy $\pi_\text{adv}$ that can defeat $\pi$ at least a $(1-p)$ fraction of the time while also using $I$ inference-compute or less per move.
We minimize over a set $\mathcal{T}$ of possible training schemes, e.g. different hyperparameters or attack algorithms.

Note that in general we can only upper bound training-compute-robustness, since minimizing over all possible adversarial-policy training schemes in $\mathcal{T}$ is intractable for large or infinite $\mathcal{T}$.
A trivial upper bound on the 50\%-level training-compute-robustness for a policy $\pi$ is the net amount of compute that was used to train $\pi$. This is because a policy can always win against itself with 50\% probability.
In fact, if the attacker has white-box access to $\pi$ then the 50\%-level white-box training-compute-robustness of any policy is \emph{zero}, as the attacker can just set $\pi_\text{adv} = \pi$ without any training.

Given any self-play algorithm $\mathcal{A}$ parameterized by training compute (i.e. $\mathcal{A}$ produces a policy given a fixed amount of compute), we can also use $\mathcal{A}$ to obtain an upper bound on $p$-level training-compute-robustness of any policy $\pi$. Namely, we can measure the minimum amount of compute (possibly infinite) needed by $\mathcal{A}$ to produce a policy that can win against $\pi$ at least a $(1 - p)$ fraction of the time.

Training-compute-robustness can be measured in different units, e.g. FLOPs, V100 GPU days, A100 GPU days, etc. A unit-less way to measure training-compute-robustness is as a fraction of the compute needed to train $\pi$. We call this \emph{relative training-compute-robustness}. The relative 50\%-level training-compute-robustness of a policy $\pi$ is always less than 1.

\subsection{Inference-Compute Robustness}
\label{app:test-compute-robustness}
Our third and final definition of robustness captures the criterion that a robust system should be able to effectively correct its own mistakes given enough time to check its work at inference-time.

We formalize the \textbf{inference-compute-robustness} of a Go policy $\pi$ against an opponent $\pi_\text{adv}$ as the rate at which $\pi$'s win-rate against $\pi_\text{adv}$ increases as a function of $\pi$'s inference-compute. The faster the rate of increase, the more inference-compute-robustness $\pi$ has. Taking the slowest win-rate scaling trend over all $\pi_\text{adv}$ in some threat-model set yields an aggregate measure of inference-compute-robustness.

An upper-bound on inference-compute-robustness can be obtained by pitting a policy against a version of \emph{itself} with a fixed inference budget. Any scaling trend which is significantly slower than this baseline scaling trend is an indication that a policy lacks inference-compute robustness. We use this baseline scaling trend in Figure~\ref{fig:inference-compute-rob} to show that none of our defended victims have strong inference-compute-robustness against our strongest adversaries. The same figure does, however, show that our defenses are fairly inference-compute-robust with respect to some of our adversaries, like \attackhnine{} and \koadv{}, even though these same adversaries demonstrate the defenses' lack of human-robustness and train-compute-robustness.

\begin{figure*}
\centering
\begin{subfigure}{0.48\textwidth}
\input{figs/plots/icr-b18.pgf}
\caption{Positional adversarial training, Dec.\ 2023}
\end{subfigure}
\begin{subfigure}{0.48\textwidth}
\input{figs/plots/icr-b18-s9997m.pgf}
\caption{Positional adversarial training, May 2024}
\end{subfigure}
\par\bigskip %
\begin{subfigure}{0.48\textwidth}
\input{figs/plots/icr-h9.pgf}
\caption{Iterated adversarial training}
\end{subfigure}
\begin{subfigure}{0.48\textwidth}
\input{figs/plots/icr-vit.pgf}
\caption{Vision transformer (ViT)}
\end{subfigure}

\caption{A version of Figure~\ref{fig:vs-visits-combined} with additional baseline scaling trends for the \defender{} vs. itself. 
In the legends, \texttt{<model>-v} is an abbreviation of \texttt{<model>-victim} to save space.
We note that the strongest \attackers{} stay above the baseline curve (i.e. the \defender{} must spend more inference compute to beat them than it needs to beat itself). This is an indication that none of our defended \defenders{} have strong inference-compute-robustness.}
\label{fig:inference-compute-rob}
\end{figure*}

\section{Training Parameters}
\label{app:training-background}

\subsection{Training Window}\label{app:training-window}

KataGo generates training data from self-play games.
The model then trains on a sample from a sliding window of the most recent training data.
The default starting window size is $m_0 = 250,000$ samples or ``rows,'' and when there are $N$ total training rows, the window size $m$
scales as a power law in $N$:\footnote{The sliding window is implemented by the script at \url{https://github.com/lightvector/KataGo/blob/eaaddd82339750d9defc70f566e6c59d7068b7b3/python/shuffle.py}, and the \texttt{-{}-help} documentation string for the script gives this equation.}
\begin{equation}\label{eq:shuffle-window-size}
  m = \frac{.4m_{0}^{.35}}{.65} \cdot \left(N^{.65} - m_{0}^{.65}\right) + m_0
.\end{equation}
Each training ``epoch'' consumes approximately 250,000 data rows and performs 1 million training steps.

All of our models, besides our self-play ViT models, involve \emph{warm-starting} from KataGo models or models trained by \Citet{wang2023adversarial}.
Warm-starting from a model, or fine-tuning a model, means we initialize our training from that model and pre-seed the training data with that model's training history.
The pre-seeding increases the window size by increasing $N$ in Equation \ref{eq:shuffle-window-size}, and it populates the training window with the pre-existing training data.
Without pre-seeding the data, the default starting window size would be small and cause over-fitting.
We could also increase the training window size by increasing $m$ without pre-seeding, but then there is a high initial cost to generate enough new data to populate the starting window.

\subsection{Configuration Parameters}\label{app:config}

\begin{table*}[ht]
    \centering
    \begin{tabular}{llllllllllllll}
        \toprule
         \textbf{Board size ($n \times n$)} & 7 & 8 & 9 & 10 & 11 & 12 & 13 & 14 & 15 & 16 & 17 & 18 & 19 \\
         \textbf{Training frequency (\%)} & 0.7 & 0.7 & 2.9 & 1.4 & 2.1 & 2.9 & 7.1 & 4.2 & 5 & 5.7 & 6.4 & 7.1 & 53.6 \\
         \bottomrule
    \end{tabular}
    \caption{Percentage of games played at each board size throughout the
    training of our adversaries. These percentages are the values used for recent KataGo training.}
    \label{tab:app:board-size-freq}
\end{table*}

The board size varies randomly between training games, allowing KataGo to learn to play Go on various board sizes.
Because we focus on 19\texttimes19 games in our evaluations, we train our \attackers{}
primarily on 19\texttimes19 games: 75/140 = 53.6\% to be precise, matching the distribution used in recent KataGo training.
Table~\ref{tab:app:board-size-freq} shows the full distribution of board sizes.
This contrasts with the attack of \citet{wang2023adversarial} who set only 35\% of games to be 19\texttimes19, following a distribution of board sizes matching those used for early KataGo training of small 6-block and 10-block models.

When training our adversaries, we disable the variance time loss (\texttt{vtimeloss} in the KataGo code),
an auxiliary loss on a model output predicting uncertainty in the game's outcome.
We disabled it following \citeauthor{wang2023adversarial}'s finding that this stabilized their early training runs, although we did not confirm its impact on our training.

Like \citeauthor{wang2023adversarial}, our adversary training uses
curricula in which the adversary plays against increasingly strong victims, switching to
a stronger victim once the adversary win rate exceeds a certain threshold.
We usually set the threshold to 75\%.
However, we increased the threshold to 90\% for higher visit count victims (typically 512 or more) since at that point higher victim visit counts substantially increase the cost of generating games, making it more computationally efficient to train at a slightly lower sample efficiency but with cheaper samples.

When training adversaries against a victim using fewer than 100 victim visits, we enable
\citeauthor{wang2023adversarial}'s pass-alive defense to prevent the adversary from learning
the degenerate ``pass attack'' that they encountered in low-visit victims.

We change several training configuration parameters listed below compared to \citeauthor{wang2023adversarial}, usually tweaking these parameters partway into training runs since we only identified or began experimenting with them after launching the runs.

\paragraph{Enabling selecting moves by the lower-confidence bound (LCB) on their utility.}
   Selecting moves by LCB is the default in evaluation but is disabled in training
   because the creator of KataGo found that
   enabling LCB reduced self-play training progress despite making evaluation stronger.\footnote{The creator of KataGo details their LCB experiments at \url{https://github.com/leela-zero/leela-zero/issues/2411}.}.
   We found that having LCB disabled led to a large strength gap between training and evaluation.
   We preferred to keep train and evaluation similar so that we could be more confident that training progress correlated with evaluation strength.

\paragraph{Adjusting other victim configuration parameters to more closely match the settings used during evaluation.} For example, parameters that govern exploration vs. exploration trade-offs (like temperature), or the KataGo ``optimism'' feature\footnote{Policy optimism is described at \url{https://github.com/lightvector/KataGo/blob/828f1bc27617f9a7dc881d11a7296856ef7c4fc0/docs/KataGoMethods.md\#optimistic-policy}. \Citeauthor{wang2023adversarial} used a version of KataGo that had not yet introduced this feature.}.

  In some training runs, we only changed a subset of these parameters because we had not yet discovered all of these parameters disparities. The full list of parameters we change in the final runs is:
  \begin{verbatim}
    antiMirror = true
    chosenMoveTemperature = 0.10
    chosenMoveTemperatureEarly = 0.50
    conservativePass = true
    cpuctExploration = 1.0
    cpuctExplorationLog = 0.45
    cpuctUtilityStdevScale = 0.85
    dynamicScoreCenterScale = 0.75
    dynamicScoreCenterZeroWeight = 0.2
    dynamicScoreUtilityFactor = 0.3
    enablePassingHacks = true
    fillDameBeforePass = true
    policyOptimism = 1.0
    rootDesiredPerChildVisitsCoeff = 0
    rootFpuReductionMax = 0.1
    rootNoiseEnabled = false
    rootNumSymmetriesToSample = 1
    rootPolicyOptimism = 0.2
    rootPolicyTemperature = 1.0
    rootPolicyTemperatureEarly = 1.0
    staticScoreUtilityFactor = 0.1
    subtreeValueBiasFactor = 0.45
    subtreeValueBiasWeightExponent = 0.85
    useNoisePruning = true
    useNonBuggyLcb = true
    useUncertainty = true
    valueWeightExponent = 0.25
  \end{verbatim}

\paragraph{Adjust adversary configuration parameters to more closely match the settings used in the latest KataGo training runs.} This involves slight adjustments in
  exploration and utility computation, as well as a small bugfix related to LCB.
  We made these changes under the assumption that later KataGo configurations are superior to early ones our initial parameter settings were based on,
  although we did not check if this made a major difference in our training. The full list of parameters we change is:
  \begin{verbatim}
    cpuctExploration = 1.05
    cpuctExplorationLog = 0.28
    dynamicScoreCenterScale = 0.75
    dynamicScoreUtilityFactor = 0.30
    rootPolicyTemperatureEarly = 1.5
    staticScoreUtilityFactor = 0.05
    subtreeValueBiasFactor = 0.30
    useNonBuggyLcb = true
  \end{verbatim}

\section{Positional Adversarial Training}

\subsection{Gift Attack}\label{app:giftattack}

This section gives more details on the \koadv{} attacking \dectwentythree{} described in Section~\ref{sec:attack-b18}.

The adversary is warm-started from \origtwotwoseven{}.
The curriculum began with \dectwentythree{} at 4 visits, increasing up to 8 visits in 1 visit increments, then doubling visits each time until 512 visits. We added the extra victim visits between 4 and 8 because after finding that a direct increase from 4 to 8 visits led to a large win rate drop and minimal
training progress.
The adversary was trained for a further 1697 V100 GPU-days and 651 million training steps, totalling 1861 GPU-days and 878 million steps (Table~\ref{tab:allattackers}).

\begin{figure*}[btp]
\centering
\input{figs/plots/win-rate-vs-gpu-days-gift.pgf}
\caption{Win rate (\%) of \koadv{} (marked \plotdiamond{}) against \dectwentythree{} throughout fine-tuning against \dectwentythree{}. The zero of the x-axis represents the win rate of \origtwotwoseven{} against \dectwentythree{} before the fine-tuning against \dectwentythree{} began.
The large drop in win rate at 700 GPU-days occurred when the curriculum prematurely increased from 128 visits to 256 visits. The adversary's win rate against \dectwentythree{} at 256 visits was poor, and it was not learning well. After we reverted the curriculum back to 128 visits, the win rate surprisingly recovered, seemingly without hindering training progress.
}
\label{fig:vs-gpu-days-gift}
\end{figure*}
\begin{figure*}[btp]
\centering
\input{figs/plots/vs-victim-checkpoints-gift.pgf}
\caption{The win rate (\%) of \koadv{} against the main KataGo training run between networks
\texttt{b18-s4975m} and \maytwentyfour{}.
The marked point \plotdiamond{} is \dectwentythree{}.
At the dashed line, KataGo's developer added positions from \contadv{} and \koadv{} into KataGo's adversarial training data.}
\label{fig:vs-victim-checkpoints-gift}
\end{figure*}

Figure~\ref{fig:vs-gpu-days-gift} shows the win rate of \koadv{} throughout training.
Figure~\ref{fig:vs-victim-checkpoints-gift} shows \koadv{}'s win rate against several \texttt{b18} KataGo networks.
Either \koadv{} is highly specialized to setting up the gift attack against \dectwentythree{}, or the gift vulnerability only appears in recent KataGo \texttt{b18} nets.
KataGo's developer suggests the former is more likely. After we disclosed this vulnerability, he examined older KataGo nets and found that they also misjudge board positions produced by \koadv{}.

At 163 V100 GPU-days (79 million training steps), we adjusted victim configuration parameters to more closely match evaluation
as described in Appendix~\ref{app:config}.

At 170 V100 GPU-days (81 million training steps), we reduced the training move limit per game from KataGo's default of 1600 moves to $900 * (\textrm{board area})/(19^2)$ moves since we noticed several games dragging out to hit the move limit due us enabling the pass-alive defense (Appendix~\ref{app:config}), which lengthens games, on low-visit victims during training. This is before the adversary had discovered the gift attack, and games were not noticeably longer than \attackhnine{}'s games at a similar point in \attackhnine{}'s training.
Still, we hypothesized this would increase training efficiency by cutting the duration of lengthy games, which cost compute and generate an excessive amount of end-game policy training data.

At 475 V100 GPU-days (220 million steps), we noticed that the adversary learned to prolong a significant portion of games using extended ko fights to hit the 900-move limit.
Normally during training, such games are scored and assigned a winner based on the final board state.
Not only does having lots of games hit the move limit significantly slow down training, but we were also worried that the final board state score does not necessarily reflect what the score would have been had the game been played out to completion. The adversary could reward hack by stalling in a state that is winning if scored prematurely but is losing if played to the end.

We therefore at 632 V100 GPU-days (307 million steps) began scoring games that hit the move limit with a score of 0 and marked them as losing for the adversary (-1 utility).
This drove the hit-move-limit rate from 59\% to 22\%.
At 836 V100 GPU-days (393 million steps), we reduced the utility of such games even further to -1.6, below the worst typically possible utility -1.35 resulting from losing a game and having the opponent control all territory on the board.
This drove the hit-move-limit down to 0\%.

\subsection{Continuous Attack}\label{app:continuous}

This section gives more details on the \contadv{} attacking \dectwentythree{} briefly mentioned in Section~\ref{sec:internaladvtraining}.

We warm-started \contadv{} from \attackbsixty{} since \citeauthor{wang2023adversarial} found \attackbsixty{} to be effective against KataGo's \texttt{b18} networks (despite it being trained against \texttt{b60} network \bsixty{}).
We trained the adversary for a further 1098 V100 GPU-days and 630 million training steps, for a total of 4476 GPU days and 1343 million training steps (Table~\ref{tab:allattackers}).

We started the curriculum at 1 victim visit against a KataGo model released August 2023, and we doubled the visits when a win rate threshold was reached. The threshold was 75\% up to 256 visits and then switched to 90\%.
We periodically (or ``continuously'', hence the name \contadv{}) updated the KataGo \texttt{b18} checkpoint used to the latest KataGo checkpoint over several months.

\begin{figure}[btp]
    \centering
    \includesvg[width=0.4\textwidth]{figs/boards/cont_example-2col.svg}
    \caption{An example of the cyclic shape formed by \contadv{} (black) in a game
    against \dectwentythree{} (white) at 65,536 victim visits.
    \ifarxiv
      \href{\demosite/positional-adversarial-training?row=0\#dec23-vs-continuous-board}{See the full game on our website}.
    \fi
    }
    \label{fig:contadv_boardstate}
\end{figure}

The final \contadv{} achieves a win rate of 91\% against \dectwentythree{} and 65\% against 4096 visits (Figure~\ref{fig:vs-visits-combined}).
Although still cyclic, unlike \citeauthor{wang2023adversarial}'s original cyclic attack the \contadv{} always forms nearly the same shape in the interior of the cycle (Figure~\ref{fig:contadv_boardstate} has an example, and the shape is also visible in Figure~\ref{fig:cyclic-heatmap-continuous}). We suspect that adversarial training shrunk \dectwentythree{}'s attack surface, which incentivizes adversaries to adopt narrower attacks in order to consistently land their exploit.

\begin{figure*}[btp]
\centering
\input{figs/plots/win-rate-vs-gpu-days-continuous.pgf}
\caption{Win rate (\%) of \contadv{} (marked \plotdiamond{}) against \dectwentythree{} throughout fine-tuning against \dectwentythree{}. The zero of the $x$-axis represents the win rate of \attackbsixty{} against \dectwentythree{} before the fine-tuning against \dectwentythree{} began.}
\label{fig:vs-gpu-days-continuous}
\end{figure*}

\begin{figure*}[btp]
\centering
\input{figs/plots/vs-victim-checkpoints-cont.pgf}
\caption{The win rate (\%) of \contadv{} against the main KataGo training run between networks \texttt{b18-s4975m} and \maytwentyfour{}.
(\texttt{b18-s4975m} is the first KataGo model with the same architecture as \dectwentythree{} and \maytwentyfour{}. It was released in March 2023.)
The marked point \plotdiamond{} is \dectwentythree{}.
At the dashed line, the KataGo developers added positions from \contadv{} and \koadv{} into KataGo's adversarial training data, which caused the win rate to drop.}
\label{fig:vs-victim-checkpoints-continuous}
\end{figure*}

Figure~\ref{fig:vs-gpu-days-continuous} shows \contadv{}'s win rate against \dectwentythree{} throughout \attacker{} training.
Figure~\ref{fig:vs-victim-checkpoints-continuous} shows \contadv{}'s win rate against several \texttt{b18} KataGo networks.
We see that \contadv{} successfully attacks all \texttt{b18} networks up until \texttt{b18-s9432m} when \contadv{} positions were introduced into KataGo's training data, at which point \contadv{}'s win rate declines substantially.
This decline was faster than when KataGo initially introduced positions from \origcyclic{} into KataGo's training data---at that time, it took several hundred million training steps to make \origcyclic{}'s win rate to dramatically drop, see \citet[Figure~L.2]{wang2023adversarial}.

\contadv{}'s full curriculum was:
\begin{itemize}
    \item \texttt{b18-s7283m} (released August 17, 2023), 1--16 visits.
    \item \texttt{b18-s7313m}, 16--32 visits.
    \item \texttt{b18-s7343m}, 32--256 visits.
    \item \texttt{b18-s7373m}, 256 visits.
    \item \texttt{b18-s7500m}, 256--512 visits.
    \item \texttt{b18-s7590m}, 1024 visits.
    \item \texttt{b18-s7620m}, 256 visits. (Here we reverted visits to 256 because earlier visit increases were due to non-representative samples of games skewing our curriculum advancement script into giving inaccurate win rate estimates.)
    \item \texttt{b18-s7680m}, 256 visits.
    \item \texttt{b18-s7740m}, 256 visits.
    \item \texttt{b18-s7830m}, 256 visits.
    \item \texttt{b18-s7890m}, 256 visits.
    \item \texttt{b18-s7950m}, 256 visits.
    \item \texttt{b18-s8010m}, 256 visits.
    \item \texttt{b18-s8071m}, 256--512 visits.
    \item \texttt{b18-s8191m}, 512 visits.
    \item \texttt{b18-s8282m}, 512 visits.
    \item \texttt{b18-s8463m} (released Dec 11 2023), 512 visits.
    This is the last curriculum checkpoint that our chosen adversary checkpoint \contadv{} at 1098 V100 GPU-days saw.
    The remaining curriculum checkpoints were seen by adversary checkpoints beyond the one we chose for main evaluations in this paper.
    \item \texttt{b18-s8588m}-v512
    \item \texttt{b18-s8678m}-v512 (released Jan 9 2024)
\end{itemize}

We initially observed a large win rate gap between training and evaluation.
To close this gap, we made only two changes to the training configuration, rather than all the changes listed in Appendix~\ref{app:config}: we enabled LCB move selection and activated optimism for the victim.

\subsection{Big Cyclic Attack}\label{app:largeadv}

This section gives more details on the \largeadv{} attacking \maytwentyfour{} described in Section~\ref{sec:attack-b18}.

\begin{figure}[btp]
\centering
\input{figs/plots/vs-visits-b18-s9997m-cont-ko.pgf}
\caption{Win rate (\%) of \contadv{} and \koadv{} against \maytwentyfour{} with varying victim visits.}
\label{fig:vs-visits-b18-s9997m-old-advs}
\end{figure}

We wanted to see whether the cyclic vulnerability was still easy to find after KataGo added
positions from \contadv{} and \koadv{} into its adversarial training data. We chose \maytwentyfour{}
as our target because as of June 2024 it was the last and strongest KataGo network with the same \texttt{b18c384} architecture as \dectwentythree{}. (In May 2024, KataGo also began releasing stronger KataGo networks with a larger \texttt{b28c512} architecture.
Running these larger networks would have required us to expend some engineering effort merging a new version of KataGo into our custom fork of KataGo that supports A-MCTS, victim-play, and ViTs.)
Figure~\ref{fig:vs-visits-b18-s9997m-old-advs} shows that \maytwentyfour{}'s adversarial training was successful in
mitigating the exploits of \contadv{} and \koadv{}, with \contadv{}'s win rate dropping from 65\% to 3\% at 4096 victim visits and \koadv{}'s win rate dropping from 91\% to 0\% (over 200 games) at 512 victim visits.

\begin{figure*}[btp]
\centering
\input{figs/plots/win-rate-vs-gpu-days-large.pgf}
\caption{Win rate (\%) of \largeadv{} (marked \plotdiamond{}) against \maytwentyfour{} throughout fine-tuning against \maytwentyfour{}. The zero of the $x$-axis represents the win rate of \attackbsixty{} against \maytwentyfour{} before the fine-tuning against \maytwentyfour{} began.}
\label{fig:vs-gpu-days-large}
\end{figure*}

We fine-tuned an adversary \largeadv{} to attack \maytwentyfour{}. Like \contadv{}, we initialized \largeadv{} to \attackbsixty{}. The training curriculum started with \maytwentyfour{} with 16 victim visits, and we doubled victim visits up until 1,024 visits. The curriculum win rate threshold was 75\% until 512 visits, then we increased it to 90\%. We used the updated victim and adversary configuration settings described in Appendix~\ref{app:config}. Training lasted for 293 V100 GPU-days and 213 million training steps.

\begin{figure*}[btp]
\centering
\input{figs/plots/vs-victim-checkpoints-large.pgf}
\caption{The win rate (\%) of \largeadv{} against the main KataGo training run between networks \texttt{b18-s4975m} and \maytwentyfour{}.
The marked point \plotdiamond{} is \maytwentyfour{}.}
\label{fig:vs-victim-checkpoints-large}
\end{figure*}

\begin{figure}[btp]
\centering
\input{figs/plots/vs-visits-b18-s8527m-large.pgf}
\caption{Win rate (\%) of \largeadv{} against \dectwentythree{} with varying victim visits.}
\label{fig:vs-visits-b18-s8527m-large}
\end{figure}

Figure~\ref{fig:vs-gpu-days-large} shows the win rate of \largeadv{} throughout training. The win rate of \largeadv{} had not plateaued, but we stopped the run early anyway because the curriculum reached high victim visit counts that become substantially more expensive to train against.
In Figure~\ref{fig:vs-victim-checkpoints-large} we see that \largeadv{} is also effective at beating old \texttt{b18} checkpoints. This is further confirmed in a human replication of the \largeadv{} attack against \dectwentythree{}.
In fact, Figure~\ref{fig:vs-visits-b18-s8527m-large} shows that \largeadv{} is stronger than \contadv{} against \dectwentythree{} at high victim visits, achieving a 35\% win rate at 65,536 victim visits compared to \contadv{}'s 27\%.
\largeadv{} was also trained for less time than \contadv{} (293 vs.\ 1098 V100 GPU-days of fine-tuning).
Our guess is that \contadv{}'s training progress was slower due to \contadv{}'s lengthy curriculum and due to by chance finding a narrow cyclic attack structure that is difficult to find local improvements for, not to due to \dectwentythree{} inherently being harder to exploit than \maytwentyfour{}.

\section{Iterated Adversarial Training}
\label{app:adversarialtraining}

\subsection{Defense}

At each iteration, we train a \defender{} \defenseiter{n} to defend against a fixed \attacker{} \attackiter{n-1}.
Figure~\ref{fig:vs-gpu-days-h} shows the training progress of each \defenseiter{n} against \attackiter{n-1}.
Figure~\ref{fig:vs-gpu-days-h-each} shows the same information but with a separate plot for each iteration.
We see that the \defender{} always made rapid progress in defending against the \attacker{}, but continued to lose a significant fraction of the time at 16 victim visits, and still suffered occasional losses at 256 victim visits.

\begin{figure*}[btp]
\centering
\input{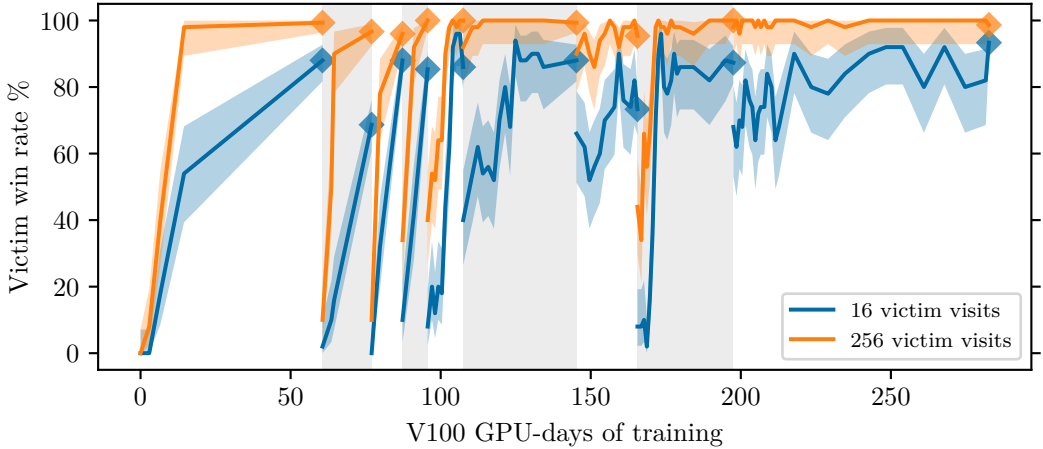}
\caption{The victim \defenseiter{n} win rate (\%) against \attackiter{n-1} throughout iterated adversarial training. Iterations are signified by alternating between a white and gray background.
The curves for \defenseiter{1} to \defenseiter{4} only have a few data points along the x-axis as intermediate checkpoints were lost.
}
\label{fig:vs-gpu-days-h}
\end{figure*}

\subsubsection{Training Settings}

The first \defenseiter{1} is warm-started from \defenseiter{0} (\cpfivezerofive{}) and is trained against \origcyclic{}.
We do not use a curriculum in \defender{} training.
We reduced the learning rate by a factor of 10 from KataGo's default since the base model \cpfivezerofive{} had been trained with a lower learning rate scale as well.
We found that fine-tuning with the default learning rate led to a large initial drop in model strength.

The \defender{} plays with 300 MCTS visits, while the \attacker{} plays with 600 A-MCTS visits.
We chose 600 A-MCTS visits for the \attacker{} to follow the default number of visits for adversary training and evaluation used by \citet{wang2023adversarial}.
We chose 300 MCTS visits for the \defender{} because it keeps the inference cost of the \defender{} similar to the \attacker{}'s---roughly $600/2=300$ visits of the adversary's A-MCTS invoke the \defender{} model, with the remaining visits invoking the smaller, cheaper \attacker{} model.

The training window size begins at 68 million rows to match the window size of \cpfivezerofive{}.\footnote{The window size was calculated from Equation~\ref{eq:shuffle-window-size} using the fact that \cpfivezerofive{} was trained on 2.9 billion rows of data.}
This is large enough that throughout defense training, all games generated in prior defense iterations remain in the training window.
Although keeping all the games in the window was not an intentional design choice, it likely contributes to each \defenseiter{n} defending well against every \attackiter{m} with $m < n$.

The \defender{} was trained with a mixture of self-play games and games against the adversary.
Self-play games help preserve general Go strength,
whereas games against the adversary focus on overcoming specific attacks.
We set the game mix to 82\% self-play and 18\% against the adversary.
This proportion was based on preliminary experiments suggesting that
training on 90\% self-play data and 10\% adversary data makes rapid progress in overcoming the adversary
without compromising general Go strength (estimated via win rate against \cpfivezerofive{}).\footnote{
For the self-play data, we downloaded \cpfivezerofive{}'s most recent self-play data from \url{https://katagotraining.org/}. The adversary data came from two experimental runs not included in this paper where we fine-tuned \cpfivezerofive{} with victim-play against \origcyclic{} (i.e.\ 100\% training games against \origcyclic{} and no self-play games) with A-MCTS and MCTS. The two runs generated 80 million training steps worth of adversary data in total.
We then performed supervised fine-tuning of \cpfivezerofive{} against a mix of these data for 25 million training steps with self-play data proportions of 25\%, 50\%, 75\%, 80\%, 85\%, 90\%, 92.5\%, 95\%, 97.5\%, and 99\%. Proportions of at most 80\% often gave large decreases in win rate against a frozen copy of \cpfivezerofive{}, and proportions of at least 95\% gave slower win rate increases against \origcyclic{}, so we picked 90\% as our final self-play proportion.
After picking the self-play proportion we also tried changing the learning rate scale-down from 10\texttimes\ to 2.5\texttimes, 5\texttimes, or 20\texttimes, but 5\texttimes\ was too large and gave worse training progress whereas 20\texttimes\ was not noticeably better than 10\texttimes.
}
Self-play games generate twice as much policy training data as games against the adversary because the model only trains on its own moves in adversary games.
Setting the proportion of selfplay games to 82\% makes the generated game data roughly match 90\% from self-play.

For simplicity, we only used Tromp-Taylor rules, since the adversaries were also only trained on these rules.
We also disabled many KataGo self-play flags (auto-komi, komi randomization, handicap games, game forking, cheap search, reduced search when winning, playing initial moves directly from policy) to simplify implementation.

In each iteration, we hand-select the final model for the subsequent iteration based on expected strength.
All else equal, we choose the checkpoint with the highest win rate against the adversary.
However, as the win rate against the adversary tends to plateau, we additionally favor checkpoints from \emph{stable} periods of training where immediately preceding and succeeding checkpoints also have high win rates.
We break ties in favor of earlier checkpoints.

\subsubsection{Defense Per-Iteration}

In this section we discuss each individual iteration in more detail.
We provide the training cost (in training steps and V100 GPU-days) of each iteration in Table~\ref{tab:iterated_training_cost}.
Additionally, we discuss any configuration changes or notable results that occurred in iterations below.

\begin{table*}[btp]
\centering
\begin{tabular}{lllll}
\toprule %
\textbf{Iteration} & \multicolumn{2}{c}{\textbf{Victim}} & \multicolumn{2}{c}{\textbf{Adversary}} \\
\textbf{$\boldsymbol{n}$} & \textbf{GPU-days} & \textbf{Steps (M)} & \textbf{GPU-days} & \textbf{Steps (M)} \\

\midrule %

1 & 61 & 61 & 238 & 150  \\

2 & 16 & 22 & 439 & 253 \\

3 & 10 & 16 & 273 & 213 \\

4 & 8 & 11 & 1195 & 983 \\

5 & 12 & 10 & 862 & 535 \\

6 & 38 & 20 & 304 & 228 \\

7 & 20 & 13 & 491 & 372 \\

8 & 32 & 32 & 308 & 230 \\

9 & 85 & 71 & 1005 & 622 \\

\textbf{Total} & \textbf{282} & \textbf{256} & \textbf{5114} & \textbf{3587} \\

\bottomrule %
\end{tabular}
\vspace{1em}
\caption{The cost of training the \defender{} \defenseiter{n} and \attacker{} \attackiter{n} at each iteration $n$ of iterated adversarial training.} %
\label{tab:iterated_training_cost} %
\end{table*}

\paragraph{Defense Iteration 1:}
The win rate at 300 victim visits against the cyclic adversary already plateaued after 14 of the 61 GPU-days (16 of 61 million steps), but we continued training in hopes of achieving a consistent 100\% win rate against the adversary.

\paragraph{Defense Iteration 4:}
An error occurred in populating the training history, where extra data from running the previous defense iteration was added for an additional 58 million steps beyond our selected checkpoint \defenseiter{3}.
We identified this error and removed the extraneous data for subsequent defense rounds.

\paragraph{Defense Iteration 6:}
In iteration 6 and 7 we unintentionally generated games
faster than we were training on them, which is why the GPU-days
relative to the number of training steps is higher.

\paragraph{Defense Iteration 9:}
We ran this iteration longer than usual because it was our final defense iteration. We also noticed that its win rate at 8 visits increased modestly (from 49\% to 74\% at the end of training), even though the training win rate at 300 visits against \attackiter{8} remained around 97\% for the entire run.

\subsection{Attack}\label{app:iterated-attack}

\begin{figure*}[btp]
\centering
\input{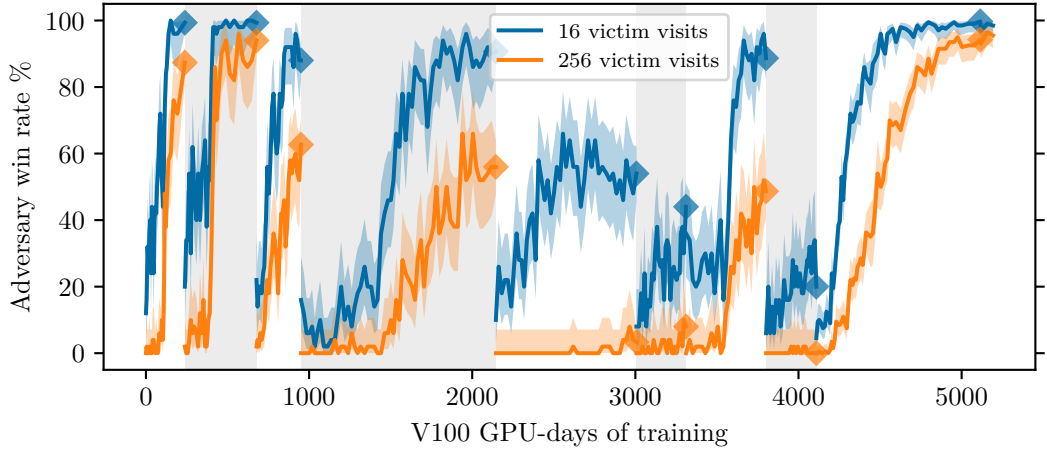}
\caption{The \attacker{} \attackiter{n} win rate (\%) against \defenseiter{n} throughout iterated adversarial training. Iterations are signified by alternating between a white and gray background.}
\label{fig:vs-gpu-days-r}
\end{figure*}

At each iteration, we train an \attacker{} \attackiter{n} to attack \defenseiter{n} warm-starting from the previous \attacker{} \attackiter{n-1}.
The very first iteration \attackiter{1} is warm-started from \attackiter{0} = \origcyclic{}.
Figure~\ref{fig:vs-gpu-days-r} shows the training progress of each \attackiter{n} against \defenseiter{n}.
Figure~\ref{fig:vs-gpu-days-r-each} shows the same information but with a separate plot for each iteration.

As can be seen in Figure~\ref{fig:win-rate-heatmap-iterated-adv}, the \attackers{} at iterations 5, 6 and 8 perform especially poorly for \defender{} visits of 16 or above.
For these iterations, Figure~\ref{fig:vs-gpu-days-r} shows that training progress significantly slowed down; the number of victim visits reached by each iteration from \attackiter{5} onwards was at most 64.
Additionally, in iterations 6 and 8, those \attackers{} were trained for relatively brief periods.
The computational expense of training potent attacks is a bottleneck to performing a large number of iterations of adversarial training.

Does this mean that \defenseiter{5}, \defenseiter{6}, and \defenseiter{8} are
more robust than the final victim \defenseiter{9} that we evaluated in
Section~\ref{sec:advtraining_results}? Our guess is
no. Figure~\ref{fig:win-rate-heatmap-iterated-adv} shows that whenever an
adversary is able to attack a victim \defenseiter{n}, that adversary also
successfully attacks \defenseiter{i} for all $i < n$. In particular,
\attackiter{9} successfully attacks every \defenseiter{n} even at 4096
\defender{} visits. This is suggestive evidence that \defenseiter{5},
\defenseiter{6}, and \defenseiter{8} are not more robust than \defenseiter{9} in terms of how many successful
attacks exist. Perhaps there is an increase in robustness in terms of
training time needed to find a successful attack, since
Figure~\ref{fig:vs-gpu-days-r} shows the adversary win rate increasing faster in
iteration 9 than in iterations 5, 6, and 8. However, it is equally
plausible that our training process has high path dependence and that
\attackiter{5}, \attackiter{6}, and \attackiter{8} got stuck in local minima
refining the first attack they stumbled upon that somewhat worked.
\attackhnine{}, initialized from \origtwotwoseven{}, demonstrates such path
dependence: it maintains qualitative similarities with \origtwotwoseven{}
(compare \attackhnine{}'s attack in Figure~\ref{fig:validateex} against Figure
J.4. in \citet{wang2023adversarial}) and plateaued in training
(Figure~\ref{fig:vs-gpu-days-attack-h9}) even though more potent attacks such as
\attackiter{9}'s attack exist. Overall it is reasonable for us to assume
that \defenseiter{9} is the most robust \defenseiter{n} we trained, though it is
not a certainty---it is merely the final \defender{} we trained before we
exhausted our compute budget.

\subsubsection{Training Settings}

The \attacker{} plays with 600 A-MCTS visits.
Initially the curriculum for each \attacker{} \attackiter{n} consisted of intermediate checkpoints from \defenseiter{n}'s training run before advancing to \defenseiter{n} with doubling visit counts.
We manually selected intermediate checkpoints by looking at the win rates of \defenseiter{n}'s intermediate checkpoints against \attackiter{n-1} and sampling checkpoints with varied win rates.
Later we found that the \defender{} \defenseiter{n} at 1 visit was always vulnerable to attack, so we simplified the curriculum by no longer using intermediate checkpoints
and instead started the curriculum at the final \defenseiter{n} checkpoint with very low visit counts.

The final \attackiter{n} model we select from a training run is always the latest model checkpoint since win rate increases fairly consistently with more training.
We did not have a consistent stopping criterion for each iteration, but we generally targeted either a high win rate at a particular number of visits or restricted the run to a rough training step budget.
Unlike in defense training, because the \attacker{} training is longer and the training window is smaller, the data from the previous iteration fully exits the training window (within 132 million training steps) in every iteration.

\subsubsection{Attack Strategies}
\label{app:iterated-qualitative}

All of the \attackers{} \attackiter{n} exploit a cyclic group, but there are still some qualitative differences. In particular, \attackiter{1} emphasizes a small alive group inside the victim's cyclic group and coaxes the victim to form a cyclic group with an eye, in contrast to the original cyclic attack of \Citet{wang2023adversarial}. \attackiter{2} creates a very large group inside the victim's cyclic group. In the middle iterations \attackiter{4} to \attackiter{6}, the inside group is small. In most attacks, the adversary sets up the inside group early and allows the victim stake out territory around it. However, in \attackiter{4} to \attackiter{6}, the adversary instead stakes out its own territory with the destined inside group on the edge. It then allows the victim come into its territory, resulting in it separating off the inside group and forming the cycle.

Meanwhile, in iterations 7 through 8, the \attacker{} forms an inside group with kos. For \attackiter{7} and \attackiter{8} there are between 1 and 3 kos -- in small sample analysis, there were most often 2 or 3 kos for \attackiter{7} and 1 for \attackiter{8}, with more variable inside group shape. With \attackiter{9}, it initially converged to 2 kos and a highly consistent inside group shape, but then abandoned the kos and started making a diamond, ``ponnuki''-like inside shape, which the victim surrounds with a square shape. This results in a small, nearly minimal inside group at the time of the final capture. An example of this is shown in Figure~\ref{fig:a9example}.

In Appendix~\ref{app:cyclic-heatmaps}, we plot heatmaps of the inside and cyclic group locations. Paralleling the qualitative analysis above, we observe differences in where they are concentrated and their sizes. We also notice some variations in victim stone concentration. Overall, we find clear but constrained evolution in the attacks. To humans, the differences do not change the difficulty of gameplay---the attacks all fit very well in the same overall type (cyclic attacks) so knowing how to beat one would almost certainly mean knowing how to beat them all. But to the KataGo \defenders{}, the representations learned do not appear to generalize smoothly between these variations.

\subsubsection{Attack Per-Iteration}

In this section we discuss each individual iteration in more detail.
We provide the training cost (in training steps and V100 GPU-days) of each iteration in Table~\ref{tab:iterated_training_cost}.
Additionally, we discuss any configuration changes or notable results that occurred in the iterations below.
We denote an intermediate checkpoint $S$ million training steps into the $n$-th iteration of defense training as \defenseiter{n}\texttt{-sSm}.

\paragraph{Attack Iteration 1:}
The curriculum consisted of \defenseiter{1}\texttt{-s4m} with 128 visits,
\defenseiter{1}\texttt{-s16m} with 32--128 visits, and \defenseiter{1} with 32--1024 visits, with a win rate threshold of 75\%.
We stopped the run due to hitting a large number of victim visits, which slowed the generation of training data.

In this iteration, we made an error when warm-starting from the original cyclic adversary.
When we copied the original cyclic adversary's training history, timestamps were erased.
Therefore, at the start of the run, the training window contained the original cyclic adversary's
After 122  of 238 V100 GPU-days (41 of 150 million training steps), all these data left the window.
The most likely effect of this error was hindering early training progress, though it is also possible that
it inadvertently helped encourage exploration in early training.

\paragraph{Attack Iteration 2:}
The curriculum consisted of
\defenseiter{2}\texttt{-s4m} with 128 visits,
\defenseiter{2}\texttt{-s5m} with 64--128 visits, and
\defenseiter{2} with 64--1024 visits, with a win rate threshold of 75\%.
Once again we stopped the run due to hitting a large number of victim visits.

\paragraph{Attack Iteration 3:}
The curriculum consisted of
\defenseiter{3}\texttt{-s5m} with 128 visits and
\defenseiter{3} with 64--256 visits.
Here, we stopped at a 75\% training win rate against 256 victim visits
because we had used about as much compute as in previous attack iterations
and considered 256 visits to be a sufficiently large number of visits that
the attack likely transfers, at least somewhat, to high visits.

\paragraph{Attack Iteration 4:}
The curriculum consisted of
\defenseiter{4}\texttt{-s5m} with 128 visits and
\defenseiter{4} with 4--256 visits.
We aimed for a training win rate of 75\% against \defenseiter{4} at 256 visits but halted early as we found training progress to be much slower than in previous iterations.

Initially, the curriculum jumped from
$\defenseiter{4}\texttt{-s5m}^\text{128 visits}$ to $\defenseiter{4}^\text{64 visits}$
but the win rate against $\defenseiter{4}^\text{64 visits}$ was very low
at 3\%, and it did not appear to be trending upwards.
After this, we reverted the curriculum back to $\defenseiter{4}\texttt{-s5m}^\text{128 visits}$ and enabled LCB move selection, which remained enabled in all subsequent attack iterations as well.
The result of enabling LCB move selection was a lower training win rate, presumably due to the victim playing more strongly.

The hope was that training
against the earlier checkpoint \defenseiter{4}\texttt{-s5m} for longer
would yield stronger performance once the curriculum advanced to \defenseiter{4},
but we still had a low win rate of 5\% when we reached $\defenseiter{4}^\text{64 visits}$.

We then changed the curriculum to reduce the starting visit count for \defenseiter{4} from 64 to 4, which worked better.
We suspect that we could have skipped the \defenseiter{4}\texttt{-s5m} intermediate checkpoint entirely and immediately initialized the curriculum
at \defenseiter{4} with 4 visits, saving 420 V100 GPU-days and 360 million steps of training.
We therefore stopped using intermediate curricula checkpoints in subsequent attack iterations and
switched to starting the curriculum against the target victim \defenseiter{n} at very low visits.

\paragraph{Attack Iteration 5:}
The curriculum consisted of
\defenseiter{5} with 4--32 visits.
We halted this run because progress was slow, and it did not look like we would reach higher
victim visits within our compute budget.

\paragraph{Attack Iteration 6:}
From this point onward, we did not anticipate reaching high victim visits during training,
so we decided to run each attack iteration for around 250 million training steps, although it
is not entirely clear that running defense iterations against weak adversaries
provides useful training signal on the defense side.
We trained \attackiter{6} against a curriculum of
\defenseiter{6} with 4--16 visits.

\paragraph{Attack Iteration 7:}
We trained \attackiter{7}
against a
curriculum of \defenseiter{7} with 4--64 visits.
We initially intended to train \attackiter{7} for only 200 million steps, but
the win rate in both training and evaluation started increasing noticeably faster
at 190 million steps (272 V100 GPU-days),
so we extended the training duration.

\paragraph{Attack Iteration 8:}
We trained \attackiter{8}
with a curriculum
of \defenseiter{8} with 4--16 visits.

\paragraph{Attack Iteration 9:}

We trained \attackiter{9} against a curriculum of
\defenseiter{9} with 4--512 visits, raising the win rate threshold from 75\% to 90\% at 256 visits.
At 82 GPU-days (74 million steps), we adjusted the victim configuration parameters to match evaluation settings
as described in Appendix~\ref{app:config} due to a large gap between training and evaluation win rates, and
because a higher training win rate was not leading to stronger evaluation strength. We
also updated adversary configuration parameters as described in
Appendix~\ref{app:config}.

No defense iteration trains against \attackiter{9}. We trained \attackiter{9}
to see whether it could successfully attack \defenseiter{9}.

\subsection{Validation Attacks}

In this section we give more details and analysis about \attackhnine{} and \stalladv{},
our ``validation'' adversaries that attempt to attack \defenseiter{9} without access to the other \defenseiter{n} models.

\subsubsection{Atari Cyclic Attack}\label{app:atari-attack}

\begin{figure*}[btp]
    \centering
    \input{figs/plots/win-rate-vs-gpu-days-attack-h9.pgf}
    \caption{Win rate of \attackhnine{} (\plotdiamond{}) against \defenseiter{9} throughout \attackhnine{} training, warm-starting (at $x=0$) from \origtwotwoseven{}. Curriculum changes are denoted by a vertical dotted line.}
    \label{fig:vs-gpu-days-attack-h9}
\end{figure*}

\begin{figure*}[t]
    \centering
    \begin{subfigure}{0.48\textwidth}
        \centering
        \includesvg[width=0.98\textwidth]{figs/boards/validateh9_m1-2col.svg}
	\caption{\attackhnine{} induces the victim to set up several bamboo joints (\textcolor{red}{$\mathbf{\times}$}). These are normally strong shapes for connecting, e.g., if black plays a triangle-marked location, white can play the other to keep the joints connected.}
    \label{fig:bamboo-ex}
    \end{subfigure}
    \quad
     \begin{subfigure}{0.48\textwidth}
        \centering
        \includesvg[width=0.98\textwidth]{figs/boards/validateh9_m2-2col.svg}
	\caption{Ultimately, \attackhnine{} threatens to split one of the bamboo joints, and the victim prevents that by playing at the triangle location. But this is a terrible mistake---on the next move, the entire cyclic group will be captured.}
    \label{fig:validateex:m2}
    \end{subfigure}
	\caption{
    The cyclic ``bamboo joint'' strategy learned by \attackhnine{}.
    \ifarxiv \href{\demosite/iterated-adversarial-training?row=0\#v9-vs-validation-board}{Explore on our website}.
    \fi
    }
    \label{fig:validateex}
\end{figure*}

In Section~\ref{sec:robustness_against_validation_\attacker{}}, we found that the final iterated adversarially trained \defender{} \defenseiter{9} can be readily exploited at low visits by the validation attack \attackhnine{}. Figure~\ref{fig:validateex} shows the complex structure of the cyclic group formed by \attackhnine{}.

\attackhnine{}'s attack is still cyclic, but with a characteristic tendency to leave many distinct stones and groups in ``atari'', i.e.\ that could be captured on the next move by \defenseiter{9}. Moreover, it sets up ``bamboo joints'' (Figure~\ref{fig:bamboo-ex}): shapes where one player has two pairs of two stones with a one-space gap between them. They are common in normal play, and often advantageous: the two sides cannot be separated, as playing in the gap still allows connection through the remaining space. \attackhnine{} induces the victim to form a large cyclic group including these bamboo joints (Figure~\ref{fig:validateex:m2}). The attack culminates by surrounding the cyclic group and finally threatening to split one of the bamboo joints. The correct play for \defenseiter{9} is to capture one of the numerous \attackhnine{} stones in atari, but \defenseiter{9} misses the danger and connects the bamboo joint, leading to the entire cyclic group being captured.

\attackhnine{} was warm-started from \origtwotwoseven{}.
We used a curriculum of \defenseiter{9} starting at 1 victim visit and doubling until it reached 512 visits.
The curriculum win rate threshold was 75\% until reaching 256 visits, at which point the threshold increased to 90\%. Figure~\ref{fig:vs-gpu-days-attack-h9} shows the training progress.
We modified the bot configurations as described in Appendix~\ref{app:config} to make the victim configuration closer to evaluation settings and the adversary configuration closer
to the latest KataGo training runs.

\subsubsection{Stalling Cyclic Attack}\label{app:stall}

We also trained another validation attack \stalladv{} against \defenseiter{9}, and like \attackhnine{}, it beats \defenseiter{9} at low visits. \stalladv{} wins 91\% of games against \defenseiter{9} at 512 victim visits but only 5\% of games at 4096.
The scaling of the win rate attack against increasing \defenseiter{9} visit count is remarkably similar to that of \attackhnine{} (Figure~\ref{fig:vs-visits-combined}).

\begin{figure*}[btp]
    \centering
    \input{figs/plots/win-rate-vs-gpu-days-stall.pgf}
    \caption{Win rate of \stalladv{} (\plotdiamond{}) against \defenseiter{9} throughout \stalladv{} training, warm-starting (at $x=0$) from \origcyclic{}. Curriculum changes are denoted by a vertical dotted line.}
    \label{fig:vs-gpu-days-stall}
\end{figure*}

\stalladv{} was initialized from the later checkpoint \origcyclic{} and used a curriculum
of \defenseiter{9} starting at 8 victim visits and doubling until it reached 512 visits, with the win rate threshold remaining at 75\% for the entire curriculum. We used the updated victim and adversary configurations described in Appendix~\ref{app:config}. The training lasted 211 V100 GPU-days and 97 million training steps. Figure~\ref{fig:vs-gpu-days-stall}
shows the win rate of \stalladv{} throughout training---training plateaued after the checkpoint we selected for \stalladv{}.

\begin{figure}[btp]
    \centering
    \includesvg[width=0.4\textwidth]{figs/boards/stall_example-2col.svg}
    \caption{An example of the cyclic shape formed by \stalladv{} (black) in a game
    against \defenseiter{9} (white) at 512 victim visits.
    \ifarxiv \href{\demosite/iterated-adversarial-training?row=0\#v9-vs-stall-board}{See the full game on our website}.
    \fi
    }
    \label{fig:stalladv_boardstate}
\end{figure}

Figure~\ref{fig:stalladv_boardstate} displays an example of the cyclic group that \stalladv{} forms. A distinctive characteristic of its attack is that \stalladv{} spends its first few moves passing (hence the name \stalladv{}). (If \stalladv{} passes as black, \defenseiter{9} could pass to immediately win the game via komi. However, there is a KataGo setting \texttt{conservativePass = true} that stops \defenseiter{9} from passing.  \stalladv{}'s attack no longer works as black if \texttt{conservativePass} is set to \texttt{false}, but had we trained with \texttt{conservativePass = false} we likely would have simply found some other cyclic attack.)

We were surprised that \stalladv{} failed to attack \attackhnine{} at high visits---one common factor between the two prior adversaries that also do not work at high visits, \attackhnine{} and \koadv{}, is that they were both initialized to \origtwotwoseven{}. Our guess prior to training \stalladv{} was that starting from \origcyclic{} was likely to find a more potent attack. \attackhnine{} is perhaps quite difficult to attack at high visits with our training setup without intermediate adversarial training checkpoints. In the next subsection we discuss further this unpredictability of training progress.

\subsubsection{Unpredictability of Training Dynamics}

Although \attackhnine{} and \stalladv{} beat \defenseiter{9} at low visit counts, we did not find a validation attack that achieved a high win rate against \defenseiter{9} at high visit counts.
However, \attackhnine{} was only trained for 6\% as much compute as \defenseiter{9}, raising the question: how well would the attack perform were we to continue this training run?
More generally, can we predict how much more compute it would take to scale \attackhnine{} to achieve, say, a 10\% win rate against \defenseiter{9} at 65,536 visits?

Unfortunately, training dynamics are hard to forecast in advance.
For instance, \origcyclic{} made little progress for a few hundred GPU-days before abruptly finding a strategy that generalized to attack \cpfivezerofive{} at high visits~\cite{wang2023adversarial,gleave2023lwcomment}.
Looking at the training progress for \attackhnine{} (Figure~\ref{fig:vs-gpu-days-attack-h9}), \attackhnine{} makes fairly consistent progress until it stalls out late in the training run.
This training curve is consistent both with it plateauing and never achieving high win rates against high victim visits (which is perhaps the fate of \stalladv{} in Figure~\ref{fig:vs-gpu-days-stall}), or it suddenly hitting a phase change in training like \origcyclic{} did and shooting up.

Therefore even with the slowing progress of \attackhnine{} late in its training and its inability to win against \defenseiter{9} at 8192 visits, we cannot conclude that \defenseiter{9} is invulnerable at high visits.
Indeed, \attackiter{9} achieves a 42\% win rate against \defenseiter{9} at 65536 visits, showing that there is available attack surface at high visits.
This is despite \attackiter{9} only training against \defenseiter{9} up to 512 visits.

This brings up another question: how can we encourage \attacker{} training to find strategies that are likely to generalize against high visits?
\contadv{}, \largeadv{}, \koadv{}, \attackiter{9}, \attackhnine{}, and \stalladv{} only trained against their victims up to 512 visits.
Yet \contadv{}, \largeadv{}, and \attackiter{9} generalize to high visits, whereas \koadv{}, \attackhnine{}, and \stalladv{} do not.
Likewise, \citet{wang2023adversarial} found one strategy that generalizes to high visits (\origcyclic{}) and one that does not (their ``pass-adversary'').

One hypothesis is that \contadv{}, \largeadv{}, and \attackiter{9} simply used more training compute.
Another is that training is highly path dependent, so initialization, training curriculum, or stochasticity matter---\contadv{} and \attackiter{9} were initialized from later adversary checkpoints and had a curriculum involving intermediate victim model checkpoints, whereas \koadv{} and \attackhnine{} were initialized from \origtwotwoseven{} and had coarser curricula not involving
intermediate victim models.
\contadv{} and \attackiter{9} were initialized from checkpoints \attackbsixty{} and \origcyclic{} that already work against some strong high-visit victim, which may be important in biasing them away from discovering another vastly different fragile strategy that only works at low visits. (\largeadv{} and \stalladv{} do not fit cleanly into this analysis---both use a late checkpoint and a coarse curriculum, yet \largeadv{} generalizes to high visits whereas \stalladv{} does not.)

An additional point of evidence for training being path dependent is that warm-starting the adversary from \cpfivezerofive{} failed to beat \cpfivezerofive{}.
In particular, the training win rate remained flat for 50 million training steps.
Although we cannot rule out that the adversary would have eventually improved, we observed clear training progress before this point in other adversary training runs.

\section{Vision Transformers}
\label{app:vit}

In this appendix, we provide an overview of our vision transformer (ViT)
architecture and describe our ViT training procedure. For full architectural
details, see our PyTorch implementation: \maybehideurl{\vitimplementation}{\vitimplementation}.

\subsection{ViT Inputs}

Our ViTs take in the same inputs as standard KataGo CNNs,
namely two tensors of spatial and global features.

The spatial features are represented by a three-dimensional binary tensor $\mathbf{S}$ taking values in $\{0, 1\}^{\texttt{height} \times \texttt{width} \times 22}$,
where $\texttt{height}$ and $\texttt{width}$ are the maximum Go board dimensions the model supports (usually 19).
In other words, each
\maybehideurl{https://senseis.xmp.net/?Intersection}{point} of the Go board has 22 binary features associated with it.
These features encode various properties such as whether a point is occupied, the color of a stone on a point, move history, and more complicated features like whether a point is involved in a potential ladder.
For an exact specification of these features, see this source file:
\maybehideurl{https://github.com/lightvector/KataGo/blob/e133f477627eeb1924e564ff2dddc993e45428ba/cpp/neuralnet/nninputs.cpp\#L2299-L2301}{\texttt{KataGo/cpp/neuralnet/nninputs.cpp}}.

The global features are represented by a real-valued vector $\mathbf{G}$ taking values in $\mathbb{R}^{19}$.
These 19 features encode properties like which of the past 5 moves were passes, and the particular ruleset the current game is using.

\begin{figure}[btp]
    \centering
    \includegraphics[width=\columnwidth]{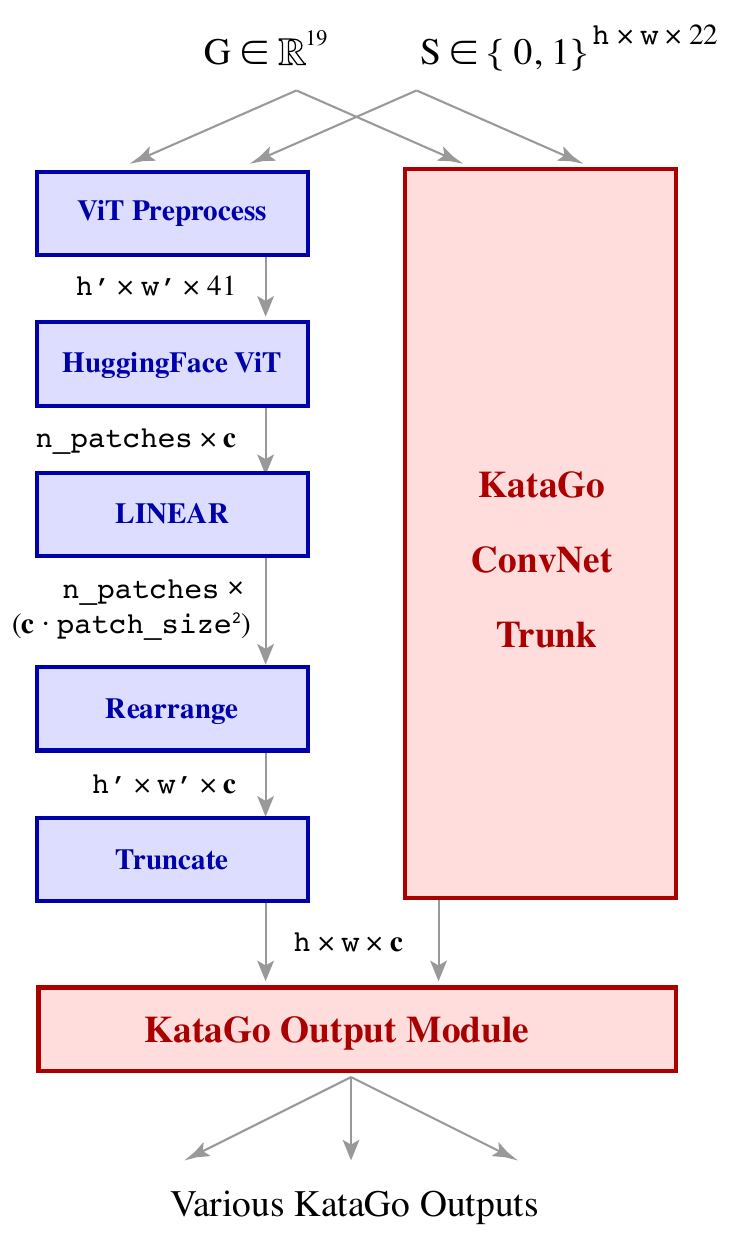}
    \caption{A diagram comparing our ViT architecture to the standard KataGo CNN architecture. Our ViT architecture replaces the KataGo \textcolor[HTML]{AA0100}{\textbf{CNN backbone}} with a \textcolor[HTML]{0006AA}{\textbf{transformer backbone}}, and reuses the KataGo CNN output layers. Boxes denote neural network components and unboxed quantities denote tensor shapes.}
    \label{fig:vit-architecture}
\end{figure}

\subsection{ViT Architecture}

Our ViT network replaces the KataGo CNN backbone with a transformer-based backbone, but reuses the same output layers as KataGo's networks (Figure~\ref{fig:vit-architecture}).
Both our transformer backbone and the KataGo's CNN backbone output a real-valued tensor with dimensions $\texttt{height} \times \texttt{width} \times c$, where $c$ is the embedding / residual-stream dimension of the network. This embedding tensor is processed by the standard KataGo output layers to produce the outputs KataGo expects networks to have: a scalar that estimates the value function, a vector that represents the next-move policy, etc. Our transformer backbone is built using a standard HuggingFace
\maybehideurl{https://huggingface.co/docs/transformers/model_doc/vit\#transformers.ViTModel}{\texttt{transformers.ViTModel}}.

\paragraph{Input Preprocessing} We zero-pad the spatial dimensions of $\mathbf{S}$ so that they are divisible by our ViT patch size (\texttt{patch\_size=2}). We then expand $\mathbf{G}$ so it has shape
$\texttt{padded\_width} \times \texttt{padded\_height} \times 19$
and concatenate it with $\mathbf{S}$ to form the actual input to our \texttt{ViTModel} of shape $\texttt{padded\_width} \times \texttt{padded\_height} \times 41$.

\paragraph{Unembedding}
The Huggingface \texttt{ViTModel} outputs a tensor of shape $\texttt{n\_patches} \times c$. We linearly project this tensor to one of size $\texttt{height} \times \texttt{width} \times c$ in the canonical way that preserves spatial locality.

\paragraph{Architecture Hyperparameters}
We tried a few different ViT architecture hyperparameters and measured how quickly they trained with supervised learning on training data from
\maybehideurl{https://katagotraining.org/}{\texttt{katagotraining.org}}. We found the following hyperparameters to work fairly well:\footnote{
To select our hyperparameters,
we tried 21 arbitrary combinations of
patch size in \{1, 2, 3, 4, 5, 7, 19\}, number of attention heads in \{6, 8, 12\},
embedding dimension in \{192, 384, 768\}, MLP dimension in \{4, 8\}\texttimes [embedding dimension],
and number of layers in \{4, 8, 16\}.
We trained each hyperparameter choice for roughly a few tens of million time steps, stopping early on many runs whose loss curves were clearly suboptimal. 
Looking at a log-log plot of training compute versus loss, we saw that the hyperparameter choice corresponding to \texttt{ViT-b4} (Table~\ref{tab:model_parameters}) did well in early training progress, so we chose it as our initial ViT model. Among models larger than \texttt{ViT-b4}, a hyperparameter setting similar to \texttt{ViT-b8} but with 8 heads did best. For simplicity we used 6 heads again for \texttt{ViT-b8} since it did not seem that number of heads affected training progress much.
We did not select \texttt{ViT-b16} via hyperparameter search but merely doubled the number of layers 
of \texttt{ViT-b8} since that was also the only change between \texttt{ViT-b4} and \texttt{ViT-b8}.
}

\begin{center}
\begin{tabular}{cccc}
    \toprule
    \textbf{Patch size} & \textbf{\# Attn. heads} & \textbf{Embed dim.} & \textbf{MLP dim.}\\
    \midrule
    \rule{0pt}{12pt} 2 & 6 & 384 & 1536 \\
    \bottomrule
\end{tabular}
\end{center}

We trained networks of varying depths,
ranging from a 4-layer ViT to a 16-layer ViT.
See Appendix section~\ref{app:vit:self-play-training} for more details.

\begin{table*}[ht]
    \centering
    \begin{tabular}{lcrrr}
        \toprule
        \textbf{Name} & \textbf{Model type} & \textbf{\# Layers} & \textbf{Embedding dim.} & \textbf{\# Parameters} \\
        \midrule
        \texttt{ViT-b4} & ViT & 4 & 384 & 7,952,501 \\
        \texttt{ViT-b8} & ViT & 8 & 384 & 15,050,357 \\
        \texttt{ViT-b16} & ViT & 16 & 384 & 29,246,069 \\
        \texttt{b6c96} & CNN & 6  & 96 & 1,001,613 \\
        \texttt{b10c128} & CNN & 10 & 128 & 2,959,329 \\
        \texttt{b15c192} & CNN & 15 & 192 & 9,875,893 \\
        \texttt{b20c256} & CNN & 20 & 256 & 23,413,525 \\
        \texttt{b40c256} & CNN & 40 & 256 & 46,632,501 \\
        \texttt{b18c384nbt} & CNN & 18 & 384 & 26,389,941 \\
        \bottomrule
    \end{tabular}
    \vspace{1em}
    \caption{Comparison of our ViT nets to KataGo CNNs in terms of depth, width, and parameter count.}
    \label{tab:model_parameters}
\end{table*}

\subsection{ViT Implementation}

KataGo implements its architectures in Python for training and C++ for self-play.
Because implementing models in C++ is fairly complex,
we only implement the ViT in Python using PyTorch.
To use ViTs during inference, we export the PyTorch model as a TorchScript model
and modify KataGo's C++ code to be able to invoke TorchScript models. Since
TorchScript models are made to be serialized and executed independently of Python,
this removes the need to implement the ViT itself in C++.

However, this comes at the cost of slower inference.
On our machines, running inference for a \texttt{b18c384nbt} or \texttt{b20c256} KataGo CNN model using TorchScript
incurred a 43\% and 28\% slowdown, respectively, compared to running them with KataGo's C++ CUDA implementation.

\subsection{Self-Play Training}
\label{app:vit:self-play-training}

In this section we describe our self-play training process for our ViT agent \vitvictim{}.

\subsubsection{Network Scaling}

In our ViT training run, we start with a small 4-block ViT network that is quick to generate data.
When the smaller model hits capacity, we switched to an 8-block network, and then finally switch to a 16-block network.
We perform each switch by first pre-training the larger model on the smaller model's data,
using the typical sliding training window method described in Appendix~\ref{app:training-window} to sample training data, but using the smaller model's existing self-play games as data.
We copy the smaller model's data into the pre-training dataset gradually in
chronological order, copying roughly one epoch's worth of data after each pre-training epoch, so pre-training
ends by training on the latest and presumably strongest data.
We discard some of the smaller model's early data under the assumption that
training on it would take extra time without much benefit due to the data being
lower quality. We still increment $N$ in Equation~\ref{eq:shuffle-window-size} to keep the window size
as large as it would have been had we not discarded the data.

\subsubsection{Training Configuration}
\label{app:vit-training-config}

Our configuration parameters matched those suggested by KataGo's example self-play configurations available in its codebase,
except that we only used Tromp-Taylor rules instead of having rule variation,
did not play any games on rectangular boards, and increased the percentage of
19\texttimes19 games to 53.6\% to match the latest KataGo training runs.
We trained exclusively on Tromp-Taylor rules because we always evaluate models under these rules, and our adversaries like \origcyclic{} were all trained only under Tromp-Taylor rules.

KataGo also seeds 14\% of its self-play games from custom positions that are rarely encountered in typical self-play~\cite{wu2023sgfpos}.
This improves play on tricky positions like Mi Yuting's Flying Dagger joseki and improves analysis on human games~\cite{wu2021sgfpos}.
Since we do not have access to this set of games, we do not include it in our ViT training run.

\subsubsection{ViT Training Run}

\begin{figure*}[btp]
\centering
\input{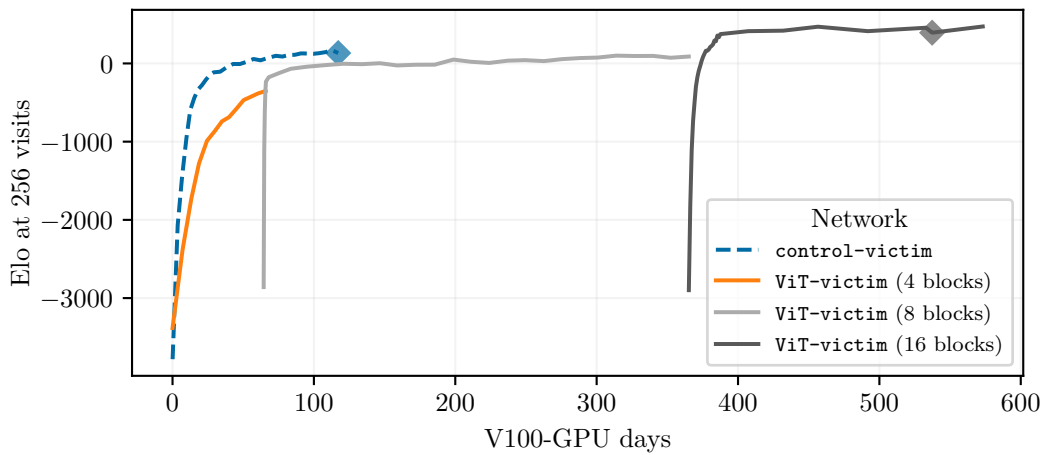}
\caption{The strength of our ViTs (\vitvictim{} = black \plotdiamond{}) throughout their training as well as a control 10-block CNN \controlbten{} (blue \plotdiamond{}) trained with the same settings.
Playing strength was estimated by playing the models against each other as well as against a few KataGo networks.
}
\label{fig:elo-vs-vit-gpu-days}
\end{figure*}

Figure~\ref{fig:elo-vs-vit-gpu-days} shows the strength of our networks throughout self-play training.
We successively trained three ViT networks (4-block, 8-block and 16-block) along with a control 10-block CNN.
Larger ViT networks reached a higher Elo but quickly saturated.
However, the ViT networks did not necessarily reach model capacity, as we were able to reach still higher Elo ratings by distilling KataGo CNN self-play training games into the ViT network.

We started with training a 4-block ViT with 600 visits using a
configuration matching an example KataGo configuration, similar to the
actual configuration used for training KataGo's 6-block and early 10-block
networks.\footnote{KataGo example configuration:
\url{https://github.com/lightvector/KataGo/blob/7488c47b6f6952f9703d9209f9afbd8d38a8afb5/cpp/configs/training/selfplay1.cfg}}
We trained for 64 GPU-days and 213 million steps.

We then switched to a 8-block ViT, pre-training it on the 4-block ViT's
latest 24.9 million data rows (100 million training steps).
After pre-training (2 V100 GPU-days, 92 million steps), the 8-block ViT was about 175 Elo stronger than the 4-block ViT at 256 visits.
We then began self-play with the 8-block ViT.
After 20 V100 GPU-days and 48 million steps of self-play, we increased the number of self-play visits to 1000 visits by
swapping our configuration to make it similar to that used for training KataGo's 10-block and 15-block
nets.\footnote{KataGo example configuration: \url{https://github.com/lightvector/KataGo/blob/7488c47b6f6952f9703d9209f9afbd8d38a8afb5/cpp/configs/training/selfplay8b.cfg}}
At 461 million steps with self-play, we gained about 264 Elo at 256 visits.
At this point we spent 301 V100 GPU-days on training the 8-block ViT. It was still making slow training progress,
but we decided to switch to a larger architecture in hopes of achieving a faster increase in playing strength.

When we switched to a 16-block ViT, we pre-trained on the latest 24.9 million data
rows generated by the 4-block ViT as well as all the data rows generated by the
8-block ViT, totaling 139 million data rows.
For this pre-training, we used data-parallel training on 8 GPUs to decrease wall-clock training time.
After training had reached 78\% of the pre-training data,
we noticed signs of overfitting: playing strength
decreased, and the ViT's value loss (loss on the model's prediction of whether a position will lead to a win or a loss)
was decreasing on training data yet increasing on validation data.
We mitigated this by increasing the minimum window size $m$ from 250000 to 10 million in
Equation~\ref{eq:shuffle-window-size}, which roughly quadrupled the current window size.
After training on 93\% of the data, we reduced the learning rate by a factor of 2
since the loss plateaued.

After pre-training (22 V100 GPU-days, 532 million steps), the 16-block ViT was 286 Elo stronger than the 8-block ViT at 256 visits.
We then started self-play. After 75 V100 GPU-days and 74 million steps of self-play,
we increased the visits to 2000 and matched a configuration similar to that used
for training KataGo's b18
models.\footnote{KataGo example configuration: \url{https://github.com/lightvector/KataGo/blob/7488c47b6f6952f9703d9209f9afbd8d38a8afb5/cpp/configs/training/selfplay8mainb18.cfg}}
At 126 V100 GPU-days and 104 million steps of self-play, we reduced the learning rate by another factor of 2.
We stopped self-play at 118 million steps of self-play, at which point we had spent 172 V100 GPU-days on 16-block training and gained another 18 Elo at 256 visits. The resulting model is \vitvictim{}.
(We were able to train the ViT for longer to boost its strength by an estimated 100 Elo at 256 visits, but we had already run evaluations on the chosen checkpoint, and the boost in the strength was not large enough to justify running new evaluations.
We suspect there is still more capacity in \vitvictim{}'s architecture because when
we trained a separate 16-block ViT on training data from \url{katagotraining.org}
generated by KataGo's stronger CNN networks, the resulting model was an estimated 277 Elo stronger than \vitvictim{} at 300 visits.)

\subsubsection{Control CNN Training Run}
\label{app:control-bten}

As a control run, we train a model \controlbten{} with a 10-block CNN architecture (\texttt{b10c128} in Table~\ref{tab:model_parameters}).
In total, we train it for 121 V100 GPU-days and 419 million steps.
We started with the same 600-visit configuration used by the 4-block ViT, and
at 29 V100 GPU-days (147 million steps), we switched to the 1000-visit configuration used by the 8-block ViT.
At 35 GPU-days and 64 GPU-days (166 million and 251 million steps), we cut the learning rate in half, and
at 110 GPU-days (396 million steps), we reduced the learning rate by 40\%.
By the end of the training run, the model was about as strong as
the 8-block ViT.

ViTs seem to be less suited to Go than than CNNs:
in Figure~\ref{fig:elo-vs-vit-gpu-days}, we see that \controlbten{} learned quicker than
the 4-block ViT and plateaued at slightly higher strength than the 8-block ViT,
despite the 8-block ViT having five times as many parameters.

\subsection{Training \vitadversary{}}
\label{app:vit:adversary}

\begin{figure*}[btp]
\centering
\input{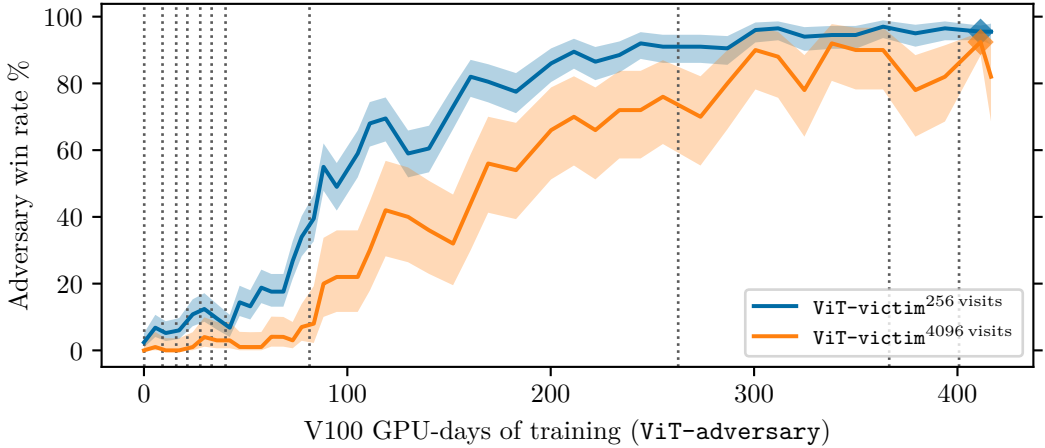}
\caption{Win rate (\%) of \vitadversary{} (\plotdiamond{}) against our superhuman ViT agent \vitvictim{} throughout \vitadversary{} training. The zero of the x-axis represents the win rate of the warm-start \origcyclic{} against \vitvictim{} before the fine-tuning against \vitvictim{} began.
Dotted lines represent victim visit increases.
}
\label{fig:vs-gpu-days-attack-vit}
\end{figure*}

Figure~\ref{fig:vs-gpu-days-attack-vit} shows the win rate of \vitadversary{} against \vitvictim{} throughout \vitadversary{}'s training.
We trained the adversary for 409 V100 GPU-days and 328 million steps, stopping the run once we had high win rates against \vitvictim{} at 32768 visits, which we estimate to be just shy of superhuman (Appendix~\ref{app:elo}).
We fine-tuned \vitadversary{} from \origcyclic{} after observing that \origcyclic{} is able to win against the final ViT at low victim visits.
We used a curriculum of \vitvictim{} starting with 1 visit and doubling until 2048 visits. The curriculum win rate threshold was 75\% until the curriculum reached 256 visits, after which the threshold was increased to 90\%.

At 262 V100 GPU-days (206 million time steps) the curriculum reached 1024 victim visits. However, we noticed that the training win rate was higher than the evaluation win rate by about 14\%, and also that the drop in win rate when the curriculum moved on to a higher visit victim
was small. We considered it desirable to train for longer at lower victim visits since it would be cheaper
to generate training data and high win rates at low visits were likely to translate to high win rates at high visits.
We therefore changed the configuration parameters to bring the victim closer to evaluation settings as described in Appendix~\ref{app:config}. This reduced training win rate, so we rewound the curriculum from 1024 visits to 256 visits. With more training, the curriculum eventually reached 1024 visits again.

\subsection{ViT Vulnerability throughout Training}
\label{app:vit:vulnerability}

\begin{figure*}[btp]
\centering
\input{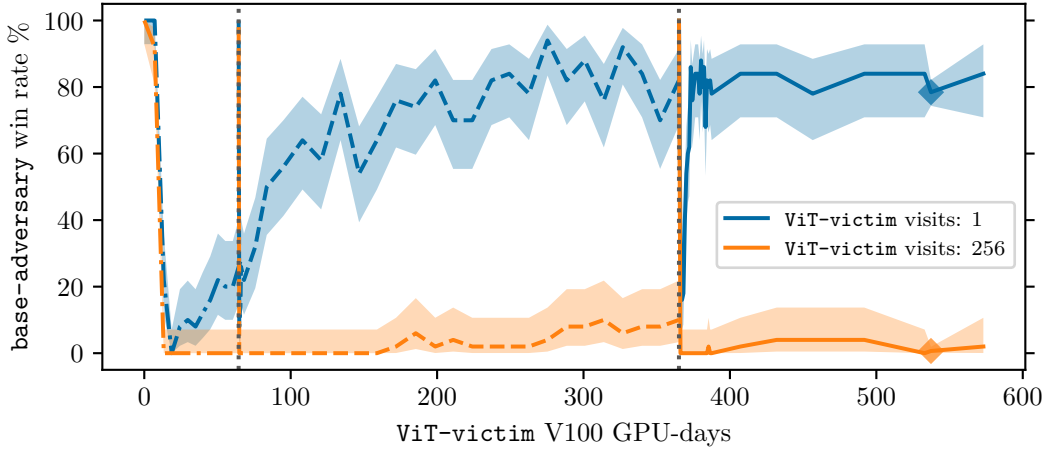}
\caption{
  Vulnerability of \vitvictim{} to \origcyclic{} throughout \vitvictim{} training.
  A dotted gray line represents switching to a larger ViT architecture,
  at which point the vulnerability drops as the larger architecture is initialized randomly but then quickly rises during pre-training.
}
\label{fig:vs-vit-gpu-days-origcyclic}
\end{figure*}
\begin{figure*}[btp]
\centering
\input{figs/plots/vs-vit-gpu-days-vitadversary.pgf}
\caption{
  Vulnerability of \vitvictim{} to \vitadversary{} throughout \vitvictim{} training.
  A dotted gray line represents switching to a larger ViT architecture,
  at which point the vulnerability drops as the larger architecture is initialized randomly but then quickly rises during pre-training.
}
\label{fig:vs-vit-gpu-days-attack-vit}
\end{figure*}

\begin{figure*}[btp]
\centering
\input{figs/plots/vs-control-b10-gpu-days-origcyclic.pgf}
\caption{
  Vulnerability of \controlbten{} to \origcyclic{} throughout \controlbten{} training.
}
\label{fig:vs-control-b10-gpu-days-origcyclic}
\end{figure*}
\begin{figure*}[btp]
\centering
\input{figs/plots/vs-control-b10-gpu-days-vitadversary.pgf}
\caption{
  Vulnerability of \controlbten{} to \vitadversary{} throughout \controlbten{} training.
}
\label{fig:vs-control-b10-gpu-days-attack-vit}
\end{figure*}

Figures~\ref{fig:vs-vit-gpu-days-origcyclic} and \ref{fig:vs-vit-gpu-days-attack-vit} show the vulnerability of \vitvictim{} to \origcyclic{} and \vitadversary{}. We observe that vulnerability to both \attackers{} develops early in training and shows no sign of decreasing. Figures~\ref{fig:vs-control-b10-gpu-days-origcyclic} and \ref{fig:vs-control-b10-gpu-days-attack-vit} show the same results for the control CNN model \controlbten{}.

\begin{figure}[btp]
\centering
\input{figs/plots/vs-visits-control-b10.pgf}
\caption{Win rate (\%) for \controlbten{} against \vitadversary{} and \origcyclic{}, with varying victim visits.}
\label{fig:vs-visits-control-b10}
\end{figure}

Figure~\ref{fig:vs-visits-control-b10} shows the win rate of \controlbten{} against
\vitadversary{} and \origcyclic{} at varying amounts of \controlbten{} visits (the corresponding plot for \vitvictim{} is Figure~\ref{fig:vs-visits-combined}). We see that \controlbten{} is also highly vulnerable to \vitadversary{}, indicating that \vitadversary{} is not conducting an architecture-specific attack.

\begin{figure*}[btp]
\centering
\input{figs/plots/win-rate-origcyclic-v1-vs-vit-elo.pgf}
\caption{
  Plot of several \vitvictim{} (black \plotdiamond{}) and \controlbten{} (blue \plotdiamond{}) training checkpoints with their playing strength on the $x$-axis and their vulnerability to \origcyclic{} at 1 victim visit on the $y$-axis.
}
\label{fig:win-rate-origcyclic-v1-vs-vit-elo}
\end{figure*}
\begin{figure*}[btp]
\centering
\input{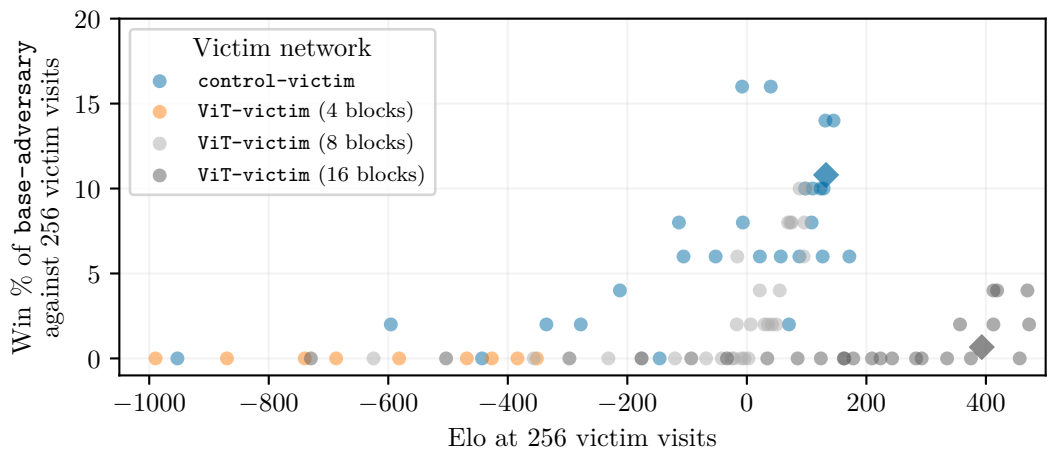}
\caption{
  Same as Figure~\ref{fig:win-rate-origcyclic-v1-vs-vit-elo} but with vulnerability at 256 victim visits on the $y$-axis.
}
\label{fig:win-rate-origcyclic-v256-vs-vit-elo}
\end{figure*}

In Figures~\ref{fig:win-rate-origcyclic-v1-vs-vit-elo} and \ref{fig:win-rate-origcyclic-v256-vs-vit-elo}, we plot how the playing strength of \vitvictim{} and \controlbten{} throughout training compares to their vulnerability to \origcyclic{}. More training yields greater strength but also increased vulnerability. \controlbten{} develops vulnerability to \origcyclic{} at a weaker strength than \vitvictim{}, suggesting that ViTs may be marginally more robust than CNNs against cyclic attacks at a given strength.

\section{Effect of adversary visit count on attack efficacy}\label{app:adv-visits}

We saw that \koadv{} achieves high win rates against \dectwentythree{} and that
\attackhnine{} and \stalladv{} both achieve high win rates against
\defenseiter{9}, all with 600 adversary visits and 512 victim visits. Here the
adversaries are performing more search than the victims, and we see in
Figure~\ref{fig:vs-visits-combined} that the adversary's win rate drops off
sharply when the victim visit count is increased. Could these adversaries merely be
winning by out-searching the victims?

\begin{figure*}[tb]
\centering
\import{figs/plots}{vs-adv-visits.pgf}
\caption{
  Win rate ($y$-axis) of our weaker adversaries at 512 victim visits and varying
  adversary visits ($x$-axis).
}
\label{fig:vs-adv-visits}
\end{figure*}

The results when we vary adversary visits (instead of victim visits)
suggests that this is not the case.  Figure~\ref{fig:vs-adv-visits} plots the win rate
of these three adversaries with varying visits while the victim visit count is
fixed at 512. We see a slower drop in win rate as adversary visits decrease.
Even with no search, the adversaries win 13--29\% of games. At 64 visits of
search the win rate is 55--91\% with \stalladv{} matching its 600-visit win
rate. Consequently, we do not think our arbitrary choice of 600 adversary visits
to be a crucial factor for the performance of our adversaries, at least at
evaluation time.

\section{Attacking on smaller boards}\label{app:board-size}

Though we focus on 19\texttimes19 games in this work, our adversaries also played games
on boards of size 7\texttimes7 to 18\texttimes18 throughout training, as described in
Appendix~\ref{app:config}. How successfully do the adversaries attack their
victims on small boards?

\begin{figure*}[tb]
\centering
\import{figs/plots}{vs-board-size.pgf}
\caption{
  Win rate of our adversaries against 512 victim visits on boards of size
  7\texttimes7 to
  19\texttimes19.
}
\label{fig:vs-board-size}
\end{figure*}

Figure~\ref{fig:vs-board-size} displays the win rate of each adversary against
their respective victim at 512 victim visits across all the board sizes.
Similar to how \citet{wang2023adversarial} only saw wins on boards of size
12\texttimes12
and larger for \origcyclic{} against \cpfivezerofive{} (Appendix F of their paper),
our adversaries only achieve wins with their characteristic adversarial
attacks on sufficiently large boards (\koadv{} and \attackiter{9} do the best
here as they start winning on boards of 11\texttimes11 and larger, whereas \attackhnine{}
the slowest and only wins on 15\texttimes15 and larger). They also achieve wins on
7\texttimes7 and 8\texttimes8
boards, but the games do not show the cyclic or gift attacks. Instead they look
like straightforward games of Go with black winning. These wins on
smaller boards are likely a result of the komi being set favorably for black (Figure
F.11 of \citet{wang2023adversarial} shows that KataGo wins as black 100\% of the time
on 7\texttimes7 and 8\texttimes8 boards when playing against itself) and the
game tree on very small boards being small enough that the adversaries can
dedicate some capacity to playing competently against their victims' most common
lines of play.

The first implication of this is that for those interested on building on our
work, it may not be fruitful to focus on small boards in spite of the reduced
compute costs. It is possible that had we increased the training frequency on
small boards that we would have uncovered adversarial attacks in that setting,
but in general we should expect that decreasing game tree size makes defense
easier for the victim.

The lead developer of KataGo considers 8\texttimes8 Go is ``tentatively (non-rigorously)
solved''~\cite{wu2023smallboards}, in that KataGo ``accurately enumerate[s] the
majority of optimal opening lines even if many less important lines are
uncertain''~\cite{katagobooks}. We are not interested in studying regimes
in which ML systems can be robust by performing nearly optimally since we do not
expect ML systems performing complex ``real-world'' tasks to be in this regime,
and we want our results to be relevant to the broader AI safety community.
Therefore very small Go board sizes are not of interest to us.
We could use a weaker victim model to compensate, but then we might no longer be
examining a superhuman system, which is one of the major reasons we are
interested in the domain of Go.
Moreover, most Go bots are optimized to play on 19\texttimes19 boards, and it
would have been a less convincing result to find adversarial attacks on settings
that the victim is not tuned for.
We recommend that future work attacking or defending KataGo should focus on
board sizes of at least 13\texttimes13 if 19\texttimes19 is too costly.

\begin{figure*}[tb]
\centering
\begin{subfigure}[t]{0.48\textwidth}
    \includesvg[width=0.98\textwidth]{figs/boards/vit-11x11-miseval-2col.svg}
    \caption{Misevaluated cyclic position}
    \label{fig:vit-misevaluate-cyclic}
\end{subfigure}
\begin{subfigure}[t]{0.48\textwidth}
    \includesvg[width=0.98\textwidth]{figs/boards/vit-11x11-miseval-modified-2col.svg}
    \caption{Correctly evaluated variation}
    \label{fig:vit-misevaluate-noncyclic}
\end{subfigure}
\caption{(\subref{fig:vit-misevaluate-cyclic}) is a cyclic position that
  \vitvictim{} misevaluates at 512 visits. This is a winning position for black,
  but across 10 trials \vitvictim{} gives an average win rate of 79.9\% for white.
  (\subref{fig:vit-misevaluate-noncyclic}) swaps one stone to break the cycle, and
  \vitvictim{} now correctly evaluates it, giving white an average win rate of
  1.7\%. (At higher visits \vitvictim{} begins evaluating
  (\subref{fig:vit-misevaluate-cyclic}) correctly. At 2048 visits \vitvictim{}
  estimates white's win rate to be 6.5\%.)
}
\label{fig:vit-misevaluate}
\end{figure*}

The second implication is that given that \vitvictim{} is vulnerable to
\vitadversary{} on boards as small as 13\texttimes13, we do not think marginally
increasing ViT depth would change its robustness to cyclic attacks. One
reasonable criticism to our vision transformer experiment is that \vitvictim{}
only has 16 layers, which may not be deep enough to perform the kind of global
analysis needed to correctly play on cyclic positions. One thing you might
expect if \vitvictim{} is merely insufficiently deep to be robust against cyclic
attacks is that \vitvictim{} would at least be more robust than CNNs at small
board sizes on which global reasoning is easier. But
Figure~\ref{fig:vs-board-size} shows \vitvictim{} is still highly vulnerable on
13\texttimes13 boards---not much different to \citet{wang2023adversarial}'s result that
\cpfivezerofive{} is still vulnerable on 12\texttimes12 boards. Moreover, we were able to
manually construct an 11\texttimes11 cyclic position that \vitvictim{} misevaluates at
512 visits (Figure~\ref{fig:vit-misevaluate}). \vitvictim{} misjudges
specifically cyclic shapes and does so egregiously, even on small board sizes.

\section{Network Strength}\label{app:elo}

\subsection{Performance of Defenses vs Base KataGo Networks}\label{app:elo:vskatago}

\begin{figure*}[btp]
\centering
\input{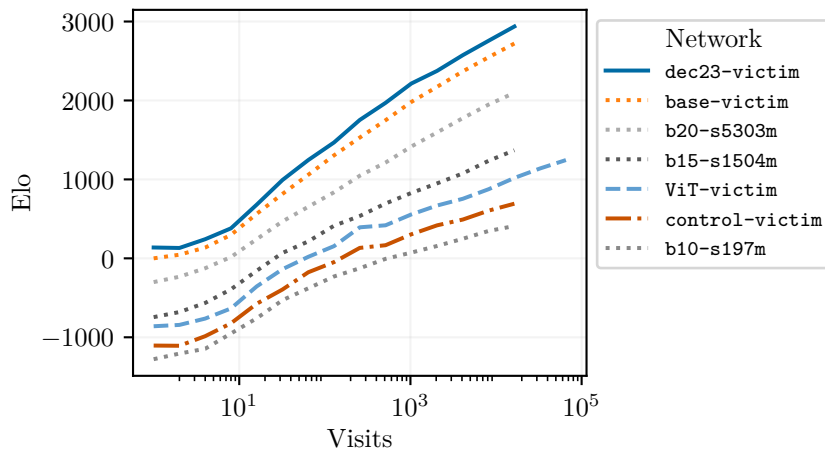}
\caption{
     Elo strength of networks (different colored lines) by visit count ($x$-axis).
     The four dotted lines are KataGo networks.
}
\label{fig:elo-by-visits}
\end{figure*}

We estimate the strength of the defended victims at playing regular Go games by pitting them against regular KataGo networks. We find that the defended victims \dectwentythree{}, \maytwentyfour{} and \defenseiter{9} all possess superhuman Go capabilities, and \vitvictim{} can play at the level of top humans.

We evaluate \dectwentythree{}, \maytwentyfour{} (positional adversarial training; Section~\ref{sec:internaladvtraining}) and \vitvictim{} (vision transformer; Section~\ref{sec:vit}) by playing games against
several KataGo networks at varying visit counts and then running a Bayesian Elo estimation algorithm.\footnote{
A script in KataGo's codebase implements this algorithm: \url{https://github.com/lightvector/KataGo/blob/master/python/summarize_sgfs.py}.}
We plot the results in Figure~\ref{fig:elo-by-visits}.
The KataGo networks we use are
\texttt{b10-s197m}, \texttt{b15-s1504m}, \texttt{b20-s5303m}, and \cpfivezerofive{}, which
\citet{wang2023adversarial} refer to \texttt{cp79}, \texttt{Original}, \texttt{cp127}, and \texttt{Latest} respectively.

We estimate that \vitvictim{} at 32768 visits is 1125 Elo stronger than \cpfivezerofive{} at 1 visit. Using \citet{wang2023adversarial}'s estimate that \cpfivezerofive{} at 1 visit would have an
Elo of 2738 on \url{goratings.org}, \vitvictim{} at 32768 visits has an
estimated Elo of 3863. This is just shy of superhuman, as the strongest historical Elo
rating on \url{goratings.org} is 3877 at the time of writing (as of 2024-05-02).
At 65536 visits, \vitvictim{} has an estimated Elo of 3974, which is superhuman.

Likewise, \dectwentythree{} at 64 visits is 1213 Elo stronger than \cpfivezerofive{} at 1 visit, giving it a superhuman estimated Elo of 3951. \maytwentyfour{} is slightly stronger with an estimated Elo of 3967 at 64 visits.

\begin{figure*}[btp]
\centering
\input{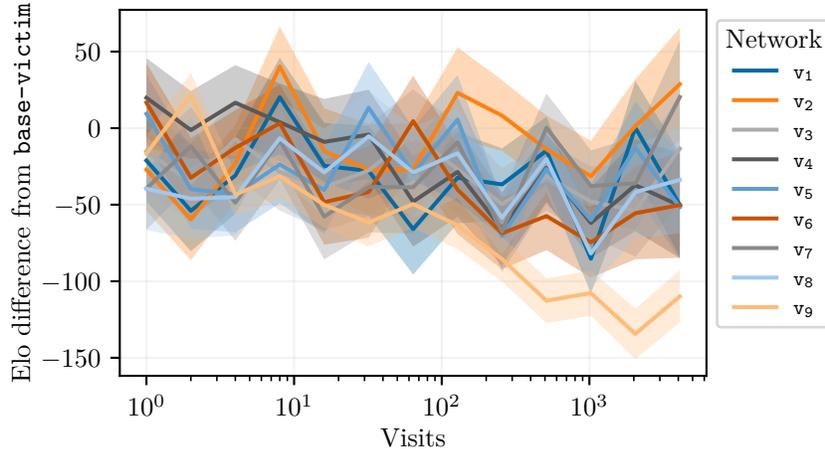}
\caption{
     Elo difference between each \defenseiter{n} to \cpfivezerofive{} at visit counts up to 4096. Shaded regions are the standard deviation of the Elo estimate.
     Each \defenseiter{n} is slightly weaker than \cpfivezerofive{}.
}
\label{fig:elo-hs}
\end{figure*}

We estimate the strength of \defenseiter{9} by playing against \cpfivezerofive{} at varying visit counts. We plot the results in Figure~\ref{fig:elo-hs}.
We estimate that \defenseiter{9} has an Elo of 5017 at 4096 visits, which is 107 points weaker than \cpfivezerofive{} but still clearly superhuman.

\subsection{Performance of \vitvictim{} Against Human Players}\label{app:elo:vithuman}

We also deployed a 64-thread, 65536 visit / move version of \vitvictim{} on the KGS Online Go server~\citep{kgs2022}.
From the previous section, we estimate this bot has a \texttt{goratings.org} Elo of 3848, around the level of a top human professional.\footnote{In the previous section we estimated a \texttt{goratings.org} Elo of 3974 for \vitvictim{} at 65536 visits. However, KataGo's internal benchmarks suggest that above 5000 visits, each search thread decreases performance by around 2 Elo (see \url{https://github.com/lightvector/KataGo/blob/v1.13.0/cpp/program/playutils.cpp\#L868}). Adjusting for the 63 additional threads gives an Elo of $3974 - 2\cdot63 = 3848$. Using multiple search threads parallelizes inference, decreasing inference latency at the cost of overall strength. That is to say, for a fixed number of visits, using fewer threads generally leads to a stronger agent. All of our training and evaluation runs in the paper are done with a single search thread unless noted otherwise.}
Our results support this: our \vitvictim{} bot achieved a
\maybehideurl{https://archive.is/J4Fjz}{peak ranking of 9th} on KGS~\citep{kgs2022}, ahead of many KataGo bots but behind others playing with stronger settings.
Since professional players rarely play on KGS, we also commissioned three games against strong professionals: our bot won two out of three, losing one largely due to a weakness that also affected early versions of KataGo (see discussion below).

\subsubsection{Public KGS Games}
Our bot played 1000 ranked games on the KGS website with members of the public, achieving a
\maybehideurl{https://www.gokgs.com/graphPage.jsp?user=ViTKata001}{peak rating of 10.98 dan} on the KGS website~\citep{kgs2022}.\footnote{The KGS website has a special rating system (\url{https://www.gokgs.com/help/rmath.html}). Official ranks are discrete and only go up to 9 dan, but KGS computes an internal Elo for all players with a minimum number of ranked games. These internal Elos can go past 9 dan. See the top rated accounts at \url{https://www.gokgs.com/top100.jsp} to see some examples of this.}
We note that bots are common on the server; we follow the standard best practice of notifying players that they are playing a bot, and our bot was approved for ranked games by the KGS administrators.
Indeed, the top ranked players on KGS are dominated by bots: our ranking of 9th puts us ahead of several KataGo bots and behind several others, though the exact configuration settings of these bots are unknown.
However, the majority of ranked games played by our \vitvictim{} bot were against human players, usually with our bot giving 1 to 6 stones of handicap to the human player.

Despite our \vitvictim{} bot having a strong showing on the KGS Online Go server, it is understood within the Go community that strong professional players rarely play on KGS. Thus our results on KGS only show that a 64-thread, 65536 visit / move version of \vitvictim{} is much stronger than many strong amateur Go players.

\subsubsection{Games Against Professional Go Players}

We therefore also commissioned a game against the 7 dan professional Yilun Yang and two games against the 4 dan professional Ryan Li. The players were informed that they were playing a bot and agreed to acknowledgement in the paper. They were also compensated at a rate greater than 4x the minimum wage in the relevant jurisdiction.

Yilun Yang played with 90 minutes base time for each player, and 5 periods of 30 seconds byo-yomi overtime. \vitvictim{} won, with Yang feeling he may have gotten behind early and missed some better ways to play in the middle game.

Ryan Li played with 5 minutes base time per player, and the same 5x30 byo-yomi overtime. \vitvictim{} lost the first game and won the second. In the first game, Li played the ``Flying Dagger'' joseki, a notoriously difficult opening corner sequence, and obtained a substantial early advantage after \vitvictim{} misplayed. Li played accurately for the rest of the game and \vitvictim{} never caught up. This joseki was a known weakness in early versions of KataGo as well; it was eventually corrected through manually adding positions from the sequence to the training run. In our training, we did not include those positions (Appendix~\ref{app:vit-training-config}), and it seems \vitvictim{} developed a similar weakness.

In the second game Li played, we requested and Li agreed to avoid that joseki, and with that constraint \vitvictim{} won.

Overall, these results indicate \vitvictim{} has some weaknesses that might lead to a lower Elo. But in general it plays at a strong professional level, in line with our original estimate. Explore the games on the accompanying
\maybehideurl{\demosite/vit\#contents}{project website}.

\section{Human Replication of Attacks}
\label{app:humanatk}

A Go expert author (\anonkellinpelrine{}) was also able to replicate several of our attacks after studying the game records but without AI assistance at attack time. Full game records, along with additional commentary on the play, are available on our
\maybehideurl{\demosite/}{website} and linked in the following sections.

\subsection{Human Replication of the Gift \Attacker{}}

Unlike the next attack in Appendix~\ref{app:humanatk-cont}, setting up the apparent shapes for the gift attack against \dectwentythree{} is relatively straightforward. It was not too challenging to produce a
\maybehideurl{\demosite/positional-adversarial-training\#dec23-vs-human-gift}{successful attack} against 1 victim visit. In particular, it was quite simple to induce \dectwentythree{} to make errors---the challenge was ensuring \dectwentythree{}'s lead was sufficiently narrow for the errors to change the game outcome.

Scaling to higher visits, however, proved difficult. Multiple attempts at 256 and 512 visits failed. We hypothesize that this is because the victim must assign enough value to the sending-two ``gift'' move to play it, but at the same time not keep searching locally and see disaster coming after the adversary's next move. These requirements are conflicting: if there are valuable areas to play elsewhere then the victim is likely to play those instead of sending-two, but if there are none, then there are none for the adversary either, so the victim is more likely to expect the adversary to continue locally and accept the gift -- and then to see the danger.

This need for some but not too much local search so that the victim plays the local ``gift'' move is in stark contrast with all versions of the cyclic attack, where the attack is more likely to succeed the less search is allocated by the victim to the locality of the vulnerability. Furthermore, at least in the versions of the attack observed so far, the number of moves that the victim needs to look ahead locally, between its deciding move and realized loss (adversary group living), is fixed and small. This again contrasts with the cyclic attack, where the deciding move can take place a virtually arbitrary amount of moves ahead of realized loss (cyclic group captured or something else lost while saving the cyclic group).

This requirement means the attack needs to balance search probabilities over the entire board to a greater and greater degree at higher visits.
By contrast, in the cyclic attacks it suffices to control the local situation to make the attack more hidden, requiring a greater victim search depth needed to notice the attack. This also fits with our empirical observations: humans can perform the attack at one visit but seemingly not at 256+, while \koadv{} can reach 512 visits and somewhat beyond but falls off very sharply after 1024 visits (Figure~\ref{fig:vs-visits-combined}). \koadv{} is likely able to balance search probabilities of the victim with much higher precision than humans can, but it becomes prohibitively difficult at high enough visits.

\subsection{Human Replication of the Continuous \Attacker{}}\label{app:humanatk-cont}

\maybehideurl{\demosite/positional-adversarial-training\#dec23-vs-human-cyclic}{This attack} was the most challenging to replicate, requiring multiple components chained together.
Despite carefully engineering the shapes of the attack like the distinctive double cut formation highlighted in Figure~\ref{fig:contadv_boardstate}, which can already be challenging, the attack still failed many times. There are likely additional subtleties or obfuscation to the attack that were non-obvious. 

In the successful attempt, the final threat against the cyclic group was also a natural move for a purpose other than attacking that group, which we hypothesize might make \dectwentythree{} less likely to search follow-up sequences attacking its cyclic group, and consequently miss the danger that it is in. The attack was performed against \dectwentythree{} playing with 512 visits. Although it would likely require many attempts and further strategizing, we believe it should still be possible for humans to achieve wins against higher visits. It is possible to engineer situations where the final sequence leading to capturing the cyclic group also poses a game-deciding threat against something else, and where the victim would have to look far ahead to see that it will not be able to defend against that threat without losing the cyclic group. This should compensate for higher search. One notable piece of evidence for this is that the critical move in the successful game is also (mis-)played by KataGo with 4096 visits. Meanwhile, the other components of the attack do not seem related to search depth and should not be harder to achieve.

We note, however, that the AI \contadv{} does not seem to use this double threat subterfuge. Thus, either it is exploiting subtle blind spots in the victim (which may be much easier with gray-box access than for a black-box human replicator), or there may be some additional strategic component of the attack that we did not find.

\subsection{Human Replication of the Big \Attacker{}}

It was comparatively easy to imitate \largeadv{}'s attack, despite us targeting \dectwentythree{} instead of \largeadv{}'s usual opponent \maytwentyfour{}, for comparability with the preceding human attack. In contrast to the continuous attack replication from Appendix~\ref{app:humanatk-cont} which took over 20 attempts against 512 victim visits, on the third attempt with the big cyclic attack approach we
\maybehideurl{\demosite/positional-adversarial-training\#dec23-vs-human-big-cyclic}{successfully attacked}
\dectwentythree{} at 4,096 visits.

The principal challenge here was making sure the inside group was big but not so big that it was invaded or caused the cyclic group to join with the victim's other groups. In addition, to minimize odds of the victim saving the cyclic group by capturing one of the surrounding ones, \anonkellinpelrine{} aimed to have many liberties on those groups. Generally, though, making the cyclic group bigger is a straightforward objective to realize.

\subsection{Human Cyclic Attack on \vitvictim{}}

\anonpelrine{} was also able to use a cyclic attack to
\maybehideurl{\demosite/vit\#vit-vs-human-cyclic}{beat \vitvictim{}}.
This attack was the easiest to execute of those discussed in this section. It was performed against a 64-thread, 65536 visit / move version of \vitvictim{}, the same used in the strength evaluation in Appendix~\ref{app:elo:vithuman}. The shape used for the inside group paralleled some of the wins by \origcyclic{} against this victim. The attack emphasized ensuring lots of liberties for the groups surrounding the cyclic one so that \vitvictim{} would have to see the danger early to have a way out.

\section{Transfer}

\begin{figure*}[tb]
\centering
\import{figs/plots}{win-rate-heatmap-full.pgf}
\caption{
  We extend Figure~\ref{fig:win-rate-heatmap-iterated-adv}'s plot of \attackers{}' win rates against various \defenders{} to include more \attackers{} on the $x$-axis and more \defenders{} on the $y$-axis.
}
\label{fig:win-rate-heatmap-full}
\end{figure*}

\begin{figure*}[tb]
\centering
\import{figs/plots}{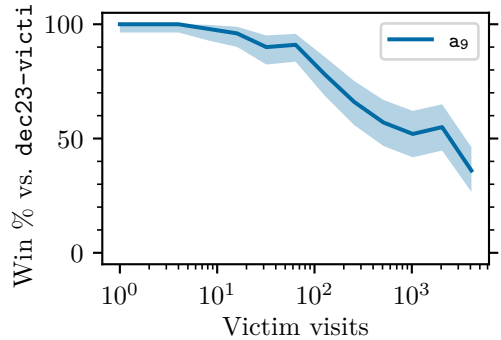}
\caption{
    Win rate ($y$-axis) of \attackiter{9} versus \dectwentythree{} at varying victim visits ($x$-axis), demonstrating considerable transfer performance.
}
\label{fig:transfer-r9-vs-b18-visits}
\end{figure*}

Figure~\ref{fig:win-rate-heatmap-full} shows the result of playing \attackers{} against a variety of \defenders{}. The ability of \defenders{} to defeat \attackers{} they were not trained against provides evidence of their robustness. 

\paragraph{\Defenders:} We find all \defenders{} remain vulnerable at extremely low amounts of search (4 \defender{} visits), although \dectwentythree{} does better than others. \cpfivezerofive{} through \defenseiter{4} progressively improve at defending against \contadv{}, after which their performance plateaus.

\paragraph{\Attackers{}:}
\attackiter{9}, trained against \defenseiter{9}, transfers surprisingly well to defeat \dectwentythree{}, winning 66\% of games at 256 visits and 36\% at 4096 visits (Figure~\ref{fig:transfer-r9-vs-b18-visits}).
\attackhnine{}, trained against \defenseiter{9}, wins 4\% of games against \dectwentythree{} at 256 victim visits. By contrast, \contadv{}, trained against \dectwentythree{}, wins 5\% of games against \defenseiter{9}.
\koadv{} does not transfer at all to other \defenders{}, achieving no wins even at 4 visits against \cpfivezerofive{} and \vitvictim{}.

\section{Compute Resources}\label{app:compute}

\subsection{Compute Infrastructure}\label{app:compute-infra}
We ran experiments using cloud computing infrastructure orchestrated with Kubernetes configured with the Kueue batch scheduler.
Each container (consisting of a self-play worker, a victim-play worker, or a model training worker) runs Ubuntu 20.04 and has 16 CPUs, 59 Gi of memory, and one GPU. %

We used A6000 GPUs for nearly all our training runs. The main exception is that
\defenseiter{1}, \defenseiter{2}, \defenseiter{3}, and \defenseiter{4}
used A100 80GB GPUs as we were trying a different compute platform.
We also used some H100 GPUs during the \koadv{} and \attackiter{9} runs, but they were mainly run on A6000 GPUs.

\subsection{Compute for Our Training Runs}

We convert our compute numbers to V100 GPU-days so that our numbers
can be straightforwardly compared to the V100-based compute estimates of \citet{wang2023adversarial}.
According to \citeauthor{wang2023adversarial}'s conversion estimates,
one A100 80GB GPU-day is 1.873 A6000 GPU-days and one A6000 GPU-day is 1.704 V100 GPU-days.
We estimate that one H100 GPU-day generated as much training data as 0.369 A6000 GPU-days.
Note we did not tune our H100 setup as we made minimal use of these GPUs.

Most of our compute estimates are measured by parsing our training logs.
However, when training \attackiter{1}, ViT, and \vitadversary{}, we made sub-optimal configuration choices that slowed down our training runs. 
For these runs, we provide idealized compute estimates by benchmarking the slow-down caused by the poor configuration and scaling our compute estimates downwards accordingly.

Our error in \attackiter{1} training was using too few game threads, a parameter controlling how many victim-play games are played at once. We were using 16--32 game threads rather than the 128--256 game threads that we used in later training runs, which gave higher training throughput.
\attackiter{1} used 703 V100 GPU-days, and we estimate that with higher game threads it would have cost 238 V100 GPU-days instead.

Our error in training our ViT networks and \vitadversary{} was using single-precision floating point rather than half-precision floating point for ViT inference. Inference with half-precision floating point is significantly faster. 
Our actual compute cost for training ViT with single-precision floating point was 
128 V100 GPU-days for the 4-block ViT, 661 V100 GPU-days for the 8-block ViT, and 457 V100 GPU-days for the 16-block ViT, totalling 1247 V100 GPU-days. We estimate that with half-precision floating point, the cost would have been 563 V100 GPU-days instead. 
For \vitadversary{}, we switched to half-precision floating point near the end of the run and spent 711.0 V100 GPU-days. We estimate that had we used half-precision floating point for the entire training run, it would have been 409 V100 GPU-days instead.

\subsection{Compute for KataGo Models}\label{app:compute-katago}

\cpfivezerofive{}, \bsixty{}, \dectwentythree{}, and \maytwentyfour{} all come from KataGo's ongoing distributed training run, which was initialized from KataGo's ``third major run.''
\citet{trainhistory2021} reports training compute estimates for the third major run, from which we can extrapolate the training cost of models from the distributed training run.
(Our compute estimate calculations are similar to those of \citet{wang2023adversarial}, except in our estimates we do not anchor on the initial 38.5 days of the third major run that generated data with smaller models, and we account for the greater search used in the distributed training run.)

In the last 118.5 out of 157 days in the third major run, the run switched from using \texttt{b20c256} nets to using \texttt{b40c256} and \texttt{b30c320} nets for self-play data generation.
The final \texttt{b20c256} net used for self-play was trained on 468,617,949 data rows whereas the third major run generated 1,229,425,124 rows in total,
so over the course of those 118.5 days, the run generated $1{,}229{,}425{,}124-468{,}617{,}949 = 760{,}807{,}175$ rows.
This segment of the run used 46 V100 GPUs, costing $118.5 \times 46 = 5451$ V100 GPU-days. The total cost of the the third run across all 157 days is 6,730 V100 GPU-days~\cite{wang2023adversarial}.

The distributed run generates data with \texttt{b40c256}, \texttt{b60c320}, and \texttt{b18c384nbt} nets, all of which have similar or higher inference cost to the \texttt{b40c256} and \texttt{b30c320} used in the third major run.\footnote{\texttt{b60c320} is a strictly larger in width and depth than \texttt{b40c256} and therefore has a higher inference cost. \url{https://github.com/lightvector/KataGo/blob/v1.14.1/python/modelconfigs.py\#L1384} states that \texttt{b40c256}, \texttt{b30c320}, and \texttt{b18c384nbt} have similar inference costs.} 
Therefore, we estimate the average inference from the distributed run is at least as expensive as the average inference from the last 118.5 days of the third major run.

Moreover, the distributed training run uses more inferences to generate each data row.
The third major run used 1000 full-search visits or 200 cheap-search visits per move, 
where full searches are used to generate high-quality policy data and cheap searches are used to play games quickly~\cite{wu2020accelerating}.
The distributed training run started with 1500 full-search visits and 250 cheap-search visits~\cite{wu2020distributedvisits}. 
Assuming inference count scales proportionally with search and that data row compute cost scales proportionally with inference count, we crudely estimate each training row 
generated with these search parameters costs $1.25\times$ as much as each training row from the third major run.
The distributed run switched to 2000 full-search visits and 350 cheap-search visits in March 2023~\cite{wu2023distributedvisits}, after about 3.211 billion data rows (including the 1.2 billion from the third major run) were generated. 
We estimate each row generated with these parameters costs $1.75\times$ as much as each third-major-run row.

Putting this all together, our training compute estimate in V100 GPU-days for a model from KataGo's distributed training run that has trained on $D \ge 1{,}229{,}425{,}124$ rows is
\begin{align*}
    6730
    + ((\min\{D,3211000000\} -1229425124) \cdot 1.25 \\
    + \max\{D-3211000000,0\} \cdot 1.75)
    \cdot \frac{5451}{760807175} 
.\end{align*}
Since \cpfivezerofive{} trained on 2,898,845,681 data rows, its estimated cost is 21681 V100 GPU-days.
\bsixty{} trained on 3,323,518,127 rows, giving a cost of 25888 V100 GPU-days;
\dectwentythree{} trained on 3,929,217,702 rows, giving a cost of 33482 V100 GPU-days;
and \maytwentyfour{} trained on 4,316,597,426 rows, giving a cost of 41511 V100 GPU-days.

For adversarial training, the last KataGo network before adversarial training began, \texttt{b60-s6729m}, was trained with 3,057,177,l18 data rows~\cite{wu2022advtrainstart}.
Based solely on the number of data rows, \dectwentythree{} has had
$(3929217702 - 3057177418) / (3323518127 - 3057177418) = 3.3$ times as much adversarial training as \bsixty{},
and \maytwentyfour{} has had 4.7 times as much as \bsixty{}.
\section{Heat Maps of Cyclic Attacks}
\label{app:cyclic-heatmaps}

In this section, we present heat maps illustrating the cyclic shapes constructed by each of our cyclic adversaries. We also plot differences between heat maps to show changes in the cyclic group constructed by different adversaries.

To construct the heat maps for an adversary, we took games where the adversary beats the victim it was trained against. 
We then inspect the board state during the move at which a large cyclic group of victim stones is captured. 
To remove board symmetries, we rotate the game board so that the center of the cyclic group is in the top-left quadrant of the board,
and flip across the major diagonal of the board to keep the center of the group above the major diagonal.
We then plot the frequency of each board square being in the captured cyclic group. We also plot the adversary's stones, the victim's other stones, and the adversary and victims' stones falling in the interior of the cyclic group at the time of capture.

\begin{figure*}[btp]
\centering
\import{figs/plots}{cyclic-heatmap-r0-v4096.pgf}
\caption{Heat map showing the cyclic attack made by \origcyclic{} against \cpfivezerofive{} with 4096 victim visits of search.}
\label{fig:cyclic-heatmap-r0-v4096}
\end{figure*}

\begin{figure*}[btp]
\centering
\import{figs/plots}{cyclic-heatmap-continuous-adversary.pgf}
\caption{Heat map showing the cyclic attack made by \contadv{} against \dectwentythree{} with 4096 victim visits of search.}
\label{fig:cyclic-heatmap-continuous}
\end{figure*}

\contadv{}'s attack against \dectwentythree{} (Figure~\ref{fig:cyclic-heatmap-continuous}) shows less variation in the cyclic group than the attack made by \origcyclic{} (Figure~\ref{fig:cyclic-heatmap-r0-v4096}): the cyclic stone heat map (Figure~\ref{fig:cyclic-heatmap-continuous}) is deeply colored throughout with few lightly colored squares. We also see a larger and consistent shape of interior adversary stones for \contadv{}, along with a pattern in interior victim stones that isn't present for the \origcyclic{}.

\begin{figure*}[btp]
\centering
\import{figs/plots}{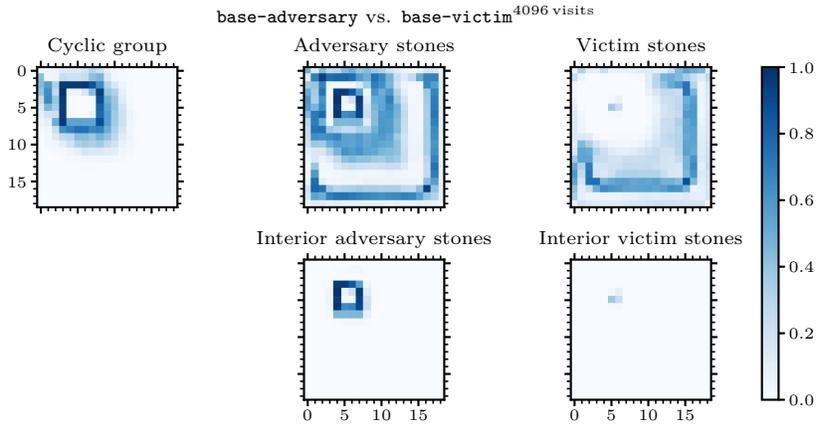}
\caption{Heat map showing the cyclic attack made by \largeadv{} against \maytwentyfour{} with 4096 victim visits of search.}
\label{fig:cyclic-heatmap-large}
\end{figure*}

Figure~\ref{fig:cyclic-heatmap-large} shows heat maps \largeadv{} against \maytwentyfour{}, displaying \largeadv{}'s distinctive, large, oblong cyclic shape.

\begin{figure*}[btp]
\centering
\import{figs/plots}{cyclic-heatmap-attack-h9.pgf}
\caption{
  Heat map showing the cyclic attack made by \attackhnine{} against \defenseiter{9}.
}
\label{fig:cyclic-heatmap-attack-h9}
\end{figure*}

Figure~\ref{fig:cyclic-heatmap-attack-h9} shows heat maps for \attackhnine{} against \defenseiter{9}, with the dark squares in the cyclic group being the bamboo joints discussed in Section~\ref{sec:attack-h9}. We also see a checkerboard pattern of adversary stones near the cyclic group. These are likely isolated pieces, mentioned in the same discussion, that could be captured if the victim saw the danger its cyclic group was in.

\begin{figure*}[btp]
\centering
\import{figs/plots}{cyclic-heatmap-stall-adversary.pgf}
\caption{
  Heat map showing the cyclic attack made by \stalladv{} against \defenseiter{9}.
}
\label{fig:cyclic-heatmap-stall}
\end{figure*}

Figure~\ref{fig:cyclic-heatmap-stall} shows heat maps for \stalladv{} against \defenseiter{9}, with \stalladv{} concentrating the cyclic group as a 6-by-6 square. Interestingly, we see a significant degree of similarity with the heatmaps of \origcyclic{} in Figure~\ref{fig:cyclic-heatmap-r0-v4096}.

\begin{figure*}[btp]
\centering
\import{figs/plots}{cyclic-heatmap-attack-vit.pgf}
\caption{
  Heat map showing the cyclic attack made by \vitadversary{} against \vitvictim{}.
}
\label{fig:cyclic-heatmap-attack-vit}
\end{figure*}

Figure~\ref{fig:cyclic-heatmap-attack-vit} shows heat maps for \vitadversary{} against \vitvictim{}, which moves the cycle into the center and forms another boundary of stones around it.

\begin{figure*}[btp]
\centering
\import{figs/plots}{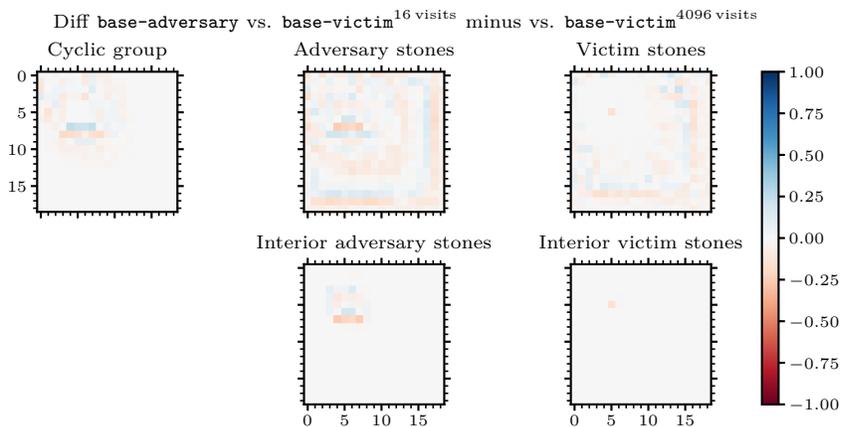}
\caption{
  Difference between the heat maps of \origcyclic{} against \cpfivezerofive{}
  with 16 (Figure~\ref{fig:cyclic-heatmap-r0}) and 4096 (Figure~\ref{fig:cyclic-heatmap-r0-v4096}) victim visits of search.
  \origcyclic{}'s attack does not change much when victim visits are increased.
}
\label{fig:cyclic-heatmap-diff-r0-v4096}
\end{figure*}

The remaining plots are heat maps for \attackiter{n} and \defenseiter{n}. We use a low victim visit count, 16, for these plots
since many \attackiter{n} do not work at high visits. Changing the number of visits usually does not change the resulting plots much---for example, with \origcyclic{} against \cpfivezerofive{}, using more victim visits does not substantially affect the shape of the cyclic attack.
Figure~\ref{fig:cyclic-heatmap-diff-r0-v4096} plots the difference between the heat map for 4096 (Figure~\ref{fig:cyclic-heatmap-r0-v4096}) and 16 (Figure~\ref{fig:cyclic-heatmap-r0}) victim visits, finding minimal differences.

\begin{figure*}[btp]
\centering
\import{figs/plots}{cyclic-heatmap-r0.pgf}
\caption{
  Heat map showing the cyclic attack made by \origcyclic{} against \cpfivezerofive{} with 16 victim visits of search.
}
\label{fig:cyclic-heatmap-r0}
\end{figure*}
\begin{figure*}[btp]
\centering
\import{figs/plots}{cyclic-heatmap-diff-r1.pgf}
\caption{
  Difference between the heat maps of \attackiter{1} (Figure~\ref{fig:cyclic-heatmap-r1}) and \origcyclic{} (Figure~\ref{fig:cyclic-heatmap-r0}).
}
\label{fig:cyclic-heatmap-diff-r1}
\end{figure*}
\begin{figure*}[btp]
\centering
\import{figs/plots}{cyclic-heatmap-r1.pgf}
\caption{
  Heat map showing the cyclic attack made by \attackiter{1} against \defenseiter{1}.
}
\label{fig:cyclic-heatmap-r1}
\end{figure*}
\begin{figure*}[btp]
\centering
\import{figs/plots}{cyclic-heatmap-diff-r2.pgf}
\caption{
  Difference between the heat maps of \attackiter{2} (Figure~\ref{fig:cyclic-heatmap-r2}) and \attackiter{1} (Figure~\ref{fig:cyclic-heatmap-r1}).
}
\label{fig:cyclic-heatmap-diff-r2}
\end{figure*}
\begin{figure*}[btp]
\centering
\import{figs/plots}{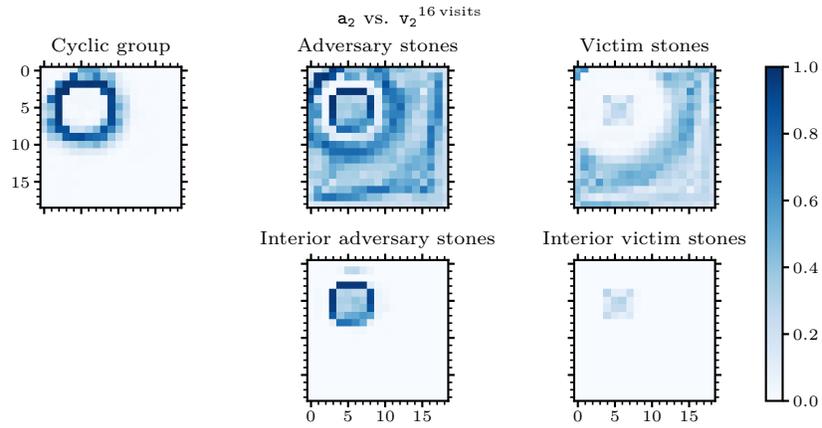}
\caption{
  Heat map showing the cyclic attack made by \attackiter{2} against \defenseiter{2}.
}
\label{fig:cyclic-heatmap-r2}
\end{figure*}
\begin{figure*}[btp]
\centering
\import{figs/plots}{cyclic-heatmap-diff-r3.pgf}
\caption{
  Difference between the heat maps of \attackiter{3} (Figure~\ref{fig:cyclic-heatmap-r3}) and \attackiter{2} (Figure~\ref{fig:cyclic-heatmap-r2}).
}
\label{fig:cyclic-heatmap-diff-r3}
\end{figure*}
\begin{figure*}[btp]
\centering
\import{figs/plots}{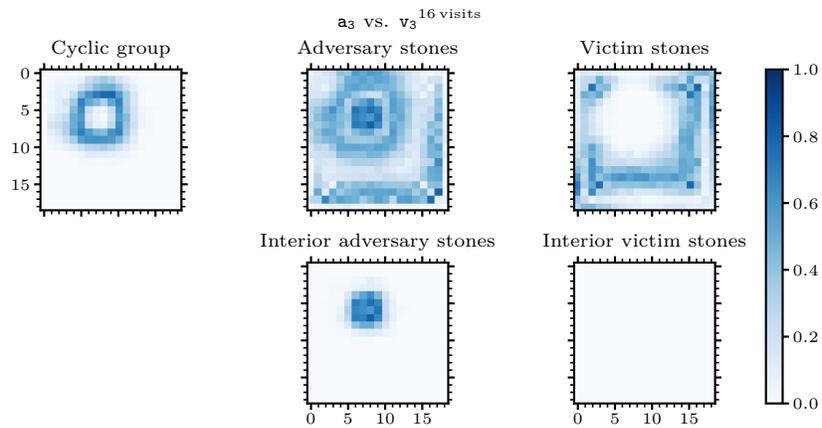}
\caption{
  Heat map showing the cyclic attack made by \attackiter{3} against \defenseiter{3}.
}
\label{fig:cyclic-heatmap-r3}
\end{figure*}
\begin{figure*}[btp]
\centering
\import{figs/plots}{cyclic-heatmap-diff-r4.pgf}
\caption{
  Difference between the heat maps of \attackiter{4} (Figure~\ref{fig:cyclic-heatmap-r4}) and \attackiter{3} (Figure~\ref{fig:cyclic-heatmap-r3}).
}
\label{fig:cyclic-heatmap-diff-r4}
\end{figure*}
\begin{figure*}[btp]
\centering
\import{figs/plots}{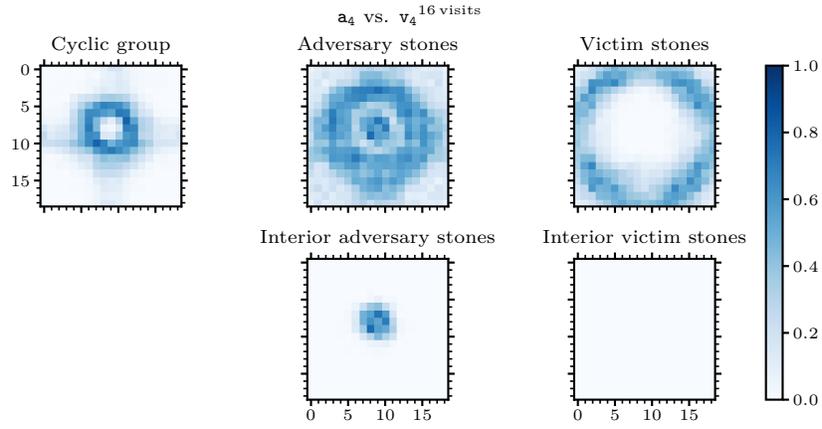}
\caption{
  Heat map showing the cyclic attack made by \attackiter{4} against \defenseiter{4}.
}
\label{fig:cyclic-heatmap-r4}
\end{figure*}
\begin{figure*}[btp]
\centering
\import{figs/plots}{cyclic-heatmap-diff-r5.pgf}
\caption{
  Difference between the heat maps of \attackiter{5} (Figure~\ref{fig:cyclic-heatmap-r5}) and \attackiter{4} (Figure~\ref{fig:cyclic-heatmap-r4}).
}
\label{fig:cyclic-heatmap-diff-r5}
\end{figure*}
\begin{figure*}[btp]
\centering
\import{figs/plots}{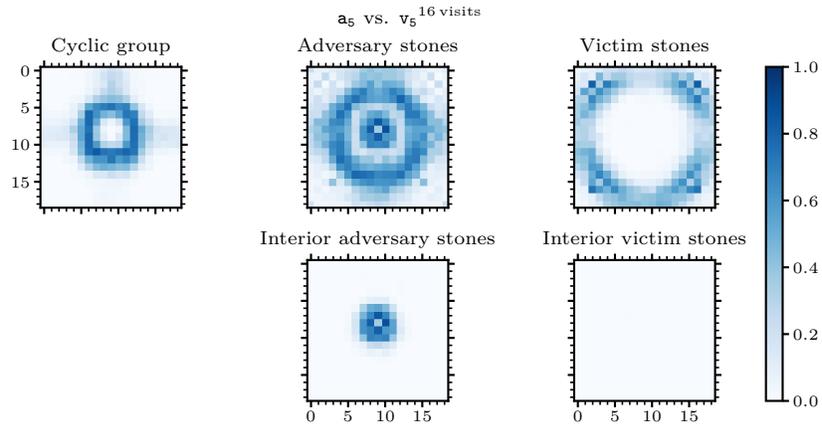}
\caption{
  Heat map showing the cyclic attack made by \attackiter{5} against \defenseiter{5}.
}
\label{fig:cyclic-heatmap-r5}
\end{figure*}
\begin{figure*}[btp]
\centering
\import{figs/plots}{cyclic-heatmap-diff-r6.pgf}
\caption{
  Difference between the heat maps of \attackiter{6} (Figure~\ref{fig:cyclic-heatmap-r6}) and \attackiter{5} (Figure~\ref{fig:cyclic-heatmap-r5}).
}
\label{fig:cyclic-heatmap-diff-r6}
\end{figure*}
\begin{figure*}[btp]
\centering
\import{figs/plots}{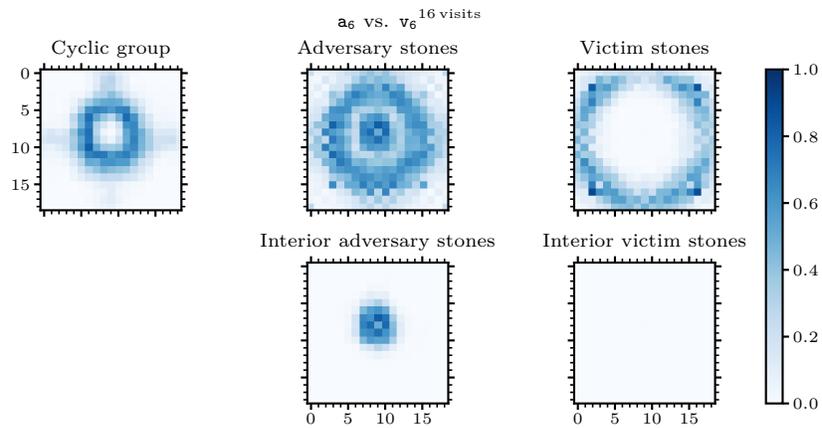}
\caption{
  Heat map showing the cyclic attack made by \attackiter{6} against \defenseiter{6}.
}
\label{fig:cyclic-heatmap-r6}
\end{figure*}
\begin{figure*}[btp]
\centering
\import{figs/plots}{cyclic-heatmap-diff-r7.pgf}
\caption{
  Difference between the heat maps of \attackiter{7} (Figure~\ref{fig:cyclic-heatmap-r7}) and \attackiter{6} (Figure~\ref{fig:cyclic-heatmap-r6}).
}
\label{fig:cyclic-heatmap-diff-r7}
\end{figure*}
\begin{figure*}[btp]
\centering
\import{figs/plots}{cyclic-heatmap-r7.pgf}
\caption{
  Heat map showing the cyclic attack made by \attackiter{7} against \defenseiter{7}.
}
\label{fig:cyclic-heatmap-r7}
\end{figure*}
\begin{figure*}[btp]
\centering
\import{figs/plots}{cyclic-heatmap-diff-r8.pgf}
\caption{
  Difference between the heat maps of \attackiter{8} (Figure~\ref{fig:cyclic-heatmap-r8}) and \attackiter{7} (Figure~\ref{fig:cyclic-heatmap-r7}).
}
\label{fig:cyclic-heatmap-diff-r8}
\end{figure*}
\begin{figure*}[btp]
\centering
\import{figs/plots}{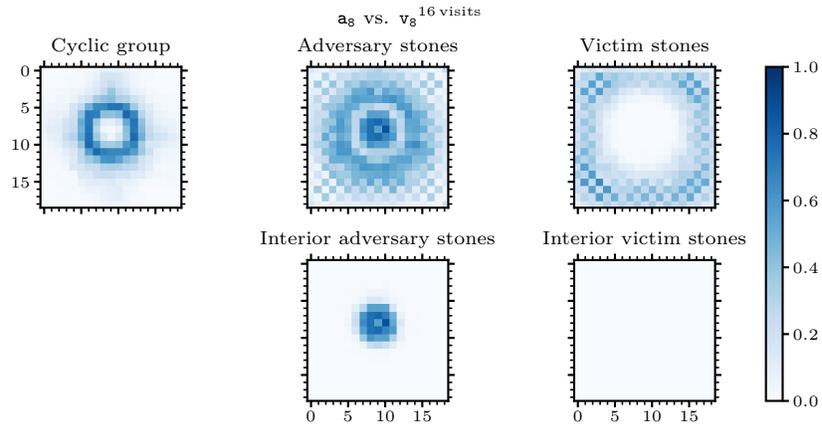}
\caption{
  Heat map showing the cyclic attack made by \attackiter{8} against \defenseiter{8}.
}
\label{fig:cyclic-heatmap-r8}
\end{figure*}
\begin{figure*}[btp]
\centering
\import{figs/plots}{cyclic-heatmap-diff-r9.pgf}
\caption{
  Difference between the heat maps of \attackiter{9} (Figure~\ref{fig:cyclic-heatmap-r9}) and \attackiter{8} (Figure~\ref{fig:cyclic-heatmap-r8}).
}
\label{fig:cyclic-heatmap-diff-r9}
\end{figure*}
\begin{figure*}[btp]
\centering
\import{figs/plots}{cyclic-heatmap-r9.pgf}
\caption{
  Heat map showing the cyclic attack made by \attackiter{9} against \defenseiter{9}.
}
\label{fig:cyclic-heatmap-r9}
\end{figure*}

Figs.~\ref{fig:cyclic-heatmap-r1} to {fig:cyclic-heatmap-r9} show heat maps for each adversary \attackiter{1} through \attackiter{9} trained in iterated adversarial training against their corresponding victims at 16 victim visits:
\begin{itemize}
    \item \attackiter{1} (Figs.~\ref{fig:cyclic-heatmap-r1} and \ref{fig:cyclic-heatmap-diff-r1}) has a less consistent structure to the stones outside the cyclic group than \origcyclic{}, which tends to form a boundary of stones near the edge of the board outside the cycle.
    \item \attackiter{2} (Figs.~\ref{fig:cyclic-heatmap-r2} and \ref{fig:cyclic-heatmap-diff-r2}) forms a larger cycle than \attackiter{1}.
    \item \attackiter{3} (Figs.~\ref{fig:cyclic-heatmap-r3} and \ref{fig:cyclic-heatmap-diff-r3}) moves the cycle towards the center on one axis, and the cycle shrinks again but with less consistent shapes.
    \item \attackiter{4} (Figs.~\ref{fig:cyclic-heatmap-r4} and \ref{fig:cyclic-heatmap-diff-r4}) moves the cycle towards the center along the other axis.
    \item \attackiter{5} (Figs.~\ref{fig:cyclic-heatmap-r5} and \ref{fig:cyclic-heatmap-diff-r5}) makes the cycle larger.
    \item \attackiter{6} (Figs.~\ref{fig:cyclic-heatmap-r6} and \ref{fig:cyclic-heatmap-diff-r6}) does not show much qualitative difference in the heat maps.
    \item \attackiter{7} (Figs.~\ref{fig:cyclic-heatmap-r7} and \ref{fig:cyclic-heatmap-diff-r7}) tends to place stones on board locations of a particular parity near the boundaries of the board, leading to a checkerboard pattern in the heat map.
    \item \attackiter{8} (Figs.~\ref{fig:cyclic-heatmap-r8} and \ref{fig:cyclic-heatmap-diff-r8}) does not show much change.
    \item \attackiter{9} (Figs.~\ref{fig:cyclic-heatmap-r9} and \ref{fig:cyclic-heatmap-diff-r9}) shrinks the cycle slightly.
\end{itemize}

\section{Extra Experimental Plots}

This section collects additional, visually large plots referenced in previous sections.

\subsection{Individual Iterated Adversarial Training Plots}

Whereas Figures~\ref{fig:vs-gpu-days-h} and \ref{fig:vs-gpu-days-r} concatenate all defense iterations into one plot and all attack iterations into another plot,
Figures~\ref{fig:vs-gpu-days-h-each} and \ref{fig:vs-gpu-days-r-each} give training progress plots for each iteration separately.

\begin{figure*}[btp]
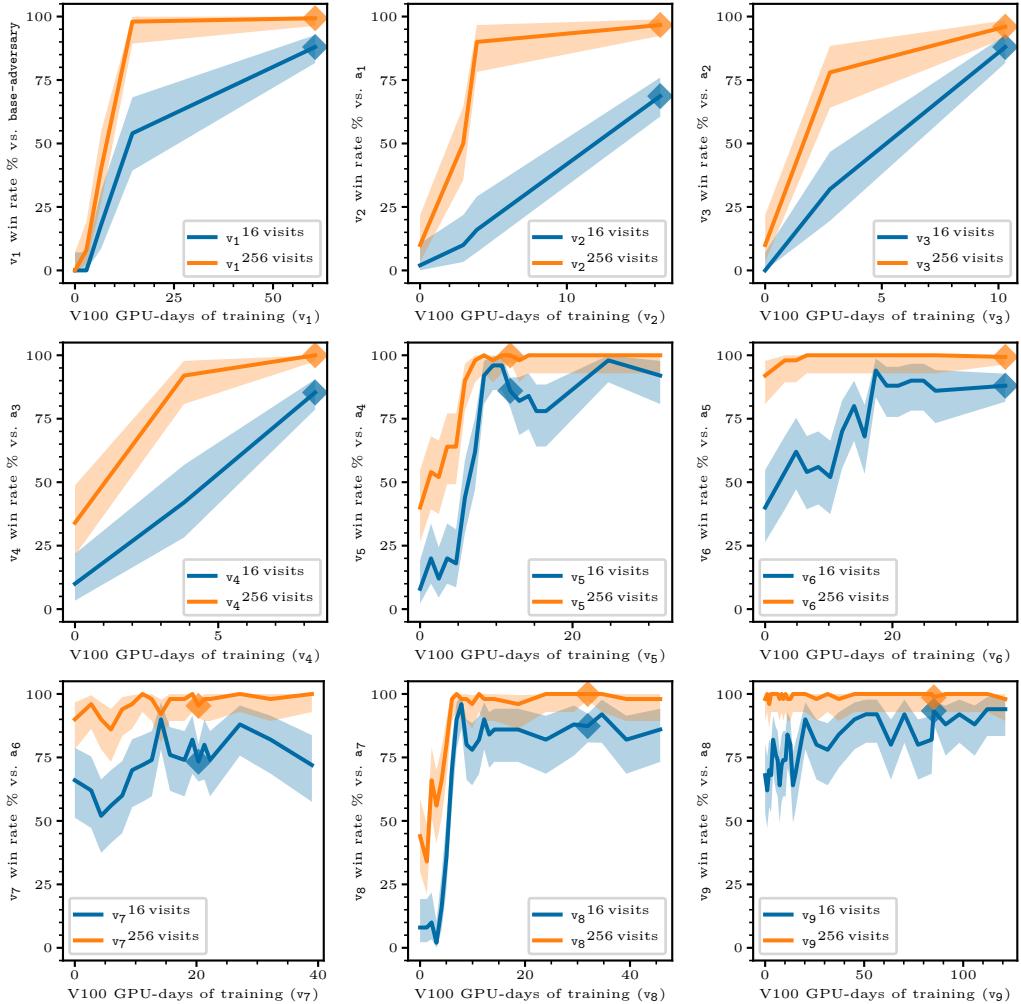

  \centering
  \begin{subfigure}{0.32\textwidth}
  \centering
  \input{figs/plots/win-rate-vs-gpu-days-h1.pgf}
  \end{subfigure}
  \begin{subfigure}{0.32\textwidth}
  \centering
  \input{figs/plots/win-rate-vs-gpu-days-h2.pgf}
  \end{subfigure}
  \begin{subfigure}{0.32\textwidth}
  \centering
  \input{figs/plots/win-rate-vs-gpu-days-h3.pgf}
  \end{subfigure}
  \begin{subfigure}{0.32\textwidth}
  \centering
  \input{figs/plots/win-rate-vs-gpu-days-h4.pgf}
  \end{subfigure}
  \begin{subfigure}{0.32\textwidth}
  \centering
  \input{figs/plots/win-rate-vs-gpu-days-h5.pgf}
  \end{subfigure}
  \begin{subfigure}{0.32\textwidth}
  \centering
  \input{figs/plots/win-rate-vs-gpu-days-h6.pgf}
  \end{subfigure}
  \begin{subfigure}{0.32\textwidth}
  \centering
  \input{figs/plots/win-rate-vs-gpu-days-h7.pgf}
  \end{subfigure}
  \begin{subfigure}{0.32\textwidth}
  \centering
  \input{figs/plots/win-rate-vs-gpu-days-h8.pgf}
  \end{subfigure}
  \begin{subfigure}{0.32\textwidth}
  \centering
  \input{figs/plots/win-rate-vs-gpu-days-h9.pgf}
  \end{subfigure}
  \caption{
    Win rate ($y$-axis) of each \defenseiter{N} (\plotdiamond{}) against \attackiter{N-1} throughout
    \defenseiter{N}'s training ($x$-axis).
    The curves for \defenseiter{1} to \defenseiter{4} only have a few data points along the $x$-axis as intermediate checkpoints were lost.
  }
  \label{fig:vs-gpu-days-h-each}
\end{figure*}

\begin{figure*}[btp]
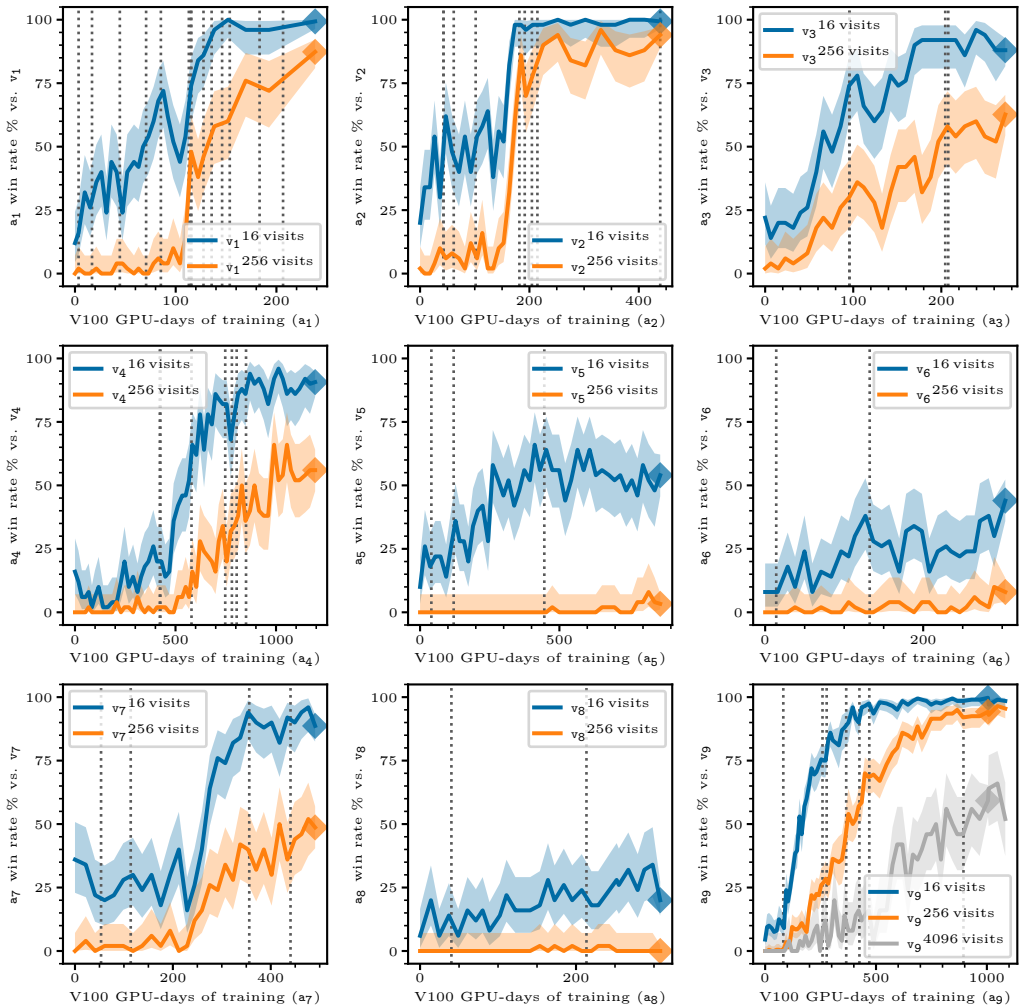

  \centering
  \begin{subfigure}{0.32\textwidth}
  \centering
  \input{figs/plots/win-rate-vs-gpu-days-r1.pgf}
  \end{subfigure}
  \begin{subfigure}{0.32\textwidth}
  \centering
  \input{figs/plots/win-rate-vs-gpu-days-r2.pgf}
  \end{subfigure}
  \begin{subfigure}{0.32\textwidth}
  \centering
  \input{figs/plots/win-rate-vs-gpu-days-r3.pgf}
  \end{subfigure}
  \begin{subfigure}{0.32\textwidth}
  \centering
  \input{figs/plots/win-rate-vs-gpu-days-r4.pgf}
  \end{subfigure}
  \begin{subfigure}{0.32\textwidth}
  \centering
  \input{figs/plots/win-rate-vs-gpu-days-r5.pgf}
  \end{subfigure}
  \begin{subfigure}{0.32\textwidth}
  \centering
  \input{figs/plots/win-rate-vs-gpu-days-r6.pgf}
  \end{subfigure}
  \begin{subfigure}{0.32\textwidth}
  \centering
  \input{figs/plots/win-rate-vs-gpu-days-r7.pgf}
  \end{subfigure}
  \begin{subfigure}{0.32\textwidth}
  \centering
  \input{figs/plots/win-rate-vs-gpu-days-r8.pgf}
  \end{subfigure}
  \begin{subfigure}{0.32\textwidth}
  \centering
  \input{figs/plots/win-rate-vs-gpu-days-r9.pgf}
  \end{subfigure}
  \caption{
      Win rate ($y$-axis) of each \attackiter{N} against \defenseiter{N} throughout
      \attackiter{N}'s training ($x$-axis).
      Dotted lines represent advancing to the next victim in the curriculum.
  }
  \label{fig:vs-gpu-days-r-each}
\end{figure*}

\subsection{Training Steps Plots}

In Figs.~\ref{fig:vs-steps-continuous} to {fig:vs-steps-r-each} we display versions of previous plots but use victim-play or self-play training steps on the $x$-axis to measure training time instead of GPU-days. We tended to use GPU-days throughout this paper since it is a unit that is more understandable for readers, but as GPU-days are machine-dependent, training steps may be more useful for other researchers who want to compare our runs to other KataGo-like training runs.

\begin{figure*}[btp]
\centering
\input{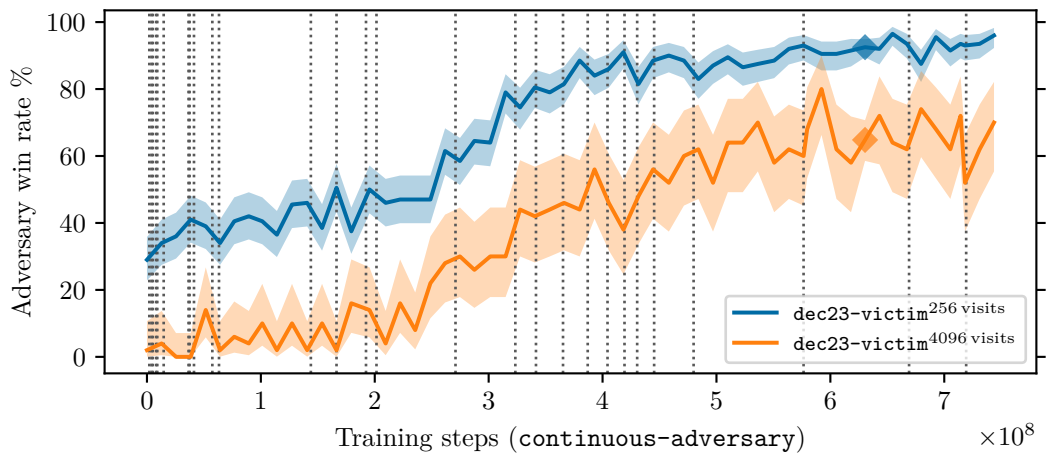}
\caption{
  This plot is the same as Figure~\ref{fig:vs-gpu-days-gift} but with training steps on the $x$-axis instead of GPU-days.
}
\label{fig:vs-steps-gift}
\end{figure*}

\begin{figure*}[btp]
\centering
\input{figs/plots/win-rate-vs-steps-continuous.pgf}
\caption{
  This plot is the same as Figure~\ref{fig:vs-gpu-days-continuous} but with training steps on the $x$-axis instead of GPU-days.
}
\label{fig:vs-steps-continuous}
\end{figure*}

\begin{figure*}[btp]
\centering
\input{figs/plots/win-rate-vs-steps-large.pgf}
\caption{
  This plot is the same as Figure~\ref{fig:vs-gpu-days-large} but with training steps on the $x$-axis instead of GPU-days.
}
\label{fig:vs-steps-large}
\end{figure*}

\begin{figure*}[btp]
\centering
\input{figs/plots/win-rate-vs-steps-h-all.pgf}
\caption{
  This plot is the same as Figure~\ref{fig:vs-gpu-days-h} but with training steps on the $x$-axis instead of GPU-days.
}
\label{fig:vs-steps-h}
\end{figure*}
\begin{figure*}[btp]
\centering
\input{figs/plots/win-rate-vs-steps-r-all.pgf}
\caption{
  This plot is the same as Figure~\ref{fig:vs-gpu-days-r} but with training steps on the $x$-axis instead of GPU-days.
}
\label{fig:vs-steps-k}
\end{figure*}

\begin{figure*}[btp]
\centering
\input{figs/plots/win-rate-vs-steps-attack-h9.pgf}
\caption{
  This plot is the same as Figure~\ref{fig:vs-gpu-days-attack-h9} but with training steps on the $x$-axis instead of GPU-days.
}
\label{fig:vs-steps-attack-h9}
\end{figure*}

\begin{figure*}[btp]
\centering
\input{figs/plots/win-rate-vs-steps-stall.pgf}
\caption{
  This plot is the same as Figure~\ref{fig:vs-gpu-days-stall} but with training steps on the $x$-axis instead of GPU-days.
}
\label{fig:vs-steps-stall}
\end{figure*}

\begin{figure*}[btp]
\centering
\input{figs/plots/elo-vs-vit-steps.pgf}
\caption{
  This plot is the same as Figure~\ref{fig:elo-vs-vit-gpu-days} but with training steps on the $x$-axis instead of GPU-days.
}
\label{fig:elo-vs-vit-steps}
\end{figure*}

\begin{figure*}[btp]
\centering
\input{figs/plots/win-rate-vs-steps-attack-vit.pgf}
\caption{
  This plot is the same as Figure~\ref{fig:vs-gpu-days-attack-vit} but with training steps on the $x$-axis instead of GPU-days.
}
\label{fig:vs-steps-attack-vit}
\end{figure*}

\begin{figure*}[btp]
\centering
\input{figs/plots/vs-vit-steps-origcyclic.pgf}
\caption{
  This plot is the same as Figure~\ref{fig:vs-vit-gpu-days-origcyclic} but with training steps on the $x$-axis instead of GPU-days.
}
\label{fig:vs-vit-steps-origcyclic}
\end{figure*}
\begin{figure*}[btp]
\centering
\input{figs/plots/vs-vit-steps-vitadversary.pgf}
\caption{
  This plot is the same as Figure~\ref{fig:vs-vit-gpu-days-attack-vit} but with training steps on the $x$-axis instead of GPU-days.
}
\label{fig:vs-vit-steps-attack-vit}
\end{figure*}
\begin{figure*}[btp]
\centering
\input{figs/plots/vs-control-b10-steps-origcyclic.pgf}
\caption{
  This plot is the same as Figure~\ref{fig:vs-control-b10-gpu-days-origcyclic} but with training steps on the $x$-axis instead of GPU-days.
}
\label{fig:vs-control-b10-steps-origcyclic}
\end{figure*}

\begin{figure*}[btp]
\centering
\input{figs/plots/vs-control-b10-steps-vitadversary.pgf}
\caption{
  This plot is the same as Figure~\ref{fig:vs-control-b10-gpu-days-attack-vit} but with training steps on the $x$-axis instead of GPU-days.
}
\label{fig:vs-control-b10-steps-attack-vit}
\end{figure*}

\begin{figure*}[btp]
  \centering
  \begin{subfigure}{0.32\textwidth}
  \centering
  \input{figs/plots/win-rate-vs-steps-h1.pgf}
  \end{subfigure}
  \begin{subfigure}{0.32\textwidth}
  \centering
  \input{figs/plots/win-rate-vs-steps-h2.pgf}
  \end{subfigure}
  \begin{subfigure}{0.32\textwidth}
  \centering
  \input{figs/plots/win-rate-vs-steps-h3.pgf}
  \end{subfigure}
  \begin{subfigure}{0.32\textwidth}
  \centering
  \input{figs/plots/win-rate-vs-steps-h4.pgf}
  \end{subfigure}
  \begin{subfigure}{0.32\textwidth}
  \centering
  \input{figs/plots/win-rate-vs-steps-h5.pgf}
  \end{subfigure}
  \begin{subfigure}{0.32\textwidth}
  \centering
  \input{figs/plots/win-rate-vs-steps-h6.pgf}
  \end{subfigure}
  \begin{subfigure}{0.32\textwidth}
  \centering
  \input{figs/plots/win-rate-vs-steps-h7.pgf}
  \end{subfigure}
  \begin{subfigure}{0.32\textwidth}
  \centering
  \input{figs/plots/win-rate-vs-steps-h8.pgf}
  \end{subfigure}
  \begin{subfigure}{0.32\textwidth}
  \centering
  \input{figs/plots/win-rate-vs-steps-h9.pgf}
  \end{subfigure}
  \caption{
    This plot is the same as Figure~\ref{fig:vs-gpu-days-h-each} but with training steps on the $x$-axis instead of GPU-days.
  }
  \label{fig:vs-steps-h-each}
\end{figure*}

\begin{figure*}[btp]
  \centering
  \begin{subfigure}{0.32\textwidth}
  \centering
  \input{figs/plots/win-rate-vs-steps-r1.pgf}
  \end{subfigure}
  \begin{subfigure}{0.32\textwidth}
  \centering
  \input{figs/plots/win-rate-vs-steps-r2.pgf}
  \end{subfigure}
  \begin{subfigure}{0.32\textwidth}
  \centering
  \input{figs/plots/win-rate-vs-steps-r3.pgf}
  \end{subfigure}
  \begin{subfigure}{0.32\textwidth}
  \centering
  \input{figs/plots/win-rate-vs-steps-r4.pgf}
  \end{subfigure}
  \begin{subfigure}{0.32\textwidth}
  \centering
  \input{figs/plots/win-rate-vs-steps-r5.pgf}
  \end{subfigure}
  \begin{subfigure}{0.32\textwidth}
  \centering
  \input{figs/plots/win-rate-vs-steps-r6.pgf}
  \end{subfigure}
  \begin{subfigure}{0.32\textwidth}
  \centering
  \input{figs/plots/win-rate-vs-steps-r7.pgf}
  \end{subfigure}
  \begin{subfigure}{0.32\textwidth}
  \centering
  \input{figs/plots/win-rate-vs-steps-r8.pgf}
  \end{subfigure}
  \begin{subfigure}{0.32\textwidth}
  \centering
  \input{figs/plots/win-rate-vs-steps-r9.pgf}
  \end{subfigure}
  \caption{
    This plot is the same as Figure~\ref{fig:vs-gpu-days-r-each} but with training steps on the $x$-axis instead of GPU-days.
  }
  \label{fig:vs-steps-r-each}
\end{figure*}

\fi

\end{document}